\documentclass{article}

\usepackage[preprint, nonatbib]{neurips_2019}

\usepackage{enumitem}

\usepackage[utf8]{inputenc} 
\usepackage[T1]{fontenc}    
\usepackage{hyperref}       
\usepackage{url}            
\usepackage{booktabs}       
\usepackage{amsfonts}       
\usepackage{nicefrac}       
\usepackage{microtype}      
\usepackage{amsmath}
\usepackage{bm}
\usepackage{graphicx}
\usepackage{xcolor}
\usepackage{subcaption}
\usepackage{wrapfig,lipsum}

\usepackage{algorithm}
\usepackage{algorithmicx}
\usepackage{algpseudocode}
\usepackage{subcaption}
\usepackage{enumitem}

\title{Exploring Model-based Planning with Policy Networks}

\author{Tingwu Wang$^{1,2}$\& Jimmy Ba$^{1,2}$ \\
$^1$ Department of Computer Science, University of Toronto\,$^2$ Vector Institute \\
\texttt{\{tingwuwang,jba\}@cs.toronto.edu}}

\newcommand{\ourmodel}{model-based policy planning}
\newcommand{\Ourmodel}{Model-Based Policy Planning}
\newcommand{\ourmodelshort}{POPLIN}
\newcommand{\ourmodelA}{model-based policy planning in action space}
\newcommand{\OurmodelA}{Model-based Policy Planning in Action Space}
\newcommand{\ourmodelP}{model-based policy planning in parameter space}
\newcommand{\OurmodelP}{Model-based Policy Planning in Parameter Space}
\newcommand{\ourmodelshortA}{POPLIN-A}
\newcommand{\ourmodelshortP}{POPLIN-P}

\begin{document}

\maketitle

\begin{abstract}
Model-based reinforcement learning (MBRL) with model-predictive control or online planning has shown great potential for locomotion control tasks in terms of both sample efficiency and asymptotic performance.
Despite their initial successes,
the existing planning methods search from candidate sequences randomly generated in the action space,
which is inefficient in complex high-dimensional environments. 
In this paper, 
we propose a novel MBRL algorithm, \ourmodel{} (\ourmodelshort{}), that combines policy networks with online planning.
More specifically, we formulate action planning at each time-step as an optimization problem using neural networks.
We experiment with both optimization w.r.t. the action sequences initialized from the policy network,
and also online optimization directly w.r.t. the parameters of the policy network.
We show that \ourmodelshort{} obtains state-of-the-art performance in the MuJoCo benchmarking environments,
being about 3x more sample efficient than the state-of-the-art algorithms, such as PETS, TD3 and SAC.
To explain the effectiveness of our algorithm,
we show that the optimization surface in parameter space is smoother than in action space.
Further more,
we found the distilled policy network can be effectively applied without the expansive model predictive control during test time for some environments such as Cheetah. Code is released in \url{https://github.com/WilsonWangTHU/POPLIN}.
\end{abstract}

\section{Introduction}
A model-based reinforcement learning (MBRL) agent learns its internal model of the world, i.e. the dynamics, from repeated interactions with the environment.
With the learnt dynamics, a MBRL agent can for example perform online planning, interact with imaginary data, or optimize the controller through dynamics, which provides significantly better sample efficiency~\cite{deisenroth2011pilco, sutton1990integrated, levine2014learning, levine2013guided}.
However, MBRL algorithms generally do not scale well with the increasing complexity of the reinforcement learning (RL) tasks in practice.
And modelling errors in dynamics that accumulate with time-steps greatly limit the applications of MBRL algorithms.
As a result, many latest progresses in RL has been made with model-free reinforcement learning (MFRL) algorithms that are capable of solving complex tasks at the cost of large number of samples~\cite{ppo, deepmindppo, trpo, dqn, lillicrap2015continuous, haarnoja2018soft}.

With the success of deep learning,
a few recent works have proposed to learn neural network-based dynamics models for MBRL.
Among them, random shooting algorithms (RS), which uses model-predictive control (MPC),
is shown to have good robustness and scalability~\cite{richards2005robust}.
In shooting algorithms, the agent randomly generates action sequences,
use the dynamics to predict the future states,
and choose the first action from the sequence with the best expected reward.
However, RS usually has worse asymptotic performance than model-free controllers~\cite{nagabandi2017neural}, 
and the authors of the the PETS algorithm~\cite{chua2018deep} suggest that the performance of RS is directly affected by the quality of the learnt dynamics.
They propose a probabilistic ensemble to capture model uncertainty,
which enables PETS algorithm to achieve both better sample efficiency and better asymptotic performance than state-of-the-art model-free controllers in environments such as Cheetah~\cite{gym}.
However, PETS is not as effective on environments with higher dimensionality.

\begin{figure}[!t]
    \centering
    \begin{subfigure}{0.24\textwidth}
        \centering
        \includegraphics[width=\linewidth]{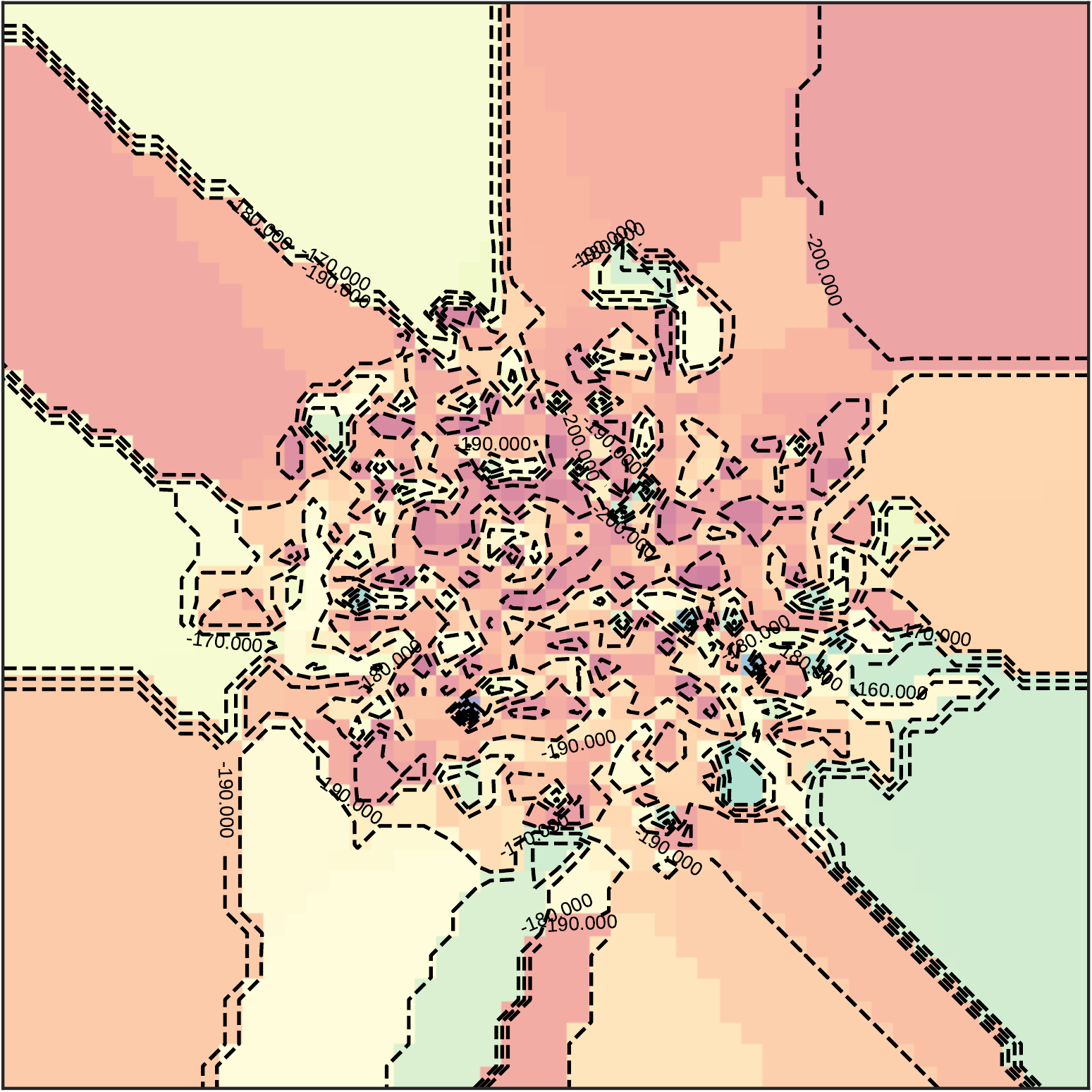}
        (a1) Reward Surface
    \end{subfigure}
    \begin{subfigure}{0.24\textwidth}
        \centering
        \includegraphics[width=\linewidth]{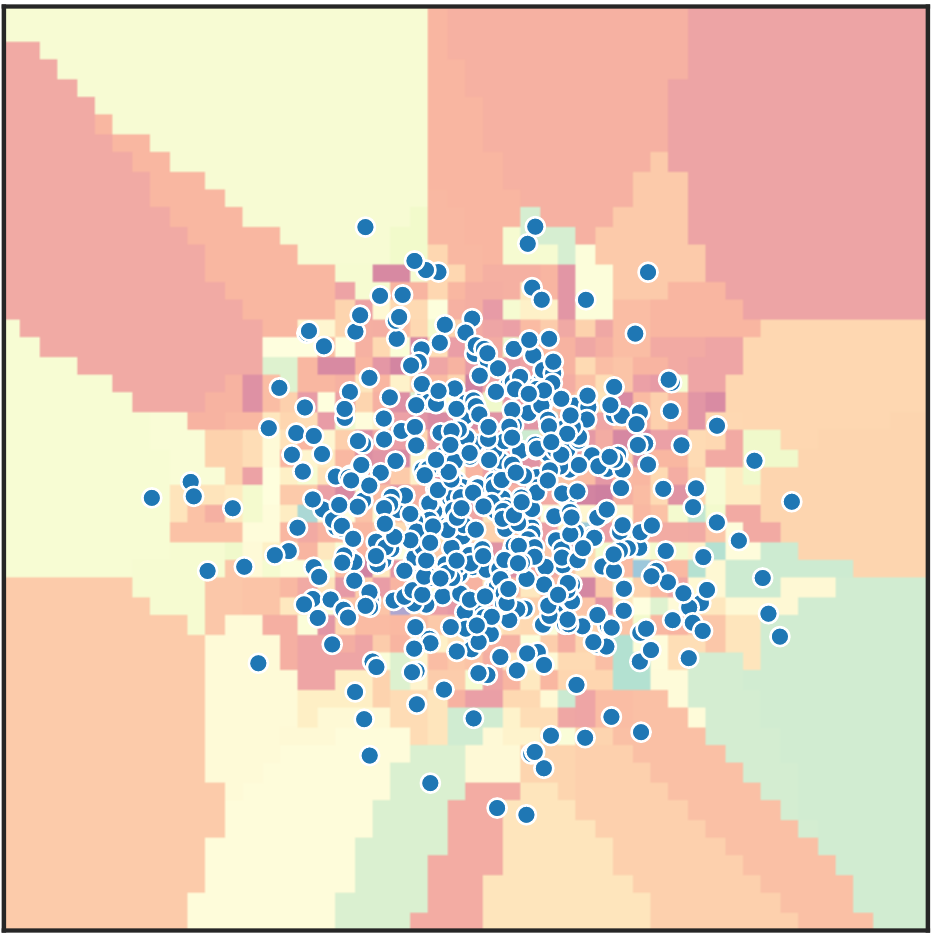}
        (a2) PETS Iteration 1
    \end{subfigure}
    \begin{subfigure}{0.24\textwidth}
        \centering
        \includegraphics[width=\linewidth]{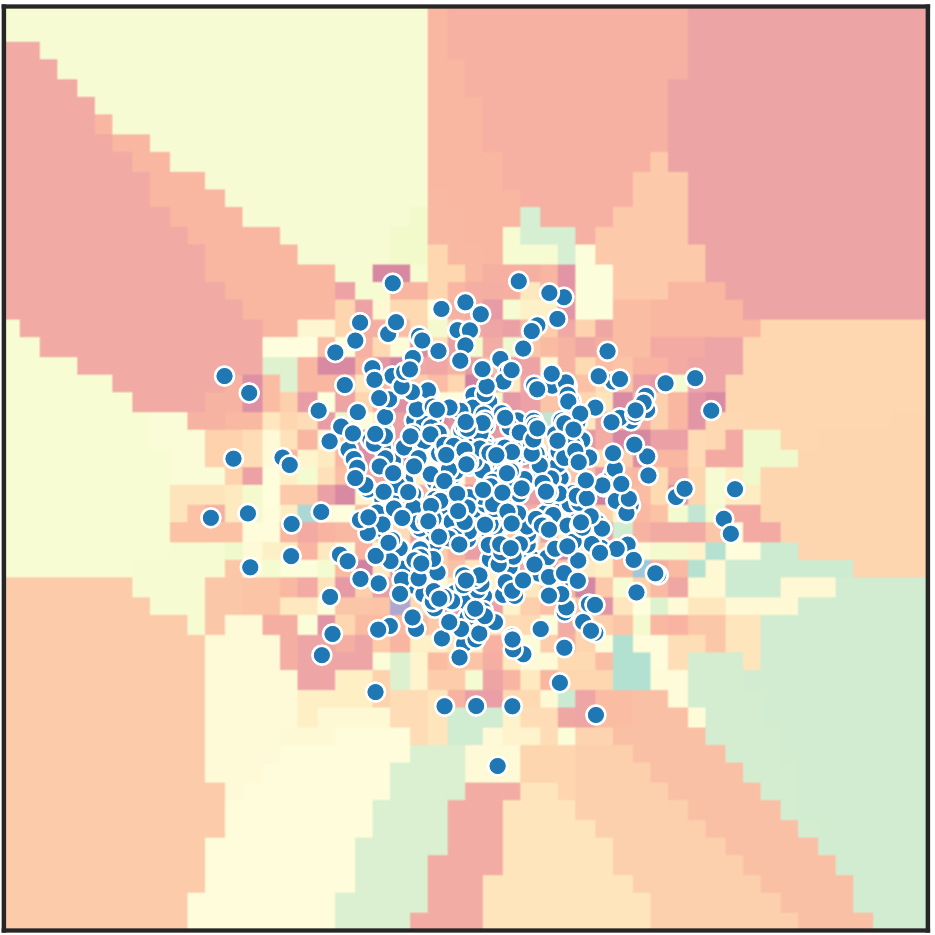}
        (a3) PETS Iteration 2
    \end{subfigure}
    \begin{subfigure}{0.24\textwidth}
        \centering
        \includegraphics[width=\linewidth]{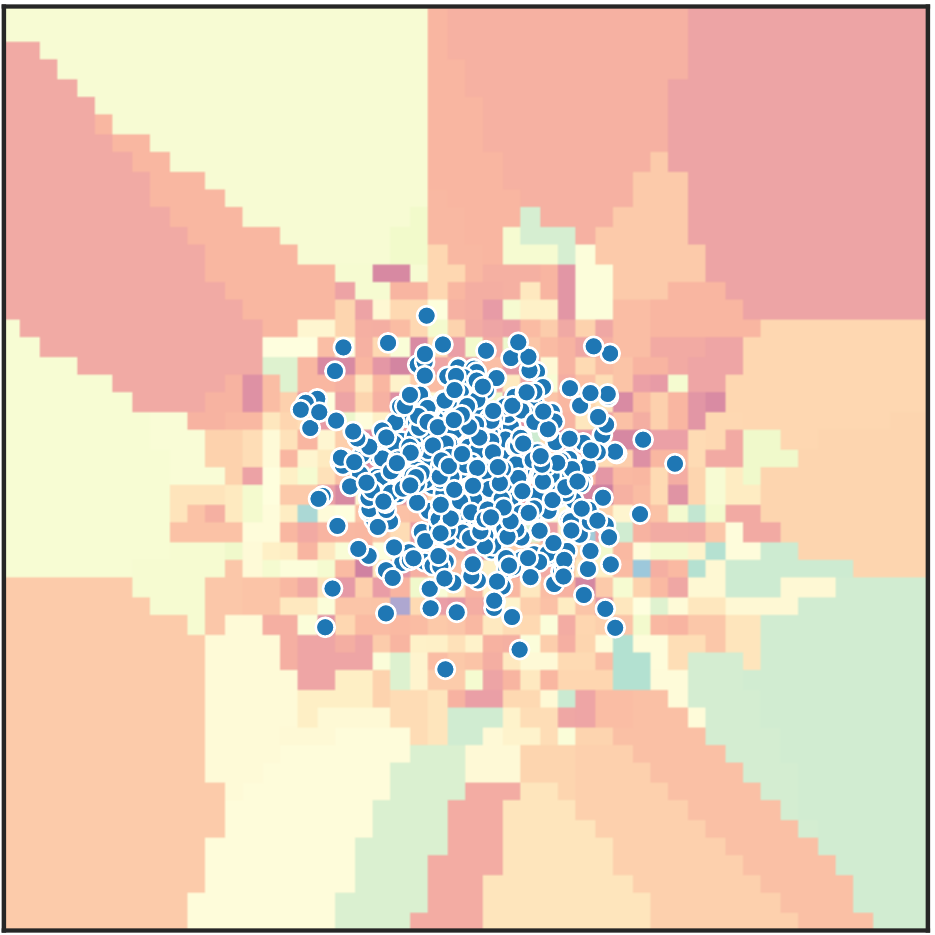}
        (a4) PETS Iteration 3
    \end{subfigure}
    
    \begin{subfigure}{0.24\textwidth}
        \centering
        \includegraphics[width=\linewidth]{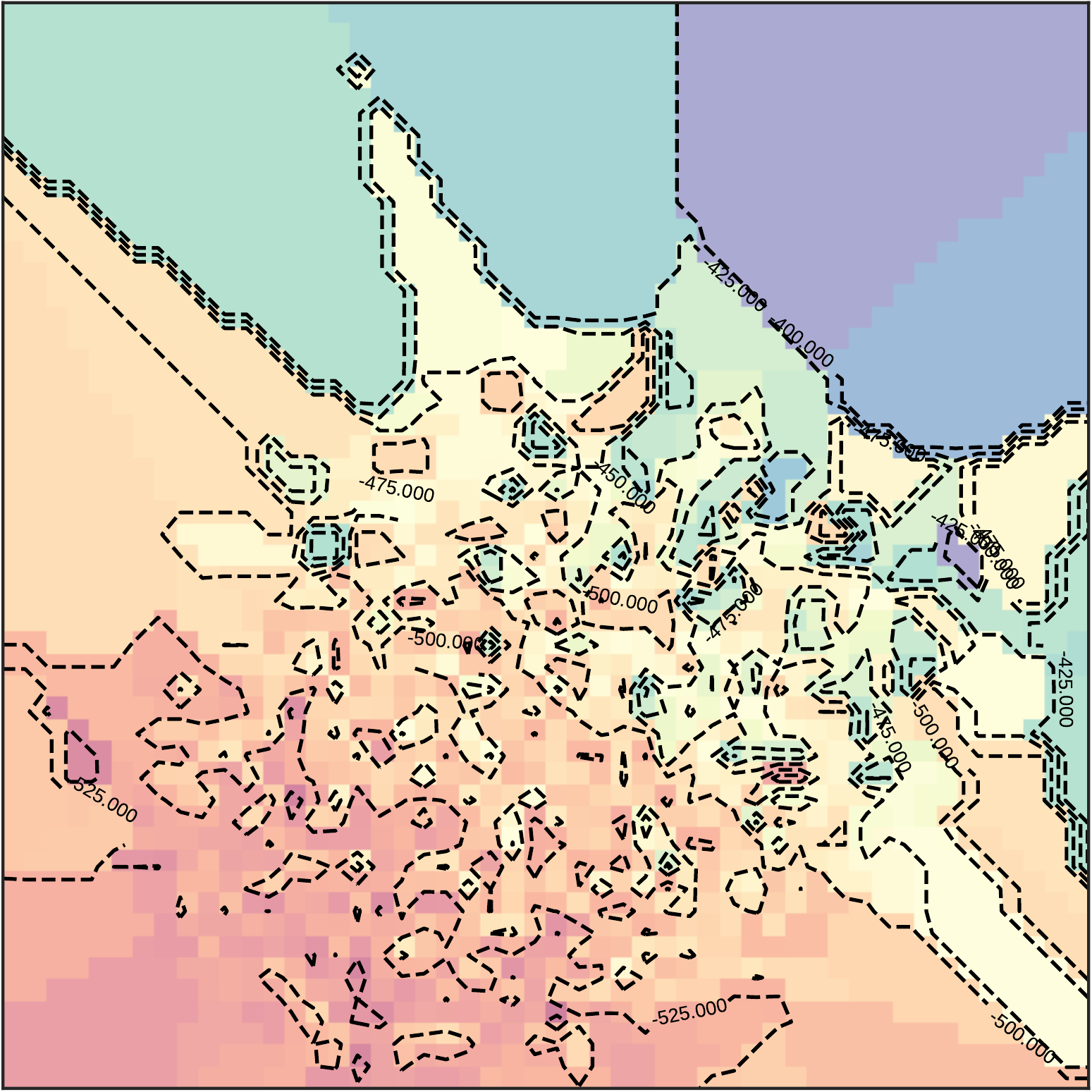}
        (b1) Reward Surface
    \end{subfigure}
    \begin{subfigure}{0.24\textwidth}
        \centering
        \includegraphics[width=\linewidth]{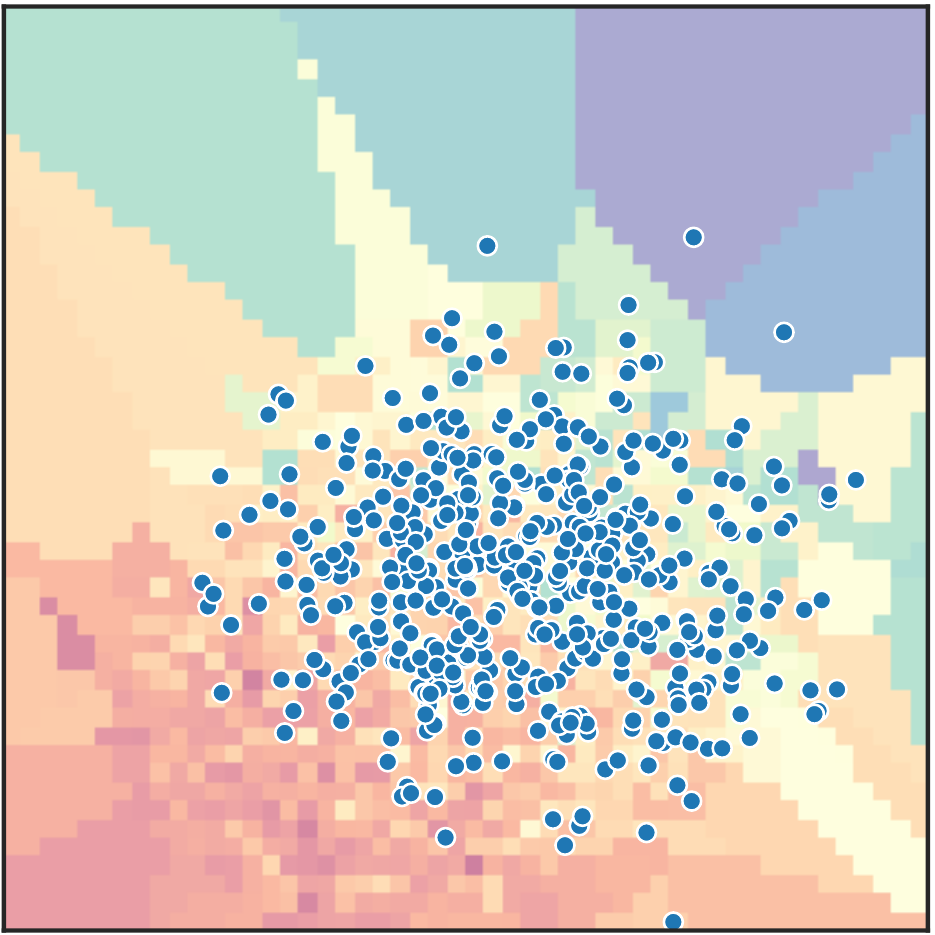}
        (b2) \ourmodelshort{} Iteration 1
    \end{subfigure}
    \begin{subfigure}{0.24\textwidth}
        \centering
        \includegraphics[width=\linewidth]{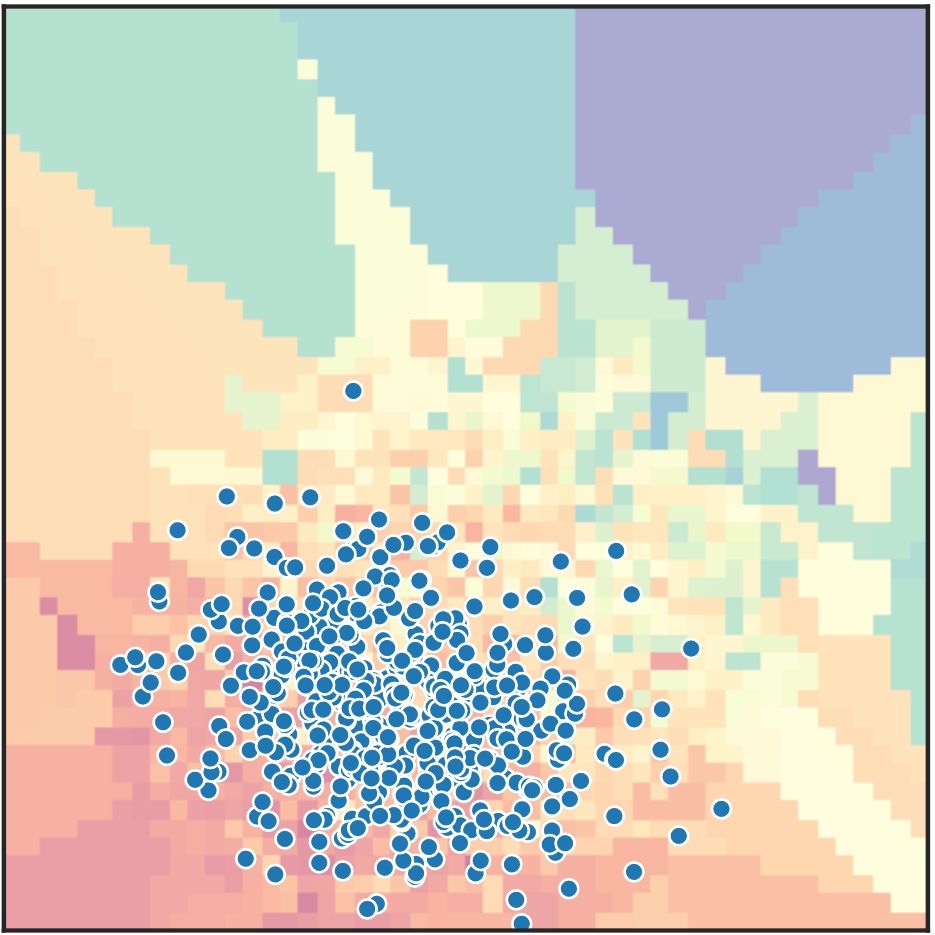}
        (b3) \ourmodelshort{} Iteration 2
    \end{subfigure}
    \begin{subfigure}{0.24\textwidth}
        \centering
        \includegraphics[width=\linewidth]{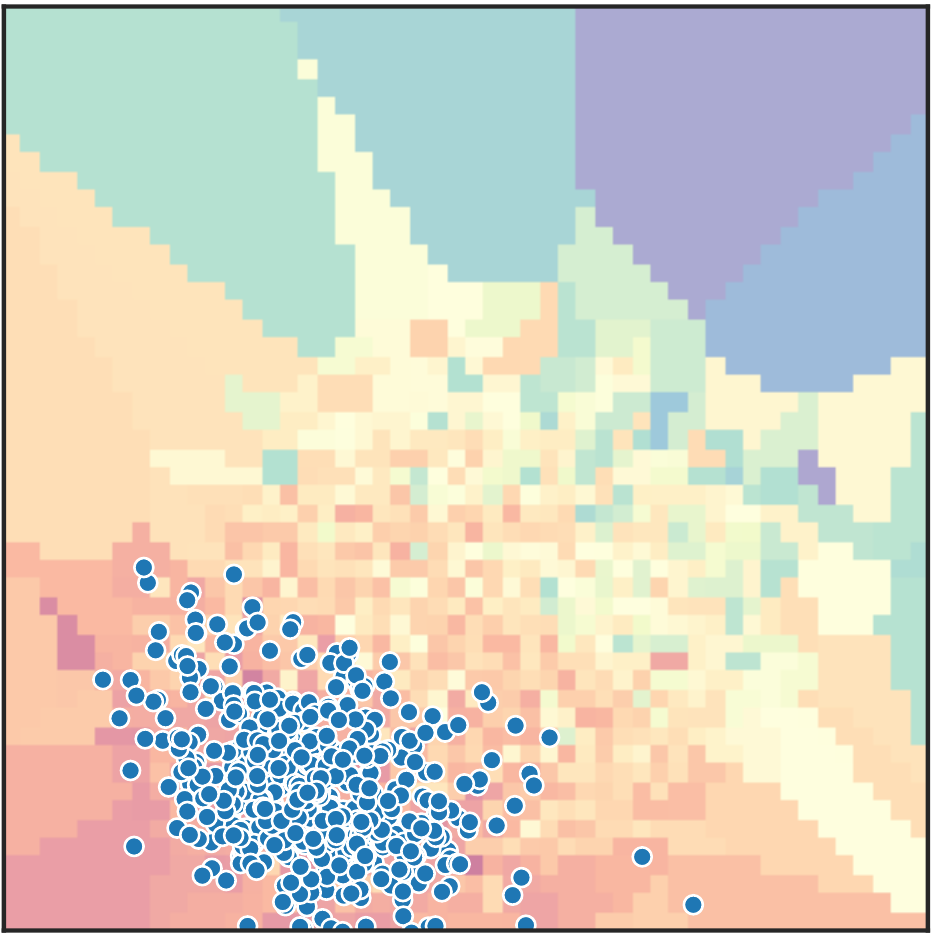}
        (b4) \ourmodelshort{} Iteration 3
    \end{subfigure}
    
    \caption{We transform each planned candidate action trajectory with PCA into a 2D blue scatter.
    The top and bottom figures are respectively the visualization of PETS~\cite{chua2018deep} and our algorithm. The red area has higher reward.
    From left to right, we show how candidate trajectories are updated,
    across different planning iterations within one time-step.
    As we can see, while both reward surface is not smooth with respect to action trajectory.
    POPLIN, using policy networks, has much better search efficiency, while PETS is stuck around its initialization.
    The details are in section~\ref{section:reward_surface}. 
    }
    \label{fig:reward_surface}
    \vspace{-0.5cm}
\end{figure}

In this paper, we explore MBRL algorithms from a different perspective, where we treat the planning at each time-step as an optimization problem.
Random search in action space, as what is being done in state-of-the-art MBRL algorithms such as PETS, is insufficient for more complex environments.
On the one hand, we are inspired by the success of AlphaGo~\cite{alphago,silver2017mastering},
where a policy network is used to generate proposals for the Monte-Carlo tree search.
On the other hand,
we are inspired by the recent research into understanding deep neural networks~\cite{nguyen2017loss, li2018visualizing, soudry2017exponentially}.
Deep neural networks, frequently observed in practices, is much less likely to get stuck in sub-optimal points.
In Figure~\ref{fig:reward_surface},
we apply principal component analysis (PCA) on the action sequences generated in each planning iteration within one time-step.
The reward surface of the action space is not smooth and prone to local-minimas.
We argue that optimization in the policy network's parameter space will be more efficient.
Furthermore, we note that the state-of-the-art MBRL algorithm with MPC cannot be applied real-time.
We therefore experiment with different policy network distillation schemes for fast control without MPC.
To sum up, the contribution of this paper is three-fold:
\begin{itemize}[leftmargin=*]
    \item We apply policy networks to generate proposals for MPC in high dimensional locomotion control problems with unknown dynamics.
    \item We formulate planning as optimization with neural networks, and propose policy planning in parameter space, which obtain state-of-the-art performance on current bench-marking environments, being about 3x more sample efficient than the previous state-of-the-art algorithm, such as PETS~\cite{chua2018deep}, TD3~\cite{fujimoto2018addressing} and SAC~\cite{haarnoja2018soft}.
    \item We also explore policy network distillation from the planned trajectories. We found the distilled policy network alone achieves high performance on environments like Cheetah without the expansive online planning.
\end{itemize}
\section{Related Work}
Model-based reinforcement learning (MBRL) has been long studied.
Dyna~\cite{sutton1990integrated, sutton1991dyna} algorithm alternately performs sampling in the real environments and optimize the controllers on the learned model of the environments.
Other pioneering work includes PILCO~\cite{deisenroth2011pilco},
where the authors model the dynamics using Gaussian Process and directly optimize the surrogate expected reward.
Effective as it is to solve simple environments,
PILCO heavily suffers the curse of dimensionality.
In~\cite{levine2014learning, levine2013guided, levine2016end, chebotar2017combining, zhang2018solar},
the authors propose guided policy search (GPS).
GPS uses iLQG~\cite{li2004iterative, todorov2005generalized, tassa2012synthesis} as the local controller,
and distill the knowledge into a policy neural network.
In SVG~\cite{heess2015learning},
the authors uses stochastic value gradient so that the stochastic policy network can be optimized by back-propagation through the learnt dynamics network with off-policy data.
Recently with the progress of model-free algorithms such as~\cite{trpo,ppo},
in~\cite{kurutach2018model, luo2018algorithmic} the authors propose modern variants of Dyna,
where TRPO~\cite{trpo} is used to optimize the policy network using data generated by the learnt dynamics.
Most recently, random shooting methods such as~\cite{nagabandi2017neural, chua2018deep} have shown its robustness and effectiveness on benchmarking environments.
PETS algorithm~\cite{chua2018deep} is considered by many to be the state-of-the-art MBRL algorithm,
which we discuss in detail in section~\ref{section:background}.
Dynamics is also used to obtain better value estimation to speed up training~\cite{gu2016continuous, feinberg2018model, buckman2018sample}.
Latent dynamics models using VAE~\cite{kingma2013auto} are commonly used to solve problems with image input~\cite{ha2018recurrent, ha2018world, hafner2018learning, kaiser2019model}.
\section{Background}\label{section:background}
\subsection{Reinforcement Learning}\label{section:bg_rl}
In reinforcement learning, the problem of solving the given task is formulated as a infinite-horizon discounted Markov decision process.
For the agent, we denote the action space and state space respectively as $\mathcal{A}$ and $\mathcal{S}$. 
We also denote the reward function and transition function as $r(s_t, a_t)$ and $f(s_{t+1}|s_t, a_t)$,
where $s_t\in \mathcal{S}$ and $a_t \in \mathcal{A}$ are the state and action at time-step $t$. 
The agent maximizes its expected total reward $J(\pi) = \mathbb{E}_{\pi}[\sum_{t=0}^{\infty}r(s_t, a_t)]$ with respect to the agent's controller $\pi$.
\subsection{Random Shooting Algorithm and PETS}\label{section:bg_pets}
Our proposed algorithm is based on the random shooting algorithm~\cite{richards2005robust}.
In random shooting algorithms~\cite{nagabandi2017neural,chua2018deep}, 
a data-set of $\mathcal{D}=\{(s_t, a_t, s_{t+1})\}$ is collected from previously generated real trajectories.
The agent learns an ensemble of neural networks denoted as $f_\phi(s_{t+1}|s_t, a_t)$,
with the parameters of the neural networks denoted as $\phi$.
In planning,
the agent randomly generates a population of $K$ candidate action sequences.
Each action sequence, denoted as $\mathbf{a} = \{a_0, ..., a_{\tau} \}$,
contains the control signals at every time-steps within the planning horizon $\tau$.
The action sequence with the best expected reward given the current dynamics network $f_\phi(s_{t+1} | s_t, a_t)$ is chosen.
RS, as a model-predictive control algorithm, only executes the first action signal and re-plan at time-step.
In PETS~\cite{chua2018deep},
the authors further use cross entropy method (CEM)~\cite{de2005tutorial,botev2013cross} to re-samples sequences near the best sequences from the last CEM iteration.
\section{\Ourmodel{}}
In this section, we describe two variants of \ourmodelshort{}:
\ourmodelA{} (\textbf{\ourmodelshortA{}}) and \ourmodelP{} (\textbf{\ourmodelshortP{}}).
Following the notations defined in section~\ref{section:bg_pets},
we first define the expected planning reward function at time-step $i$ as follows:
\begin{equation}
    \mathcal{R}(s_i, \mathbf{a}_i) = \mathbb{E}\left[\sum_{t=i}^{i+\tau} r(s_t, a_t)\right], \, \mbox{where\,} s_{t+1} \sim f_\phi(s_{t+1}|s_t, a_t).\\
\end{equation}
The action sequence $\mathbf{a}_i=\{a_i, a_{i+1}, ..., a_{i+\tau}\}$ is generated by the policy search module,
as later described in Section~\ref{section:pocem} and~\ref{section:pwcem}.
The expectation of predicted trajectories $\{s_i, s_{i+1}, ..., s_{i+\tau}\}$ is estimated by creating $P$ particles from the current state.
The dynamics model $f_\phi^{k, t}(s_{t+1}|s_t,a_t)$ used by $k^{th}$ particle at time-step $t$ is sampled from deterministic ensemble models or probabilistic ensemble models.
To better illustrate,
throughout the paper we treat this dynamics as a fixed deterministic model, i.e. $f_\phi^{k,t} \equiv f_\phi$.
This does not violate the math in this paper, and we refer readers to PETS~\cite{chua2018deep} for details.

\subsection{\OurmodelA{}}\label{section:pocem}
In \ourmodelA{} (\ourmodelshortA{}),
we use a policy network to generate good initial action distribution.
We denote the policy network as $\pi(s_t)$. 
Once the policy network proposes sequences of actions on the expected trajectories,
we add Gaussian noise to the candidate actions and use CEM to fine-tune the mean and standard deviation of the noise distribution.

Similar to defining $\mathbf{a}_i=\{a_i, a_{i+1}, ..., a_{i+\tau}\}$,
we denote the noise sequence at time-step $t$ with horizon $\tau$ as $\bm{\delta}_i = \{\delta_i, \delta_{i+1}, ..., \delta_{i+\tau}\}$.
We initialize the noise distribution as a Gaussian distribution with mean $\mu_0 = \bm{0}$ and covariance $\Sigma_0 = \sigma_0^2\bm{I}$,
where $\sigma_0^2$ is the initial noise variance.
In each CEM iteration, we first sort out the sequences with the top $\xi$ expected planning reward, 
whose noise sequences are denoted as $\{\bm{\delta}_i^0, \bm{\delta}_i^1, ..., \bm{\delta}_i^\xi\}$.
Then we estimate the noise distribution of the elite candidates, i. e.,
\begin{equation}
    \Sigma' \gets \mbox{Cov}(\{\bm{\delta}_i^0, \bm{\delta}_i^1, ..., \bm{\delta}_i^\xi\}), \, 
    \mu' \gets \mbox{Mean}(\{\bm{\delta}_i^0, \bm{\delta}_i^1, ..., \bm{\delta}_i^\xi\}).
\end{equation}
The elite distribution ($\mu', \Sigma'$) in CEM algorithm is used to update the candidate noise distribution as $\mu = (1-\alpha)\mu + \alpha\mu',\, \Sigma = (1-\alpha)\Sigma+\alpha\Sigma'$.
For every time-step, several CEM iterations are performed by candidate re-sampling and noise distribution updating.
We provide detailed algorithm boxes in appendix~\ref{appendix:section_diagrams}.
We consider the following two schemes to add action noise.

\textbf{\ourmodelshortA{}-Init:}
In this planning schemes,
we use the policy network only to propose the initialization of the action sequences.
When planning at time-step $i$ with observed state $s_i$,
we first obtain the initial reference action sequences,
denoted as $\mathbf{\hat{a}}_i = \{\hat{a}_i, \hat{a}_{i+1}, ..., \hat{a}_{i+\tau}\}$,
by running the initial forward pass with policy network.
At each planning time-step $t,\,\mbox{where}\, i\le t\le i+\tau$, we have
\begin{equation}\label{equation:init_actions}
    \hat{a}_t = \pi(\hat{s_t}), \,\mbox{where} \, \hat{s}_{t} = f_\phi(\hat{s}_{t-1}, a_{t-1}),\, \hat{s_i} = s_i
\end{equation}
Then the expected reward given search noise $\bm{\delta}_i$ will be:
\begin{equation}
    \mathcal{R}(s_i, \bm{\delta}_i) = \mathbb{E}\left[\sum_{t=i}^{i+\tau} r(s_t, \hat{a}_t + \delta_t)\right], \, \mbox{where\,\,} s_{t+1} = f_\phi(s_{t+1}|s_t, \hat{a}_t + \delta_t).
\end{equation}

\textbf{\ourmodelshortA{}-Replan:}
\ourmodelshortA{}-Replan is a more aggressive planning schemes,
which always re-plans the controller according the changed trajectory given the current noise distribution.
If we had the perfect dynamics network and the policy network,
then we expect re-planning to achieve faster convergence the optimal action distribution.
But it increases the risk of divergent behaviors.
In this case, the expected reward for each trajectory is 
\begin{equation}
    \mathcal{R}(s_i, \bm{\delta}_i) = \mathbb{E}\left[\sum_{t=i}^{i+\tau} r(s_t, \pi(s_t) + \delta_t)\right], \, \mbox{where\,\,} s_{t+1} = f_\phi(s_{t+1}|s_t, \pi(s_t) + \delta_t).
\end{equation}

\subsection{\OurmodelP{}}\label{section:pwcem}
While planning in the action space is a natural extension of the original PETS algorithm,
we found it provides little performance improvement in complex environments.
One potential reason is that \ourmodelshortA{} still performs CEM searching in action sequence space,
where the conditions of convergence for CEM is usually not met.
Let's assume that a robot arm needs to either go left or right to get past the obstacle in the middle.
In CEM planning in the action space, the theoretic distribution mean is always going straight, which fails the task.

Indeed, planning in action space is a non-convex optimization whose surface has lots of holes and peaks.
Recently, much research progress has been made in understanding why deep neural networks are much less likely to get stuck in sub-optimal points~\cite{nguyen2017loss, li2018visualizing, soudry2017exponentially}.
And we believe that planning in parameter space is essentially using deeper neural networks.
Therefore, we propose \ourmodelP{} (\ourmodelshortP{}).
Instead of adding noise in the action space,
\ourmodelshortP{} adds noise in the parameter space of the policy network.
We denote the parameter vector of policy network as $\theta$,
and the parameter noise sequence starting from time-step $i$ as $\bm{\omega}_i = \{\omega_i, \omega_{i+1}, ..., \omega_{i+\tau}\}$.
The expected reward function is now denoted as
\begin{equation}
    \mathcal{R}(s_i, \bm{\omega}_i) = \mathbb{E}\left[\sum_{t=i}^{i+\tau} r\left(s_t, \pi_{\theta + \omega_t}(s_t)\right)\right], \, \mbox{where\,\,} s_{t+1} = f_\phi(s_{t+1}|s_t, \pi_{\theta + \omega_t}(s_t)).
\end{equation}
Similarly, we update the CEM distribution towards the following elite distribution:
\begin{equation}
    \Sigma' \gets \mbox{Cov}(\{\bm{\omega}_i^0, \bm{\omega}_i^1, ..., \bm{\omega}_i^\xi\}), \, 
    \mu' \gets \mbox{Mean}(\{\bm{\omega}_i^0, \bm{\omega}_i^1, ..., \bm{\omega}_i^\xi\}).
\end{equation}
We can force the policy network noise within the sequence to be consistent, i.e. $\omega_i = \omega_{i+1} = ... =\omega_{i+\tau}$, which we name as \textbf{\ourmodelshortP{}-Uni}.
This reduces the size of the flattened noise vector from $(\tau + 1)|\theta|$ to $|\theta|$,
and is more consistent in policy behaviors.
The noise can also be separate for each time-step, which we name as \textbf{\ourmodelshortP{}-Sep}.
We benchmark both schemes in section~\ref{section:ablation_study}.

\textbf{Equivalence to stochastic policy with re-parameterization trick:}
Stochastic policy network encourages exploration,
and increases the robustness against the impact of compounded model errors.
\ourmodelshortP{}, which inserts exogenous noise into the parameter space,
can be regarded as stochastic policy network using re-parameterization trick,
which natural combines stochastic policy network with planning.

\subsection{Model-predictive Control and Policy Control}
MBRL with online re-planning or model-predictive control (MPC) is effective,
but at the same time time-consuming.
Many previous attempts have tried to distill the planned trajectories into a policy network~\cite{levine2014learning, levine2013guided, chebotar2017combining, zhang2018solar},
and control only with policy network.
In this paper, we define two settings of using \ourmodelshort{}:
\textbf{MPC Control} and \textbf{Policy Control}.
In MPC control, the agent uses policy network during the online planning and only execute the first action.
In policy control, the agent directly executes the signal produced by the policy network given current observation,
just like how policy network is used in MFRL algorithms.
We show both performance of \ourmodelshort{} in this paper.
\subsection{Policy Distillation Schemes}
The agents iterate between interacting with the environments,
and distilling the knowledge from planning trajectory into a policy network.
We consider several policy distillation schemes here, and discuss their effectiveness in the later experimental section.

\textbf{Behavior cloning (BC)}: BC can be applied both to \ourmodelshortA{} and \ourmodelshortP{},
where we minimize the squared L2 loss as
\begin{equation}
    \min_{\theta} \, \mathbb{E}_{s,\,a\in \mathcal{D}}||\pi_{\theta}(s) - a||^2,
\end{equation}
where $\mathcal{D}$ is the collection of observation and planned action from real environment.
When applying BC to \ourmodelshortP{}, we fix the parameter noise of the target network to be zeros.

\textbf{Generative adversarial network training (GAN)}~\cite{goodfellow2014generative}:
GAN can be applied to \ourmodelshortP{}.
We consider the following fact.
During MPC control, 
the agent only needs to cover the best action sequence in its action sequence distribution.
Therefore, instead of point-to-point supervised training such as BC,
we can train the policy network using GAN:
\begin{equation}
    \min_{\pi_\theta}\max_{\psi} \, \mathbb{E}_{s,\,a\in \mathcal{D}}\log(D_{\psi}(s, a)) +
    \mathbb{E}_{s\in \mathcal{D}, \,z\sim\mathcal{N}(\bm{0}, \sigma_0\bm{I})}\log(1 - D_{\psi}(s, \pi_{\theta + z}(s))),
\end{equation}
where a discriminator $D$ parameterized by $\psi$ is used,
and we sample the random noise $z$ from the initial CEM distribution $\mathcal{N}(\bm{0}, \sigma_0\bm{I})$.

\textbf{Setting parameter average (AVG)}:
AVG is also applicable to \ourmodelshortP{}.
During interaction with real environment, we also record the optimized parameter noise in to the data-set,
i. e. $\mathcal{D} = \{(s, \omega)\}$.
And we sacrifice the effectiveness of the policy control and only use policy network as a good search initialization.
The new parameter is updated as
\begin{equation}
    \theta = \theta + \frac{1}{|\mathcal{D}|}\sum_{\omega \in \mathcal{D}}\omega.
\end{equation}

\section{Experiments}
\vspace{-0.2cm}
In section~\ref{section:benchmarking_performance},
we compare \ourmodelshort{} with existing algorithms.
We also show the policy control performance of \ourmodelshort{} with different training methods in section~\ref{section:policy_test}.
In section~\ref{section:reward_surface}, we provide explanations and analysis for the effectiveness of our proposed algorithms by exploring and visualizing the reward optimization surface of the planner.
In section~\ref{section:ablation_study},
we study the sensitivity of our algorithms with respect to hyper-parameters, and show the performance of different algorithm variants.
\subsection{MuJoCo Benchmarking Performance}\label{section:benchmarking_performance}
\begin{figure}[!t]
    \centering
    \begin{subfigure}{.24\textwidth}
        \centering
        \includegraphics[width=\linewidth]{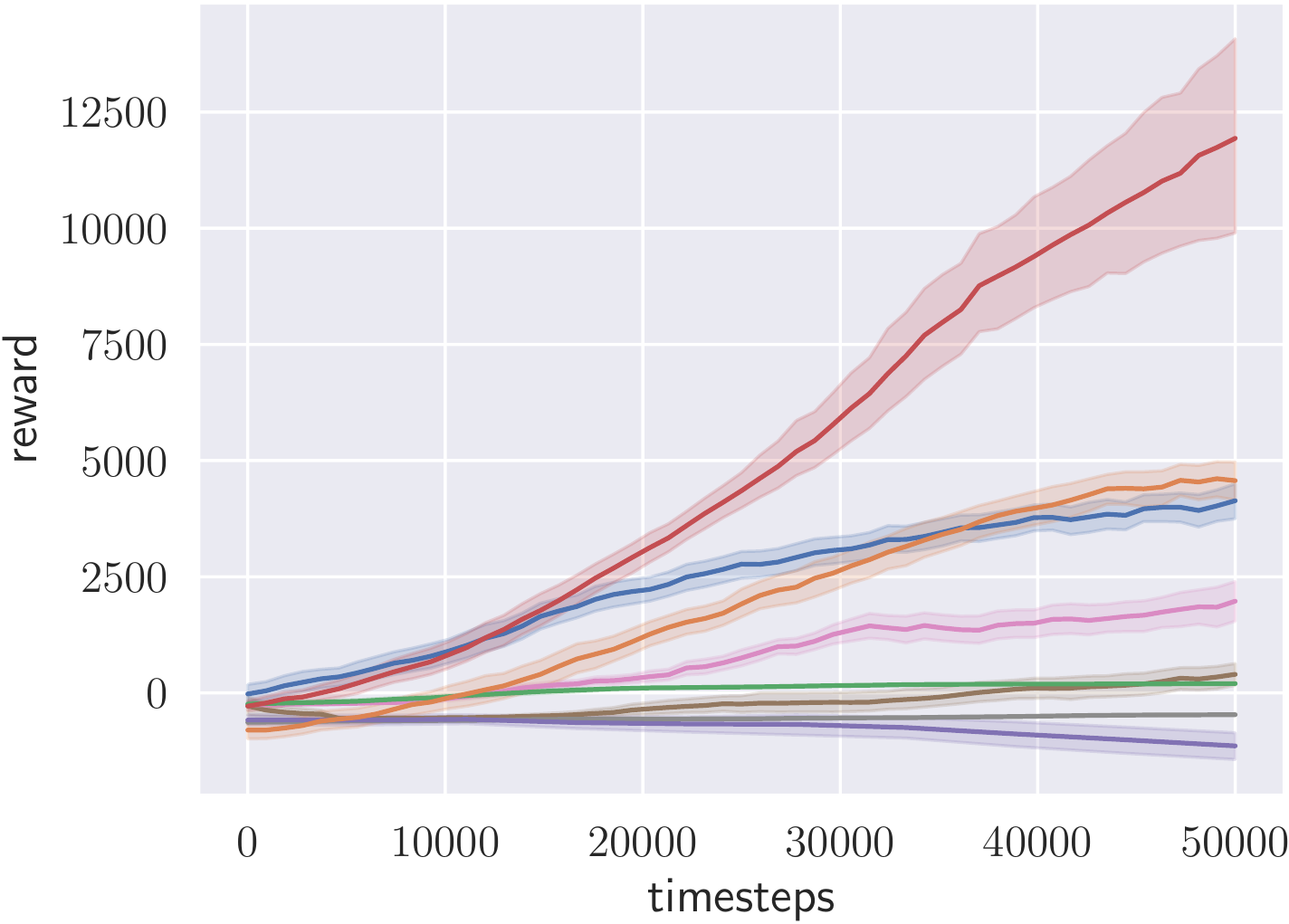}
        (a) Cheetah
    \end{subfigure}%
    \begin{subfigure}{.24\textwidth}
        \centering
        \includegraphics[width=\linewidth]{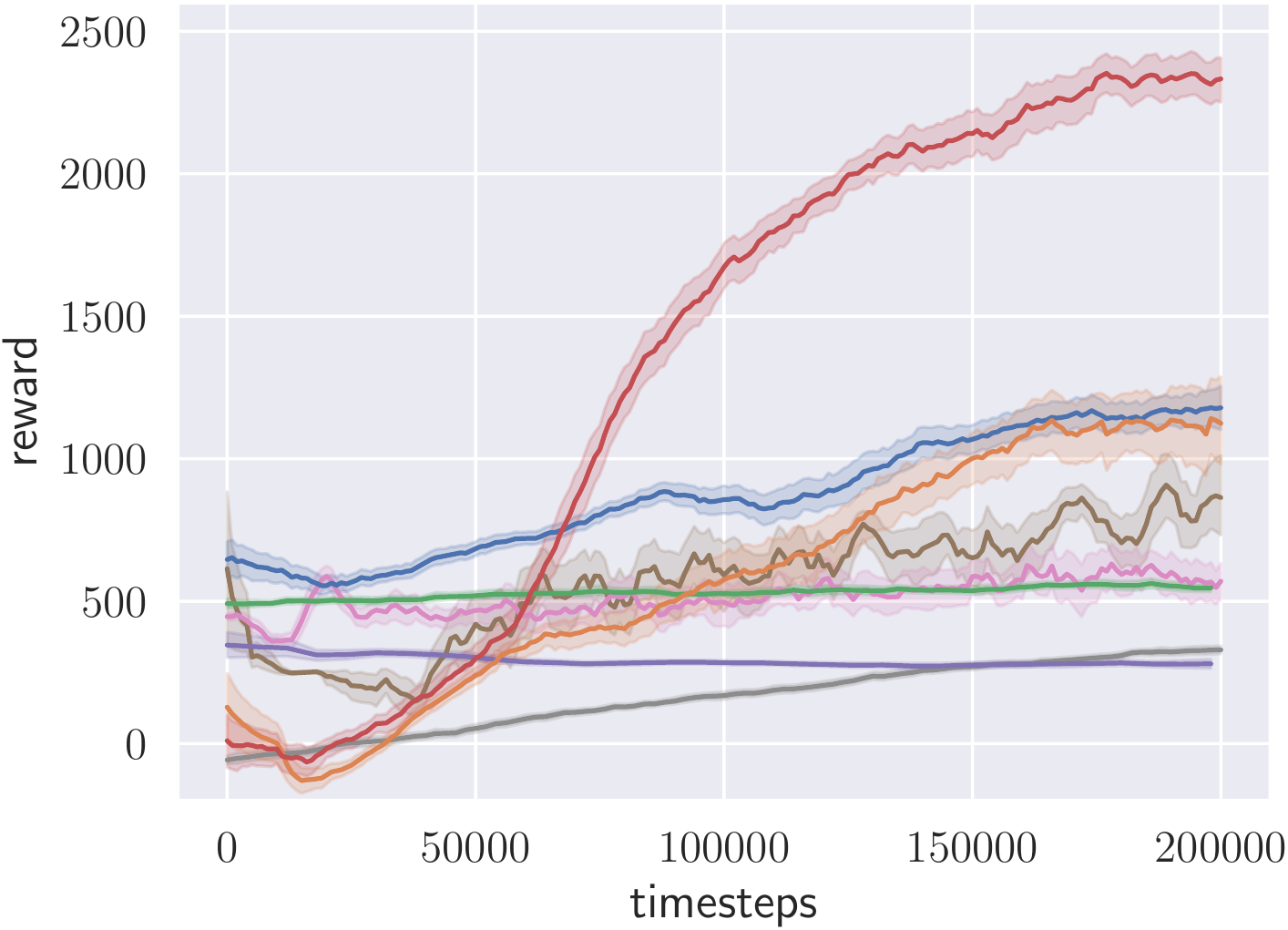}
        (b) Ant
    \end{subfigure}%
    \begin{subfigure}{.24\textwidth}
        \centering
        \includegraphics[width=\linewidth]{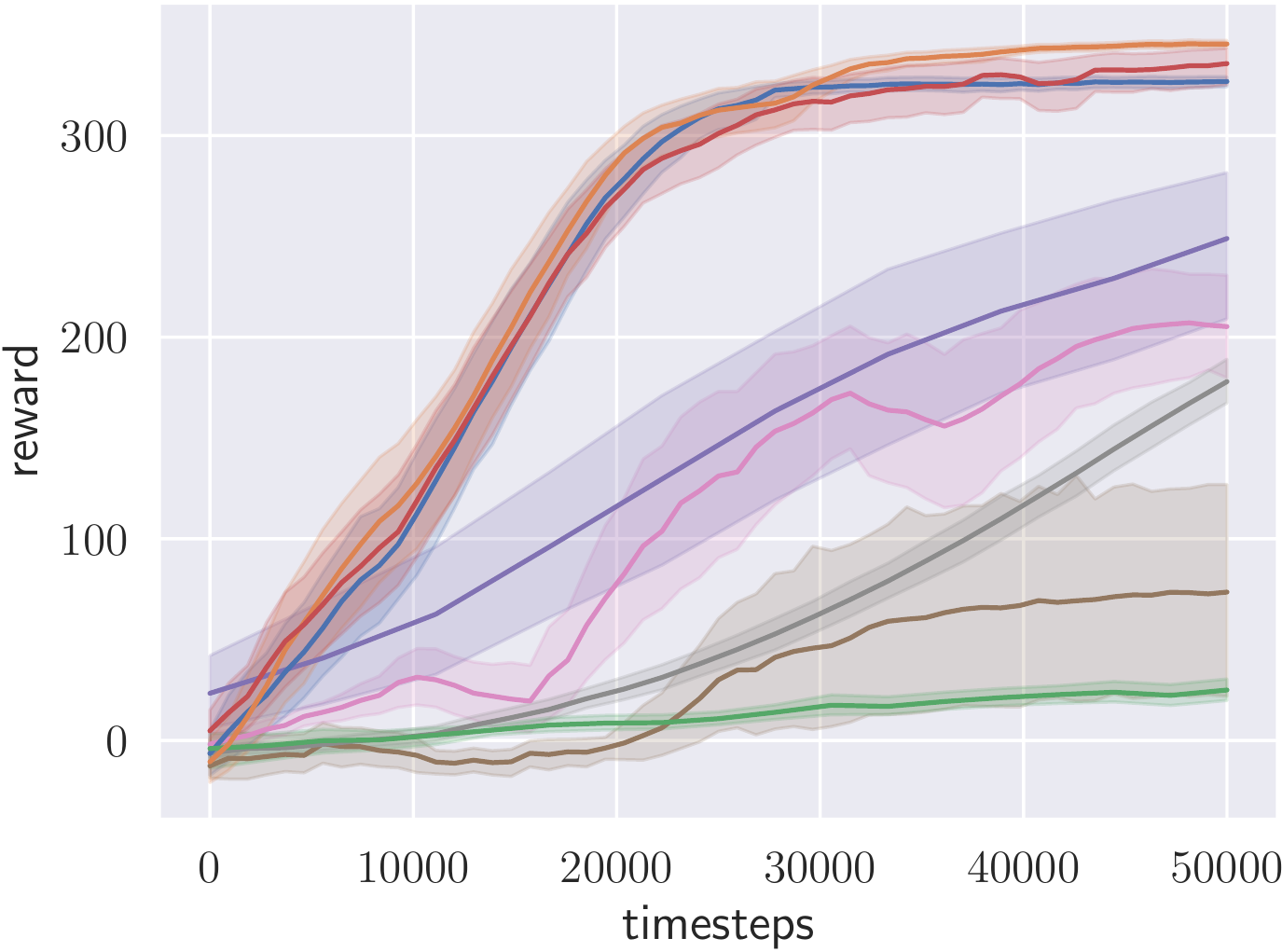}
        (c) Swimmer
    \end{subfigure}%
    \begin{subfigure}{.24\textwidth}
        \centering
        \includegraphics[width=\linewidth]{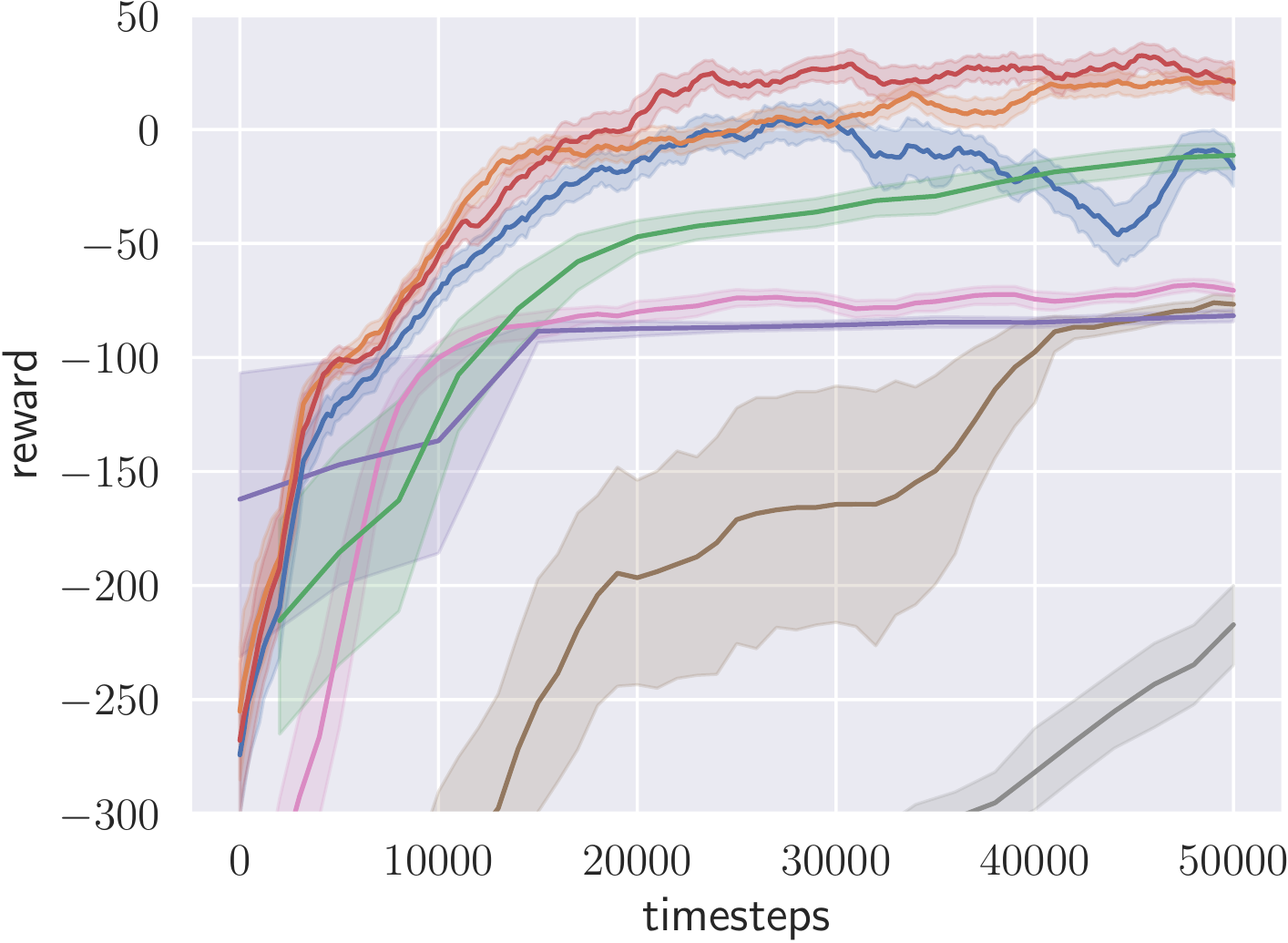}
        (d) Acrobot
    \end{subfigure}
    \begin{subfigure}{1\textwidth}
        \centering
        \includegraphics[width=\linewidth]{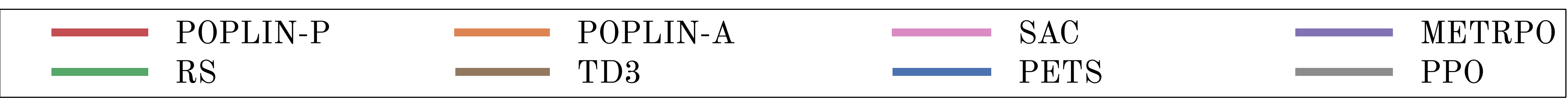}
    \end{subfigure}
    \caption{
    %
    %
    %
    %
    %
    Performance curves of \ourmodelshortP{}, \ourmodelshortA{} and other state-of-the-art algorithms on different bench-marking environments. 4 random seeds are run for each environment, and the full figures of all 12 MuJoCo environments are summarized in appendix~\ref{fig:full_benchmarking_performance}.} 
    \label{fig:benchmarking_performance}
\end{figure}
\vspace{-0.1cm}
\begin{table}[!t]
\resizebox{1\textwidth}{!}{\begin{tabular}{l|c|c|c|c|c|c}
\toprule
& Cheetah  & Ant            & Hopper       & Swimmer      & Cheetah-v0   & Walker2d        \\
\midrule
POPLIN-P (ours) & \textbf{12227.9 $\pm$ 5652.8} & \textbf{2330.1 $\pm$ 320.9} & \textbf{2055.2 $\pm$ 613.8} & \textbf{334.4 $\pm$ 34.2}   & \textbf{4235.0 $\pm$ 1133.0} & \textbf{597.0 $\pm$ 478.8}   \\
POPLIN-A (ours) & 4651.1 $\pm$ 1088.5  & 1148.4 $\pm$ 438.3 & 202.5 $\pm$ 962.5  & \textbf{344.9 $\pm$ 7.1}    & 1562.8 $\pm$ 1136.7 & -105.0 $\pm$ 249.8  \\
PETS~\cite{chua2018deep}    & 4204.5 $\pm$ 789.0   & 1165.5 $\pm$ 226.9 & 114.9 $\pm$ 621.0  & 326.2 $\pm$ 12.6   & 2288.4 $\pm$ 1019.0 & 282.5 $\pm$ 501.6   \\
METRPO~\cite{kurutach2018model}   & -744.8 $\pm$ 707.1   & 282.2 $\pm$ 18.0   & 1272.5 $\pm$ 500.9 & 225.5 $\pm$ 104.6  & 2283.7 $\pm$ 900.4  & -1609.3 $\pm$ 657.5 \\
TD3~\cite{fujimoto2018addressing}      & 218.9 $\pm$ 593.3    & 870.1 $\pm$ 283.8  & 1816.6 $\pm$ 994.8 & 72.1 $\pm$ 130.9   & 3015.7 $\pm$ 969.8  & -516.4 $\pm$ 812.2  \\
SAC~\cite{haarnoja2018soft}      & 1745.9 $\pm$ 839.2   & 548.1 $\pm$ 146.6  & 788.3 $\pm$ 738.2  & 204.6 $\pm$ 69.3   & 3459.8 $\pm$ 1326.6 & 164.5 $\pm$ 1318.6  \\\midrule
Training Time-step & 50000 & 200000 & 200000 & 50000 & 200000 & 200000 \\
\midrule
         & Reacher3D      & Pusher       & Pendulum     & InvertedPendulum  & Acrobot       & Cartpole      \\
\midrule
POPLIN-P (ours) & \textbf{-29.0} $\pm$ 25.2     & \textbf{-55.8 $\pm$ 23.1}   & 167.9 $\pm$ 45.9   & \textbf{-0.0 $\pm$ 0.0}     & \textbf{23.2 $\pm$ 27.2}     & \textbf{200.8 $\pm$ 0.3}     \\
POPLIN-A (ours) & \textbf{-27.7} $\pm$ 25.2     & \textbf{-56.0 $\pm$ 24.3}   & \textbf{178.3 $\pm$ 19.3}   & \textbf{-0.0 $\pm$ 0.0}     & 20.5 $\pm$ 20.1     & \textbf{200.6 $\pm$ 1.3}     \\
PETS~\cite{chua2018deep}     & -47.7 $\pm$ 43.6     & \textbf{-52.7 $\pm$ 23.5}   & 155.7 $\pm$ 79.3   & -29.5 $\pm$ 37.8   & -18.4 $\pm$ 46.3    & 199.6 $\pm$ 4.6     \\
METRPO~\cite{kurutach2018model}   & -43.5 $\pm$ 3.7      & -98.5 $\pm$ 12.6   & 174.8 $\pm$ 6.2    & -29.3 $\pm$ 29.5   & -78.7 $\pm$ 5.0     & 138.5 $\pm$ 63.2    \\
TD3~\cite{fujimoto2018addressing}      & -331.6 $\pm$ 134.6   & -216.4 $\pm$ 39.6  & 168.6 $\pm$ 12.7   & -102.9 $\pm$ 101.0 & -76.5 $\pm$ 10.2    & -409.2 $\pm$ 928.8  \\
SAC~\cite{haarnoja2018soft}      & -161.6 $\pm$ 43.7    & -227.6 $\pm$ 42.2  & 159.5 $\pm$ 12.1   & -0.2 $\pm$ 0.1     & -69.4 $\pm$ 7.0     & 195.5 $\pm$ 8.7    \\\midrule
Training Time-step & 50000 & 50000 & 50000 & 50000 & 50000 & 50000 \\
\bottomrule
\end{tabular}}
\vspace{0.1cm}
\caption{The training time-step varies from 50,000 to 200,000 depending on the difficulty of the tasks. The performance is averaged across four random seeds with a window size of 3000 time-steps at the end of the training.}\label{table:benchmarking_stats}
\vspace{-0.5cm}
\end{table}

In this section,
we compare \ourmodelshort{} with existing reinforcement learning algorithms including PETS~\cite{chua2018deep},
GPS~\cite{levine2016end},
RS~\cite{richards2005robust},
MBMF~\cite{nagabandi2017neural},
TD3~\cite{fujimoto2018addressing}
METRPO~\cite{kurutach2018model},
PPO~\cite{ppo, deepmindppo},
TRPO~\cite{trpo} and SAC~\cite{haarnoja2018soft},
which includes the most recent progress of both model-free and model-based algorithms.
We examine the algorithms with 12 environments,
which is a wide collection of environments from OpenAI Gym~\cite{gym} and the environments proposed in PETS~\cite{chua2018deep},
which are summarized in appendix~\ref{appendix:environments}.
Due to the page limit and to better visualize the results,
we put the complete figures and tables in appendix~\ref{appendix:full_benchmarking}.
And in Figure~\ref{fig:benchmarking_performance} and Table~\ref{table:benchmarking_stats},
we show the performance of our algorithms and the best performing baselines.
The hyper-parameter search is summarized in appendix~\ref{appendix:hyper_search}.

As shown in Table~\ref{table:benchmarking_stats}, \ourmodelshort{} achieves state-of-the-art performance in almost all of the environments,
solving most of the environments with 200,000 or 50,000 time-steps,
which is much less than the 1 million time-steps commonly used in MFRL algorithms.
\ourmodelshortA{} has the best performance in simpler environments such as Pendulum, Cart-pole, Swimmer.
But on complex environments such as Ant, Cheetah or Hopper, \ourmodelshortA{} does not have obvious performance gain compared with PETS.
\ourmodelshortP{} on the other hand,
has consistent and stable performance among different environments.
\ourmodelshortP{} is significantly better than all other algorithms in complex environments such as Ant and Cheetah.
However, like other model-based algorithms,
\ourmodelshort{} cannot efficient solve environments such as Walker and Humanoid.
Although in the given episodes, \ourmodelshort{} has better sample efficiency,
gradually model-free algorithms will have better asymptotic performance.
We view this as a bottleneck of our algorithms and leave it to future research.
\subsection{Policy Control Performance}\label{section:policy_test}
In this section, we show the performance of \ourmodelshort{} without MPC.
To be more specific, we show the performance with the Cheetah, Pendulum, Pusher and Reacher3D,
and we refer readers to appendix~\ref{appendix:full_testing} for the full results.
We note that policy control is not always successful, 
and in environments such as Ant and Walker2D, the performance is almost random.
In simple environments such as Pusher and Reacher3D, \ourmodelshortA{} has the best MPC performance,
but has worse policy control performance compared with \ourmodelshortP{}-BC and \ourmodelshortP{}-GAN.
At the same time, both \ourmodelshortP{}-BC and \ourmodelshortP{}-GAN are able to efficiently distill the knowledge from planned trajectory.
Which one of \ourmodelshortP{}-BC and \ourmodelshortP{}-GAN is better depends on the environment tested,
and they can be used interchangeably.
This indicates that \ourmodelshortA{}, which uses a deterministic policy network,
is more prone to distillation collapse than \ourmodelshortP{},
which can be interpreted as using a stochastic policy network with reparameterization trick.
\ourmodelshortP{}-Avg, which only use policy network as optimization initialization has good MPC performance,
but sacrifices the policy control performance.
In general, the performance of policy control lags behind MPC control.
\begin{figure}[!t]
    \centering
    \begin{subfigure}{.24\textwidth}
        \centering
        \includegraphics[width=\linewidth]{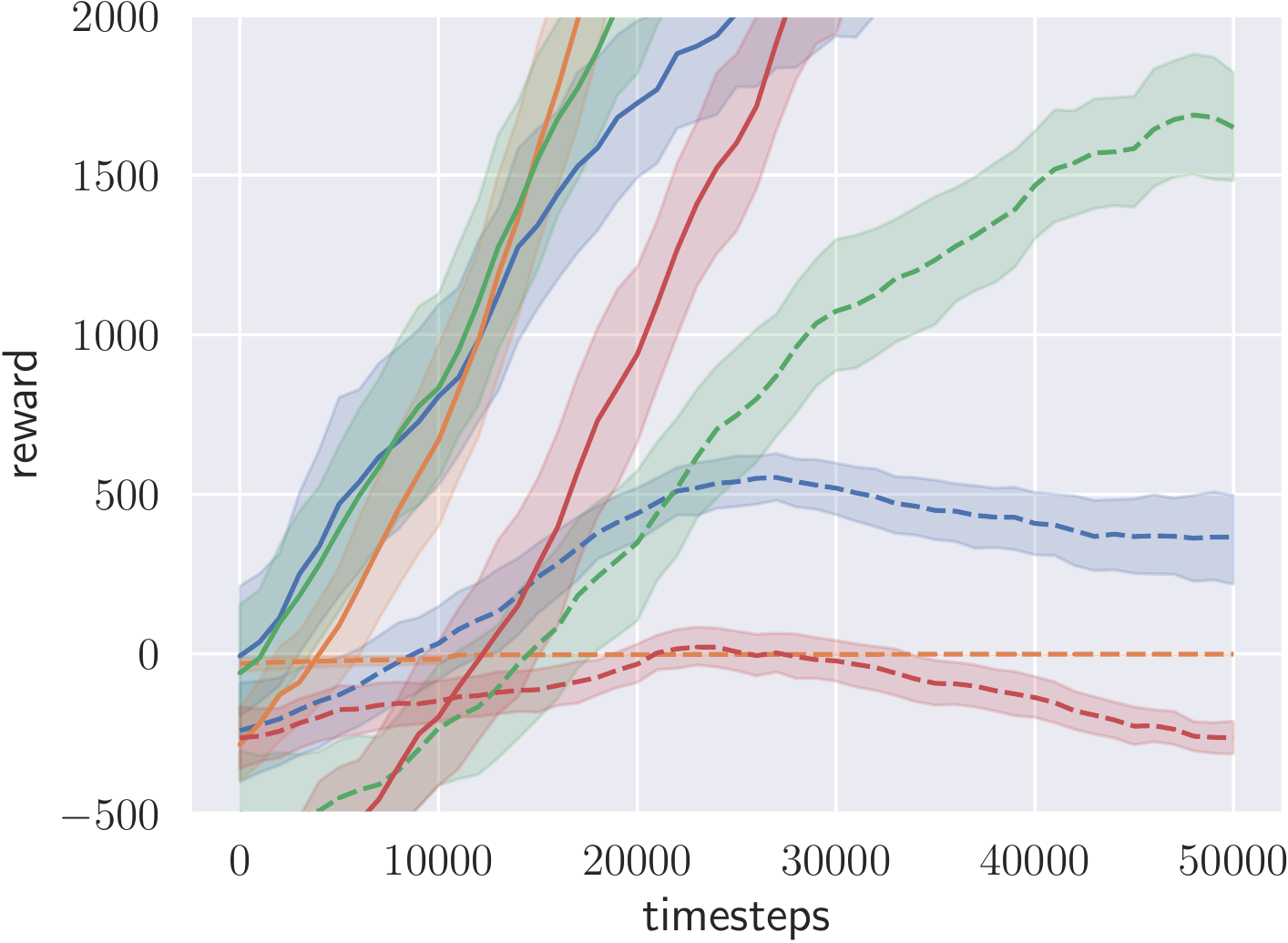}
        (a) Cheetah
    \end{subfigure}
    \begin{subfigure}{.24\textwidth}
        \centering
        \includegraphics[width=\linewidth]{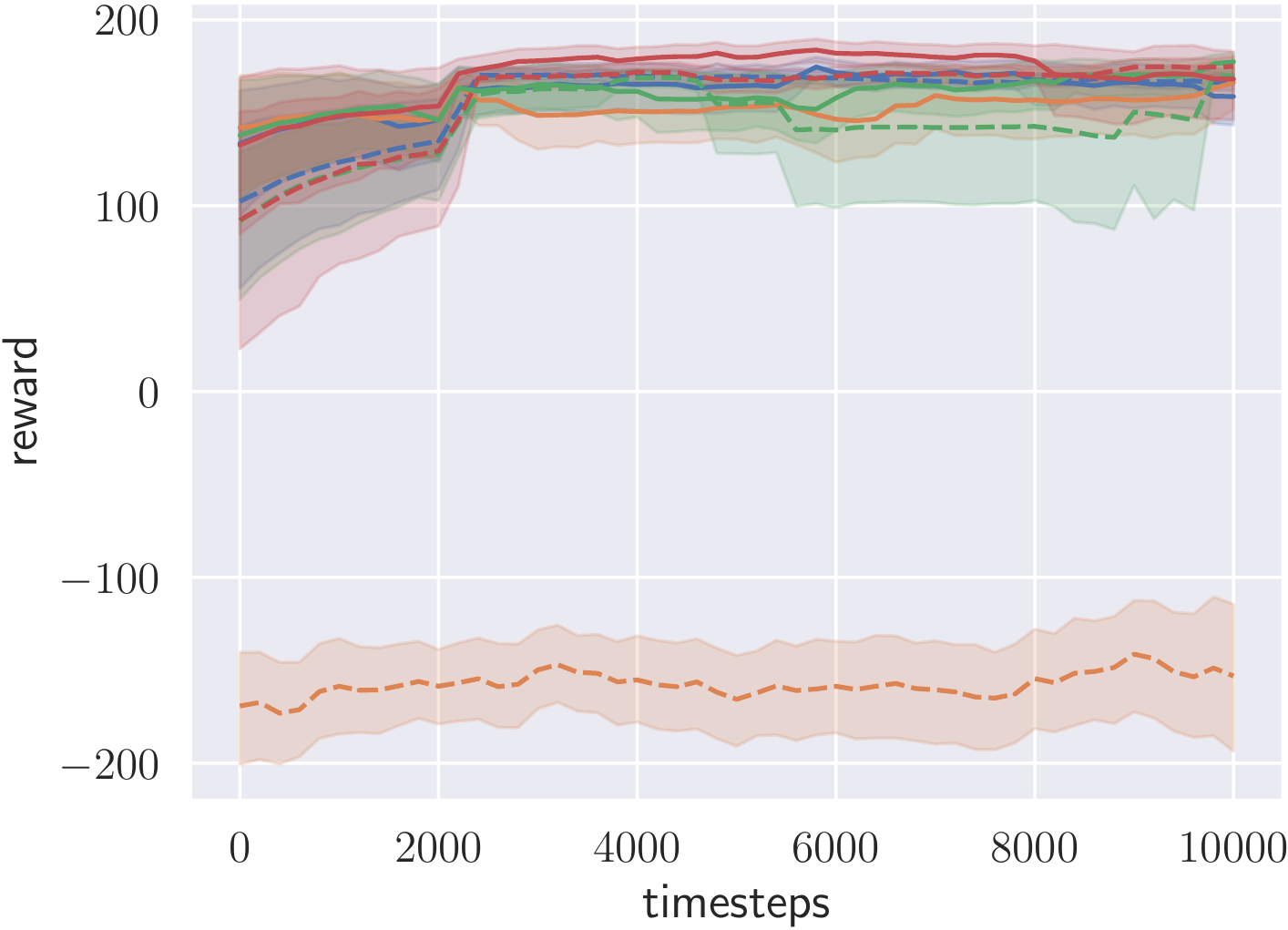}
        (b) Pendulum
    \end{subfigure}
    \begin{subfigure}{.24\textwidth}
        \centering
        \includegraphics[width=\linewidth]{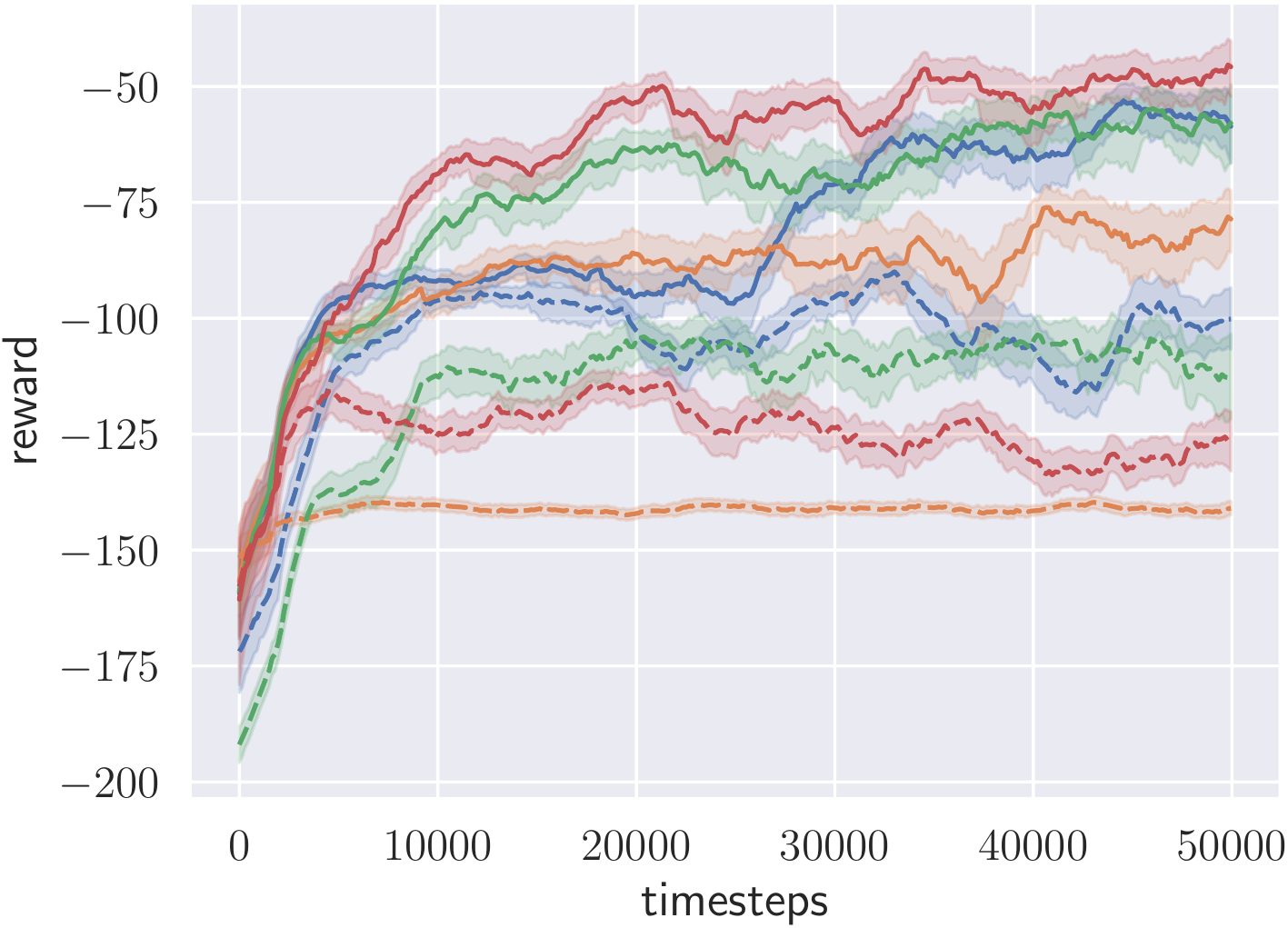}
        (c) Pusher
    \end{subfigure}
    \begin{subfigure}{.24\textwidth}
        \centering
        \includegraphics[width=\linewidth]{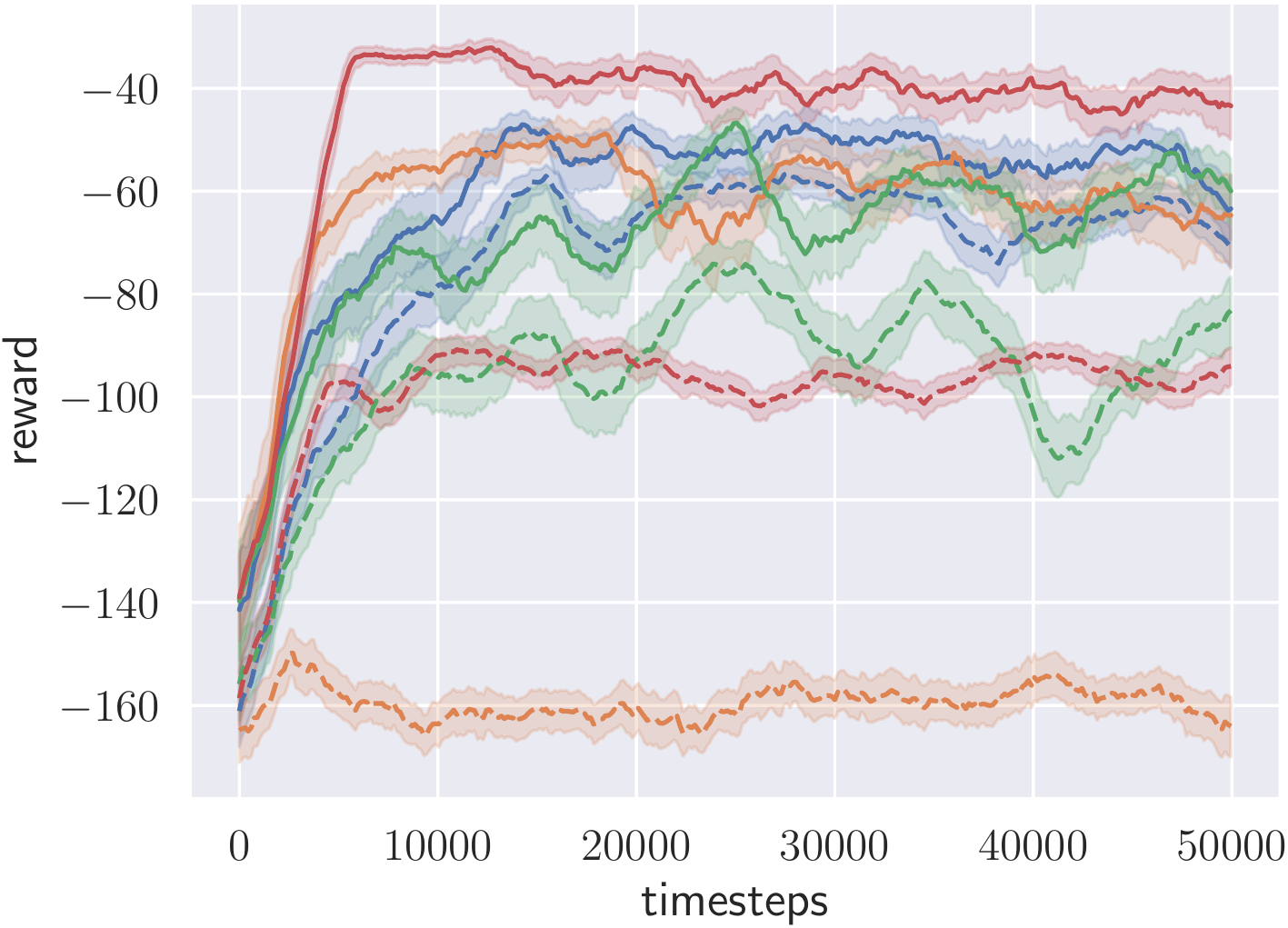}
        (e) Reacher3D
    \end{subfigure}%
    
    \begin{subfigure}{1\textwidth}
        \centering
        \includegraphics[width=\linewidth]{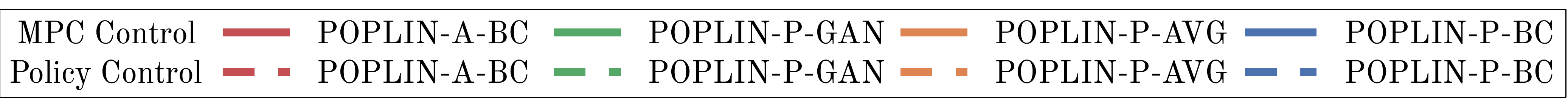}
    \end{subfigure}
    \caption{The MPC control and policy control performance of the proposed \ourmodelshortA{}, and \ourmodelshortP{} with its three training schemes, which are namely behavior cloning (BC), generative adversarial network training (GAN) and setting parameter average (Avg).}
    \label{fig:test_performance}
\end{figure}
\subsection{Search Effectiveness and Reward Surface}\label{section:reward_surface}
\begin{figure}[!t]
    \centering
    \begin{subfigure}{.32\textwidth}
        \centering
        \includegraphics[width=\linewidth]{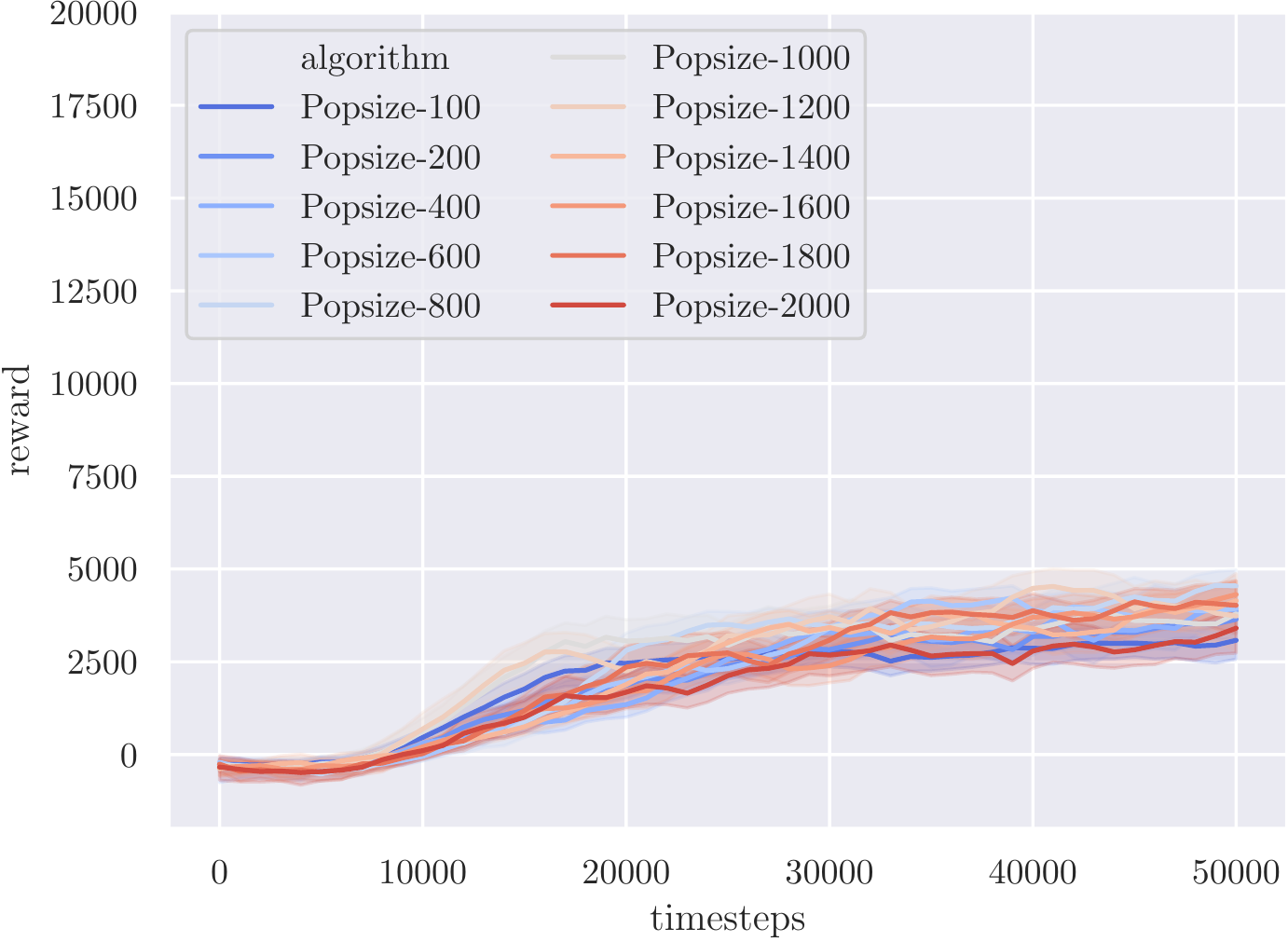}
        (a) PETS
    \end{subfigure}
    \begin{subfigure}{.32\textwidth}
        \centering
        \includegraphics[width=\linewidth]{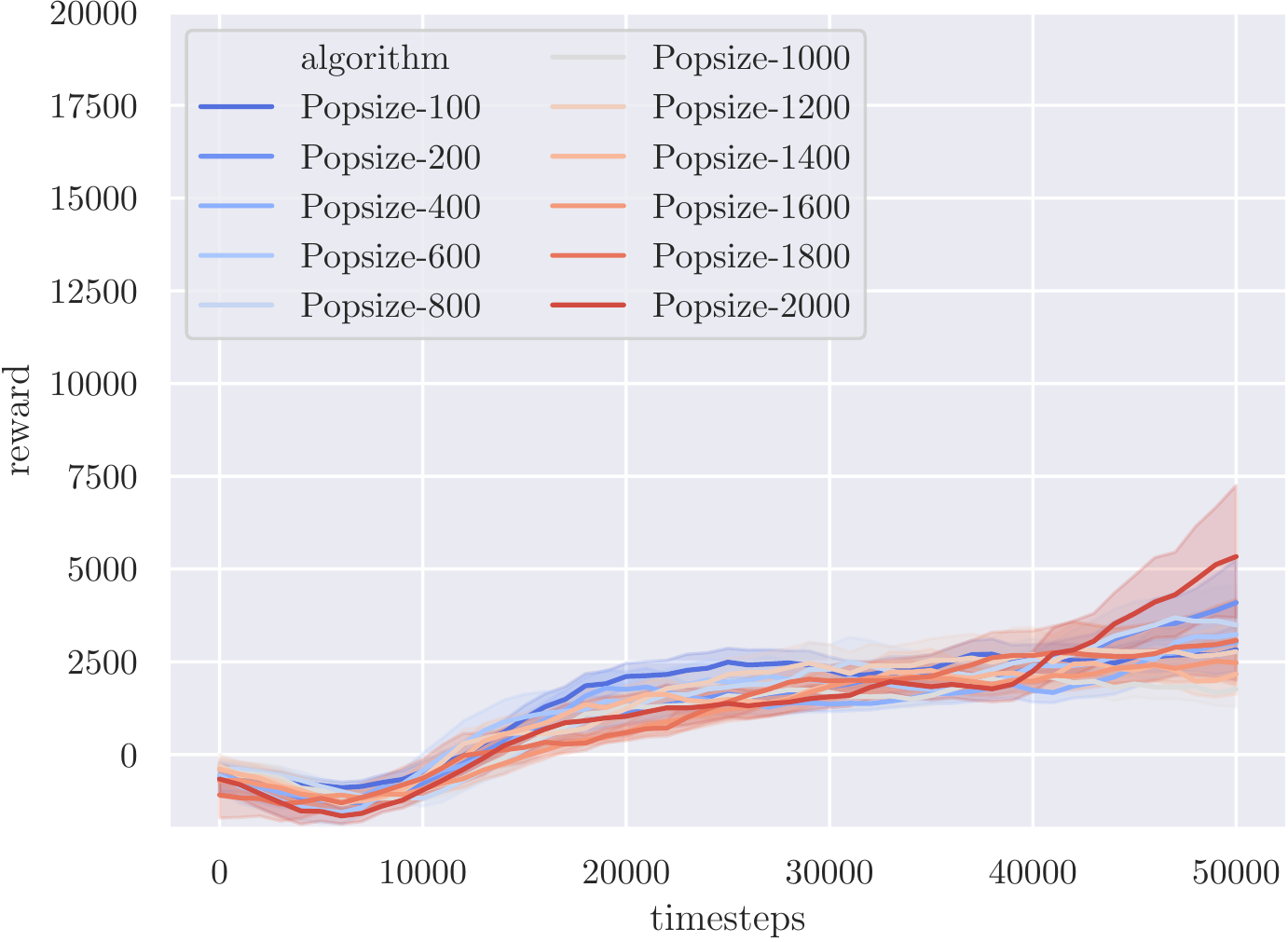}
        (b) \ourmodelshortA{}
    \end{subfigure}
    \begin{subfigure}{.32\textwidth}
        \centering
        \includegraphics[width=\linewidth]{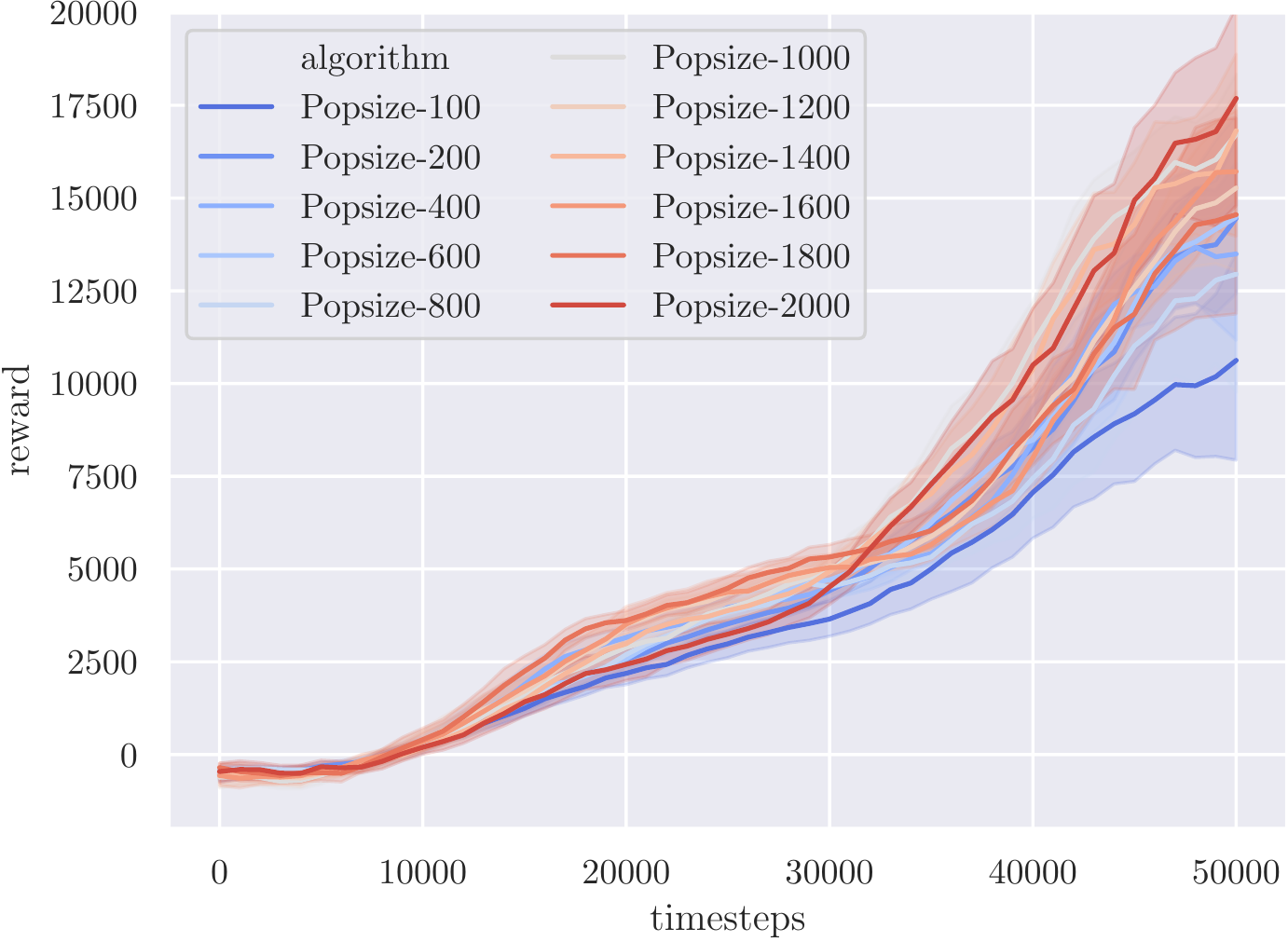}
        (c) \ourmodelshortP{}
    \end{subfigure}%
    \caption{The performance of PETS, \ourmodelshortA{}, \ourmodelshortP{} using different population size of candidates. The variance of the candidates trajectory $\sigma$ in \ourmodelshortP{} is set to 0.1.}
    \label{fig:popsize}
    \vspace{-0.3cm}
\end{figure}
\begin{figure}[!t]
    \centering
    \begin{subfigure}{.15\textwidth}
        \centering
        \includegraphics[width=\linewidth]{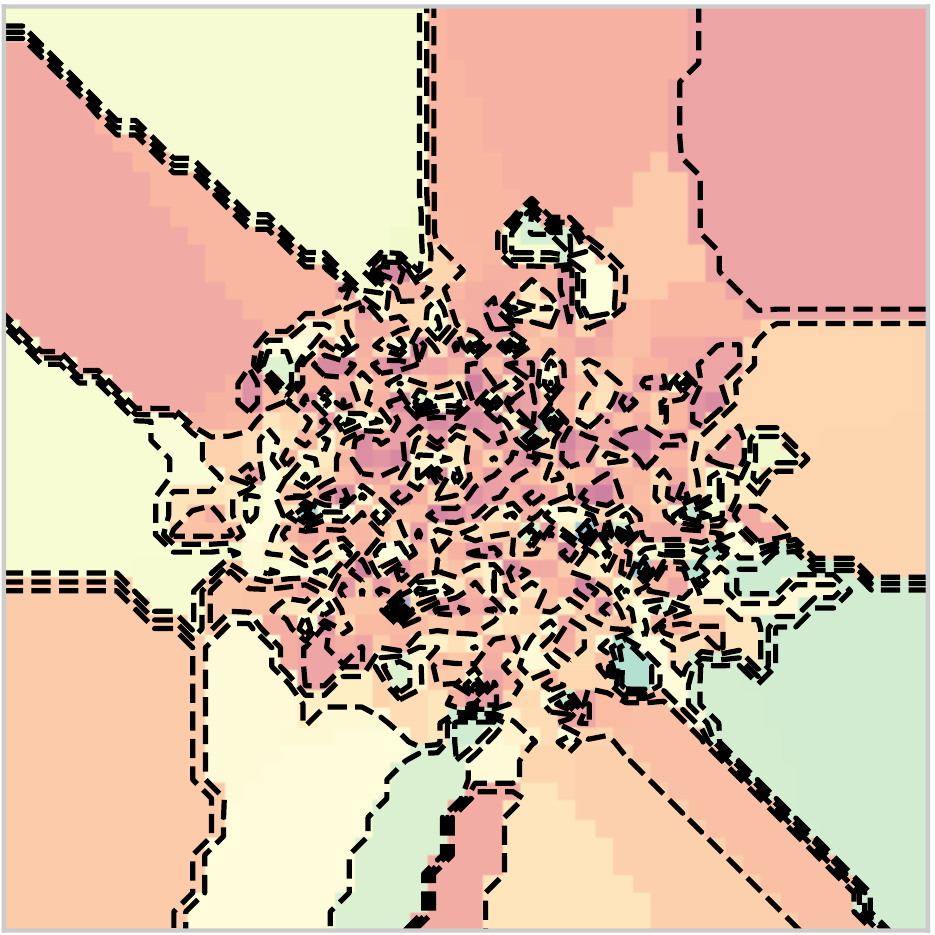}
        (a) PETS Surface
    \end{subfigure}
    \begin{subfigure}{.15\textwidth}
        \centering
        \includegraphics[width=\linewidth]{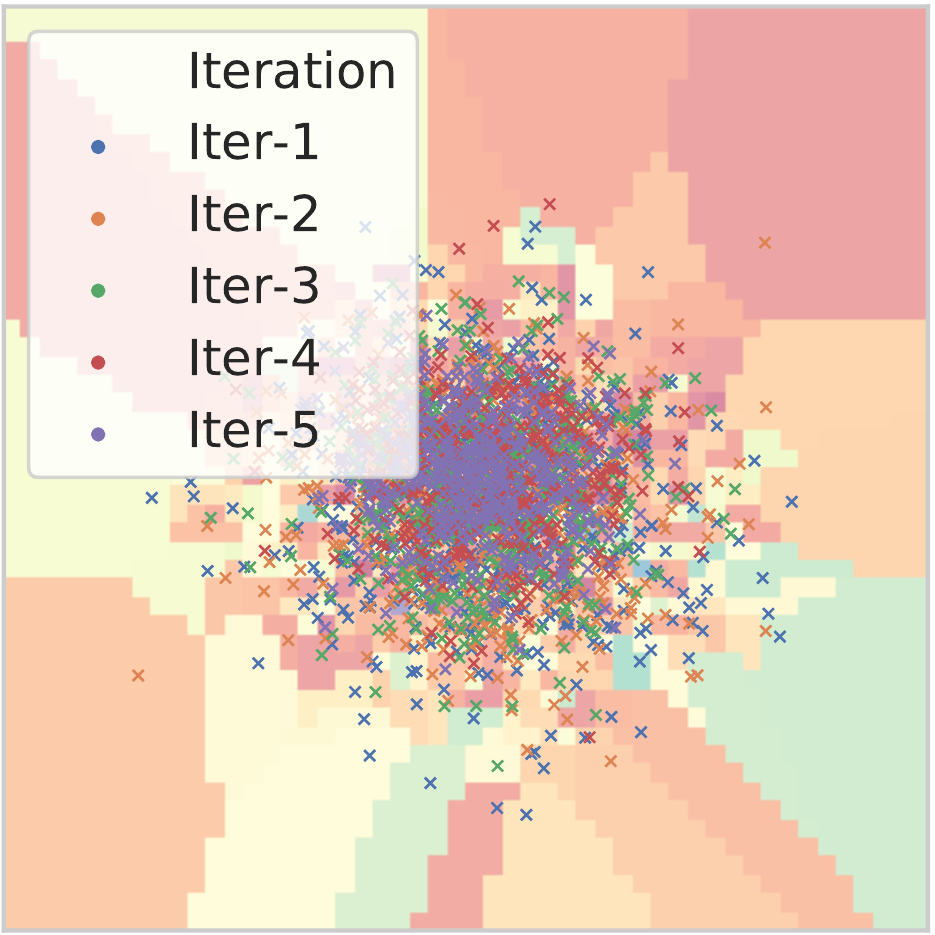}
        (b) PETS Population
    \end{subfigure}
    \begin{subfigure}{.15\textwidth}
        \centering
        \includegraphics[width=\linewidth]{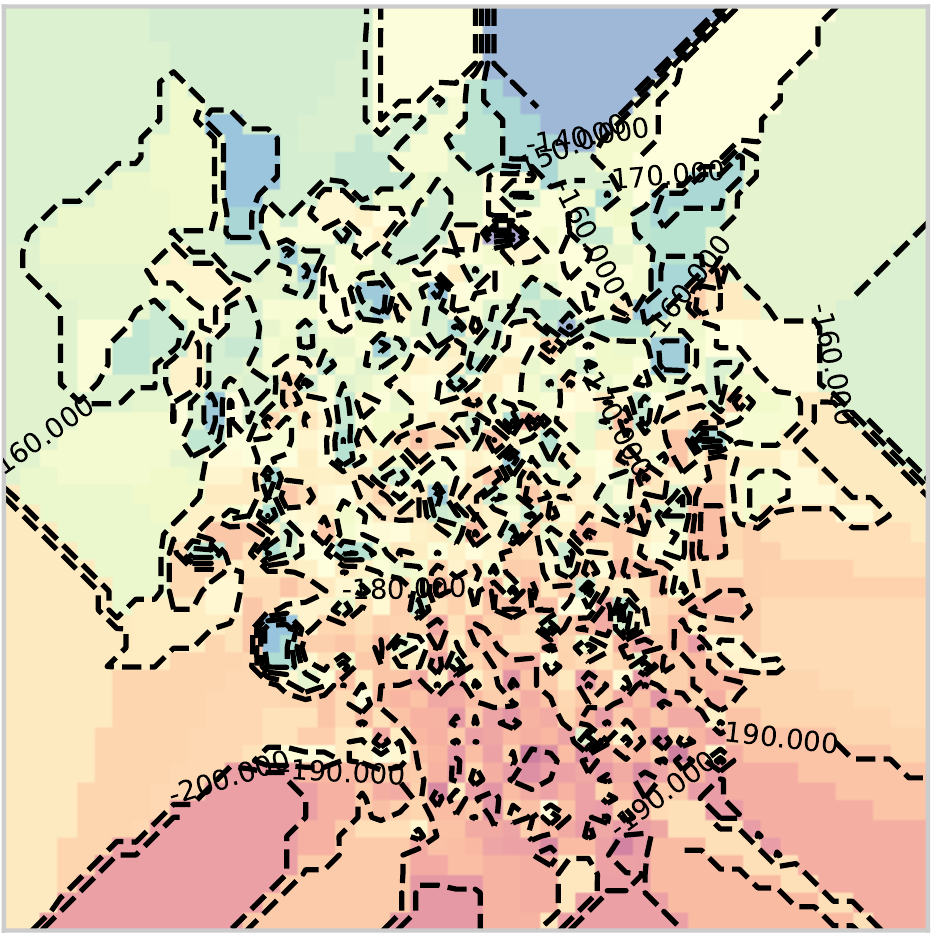}
        (c) \ourmodelshortA{} Surface
    \end{subfigure}
    \begin{subfigure}{.15\textwidth}
        \centering
        \includegraphics[width=\linewidth]{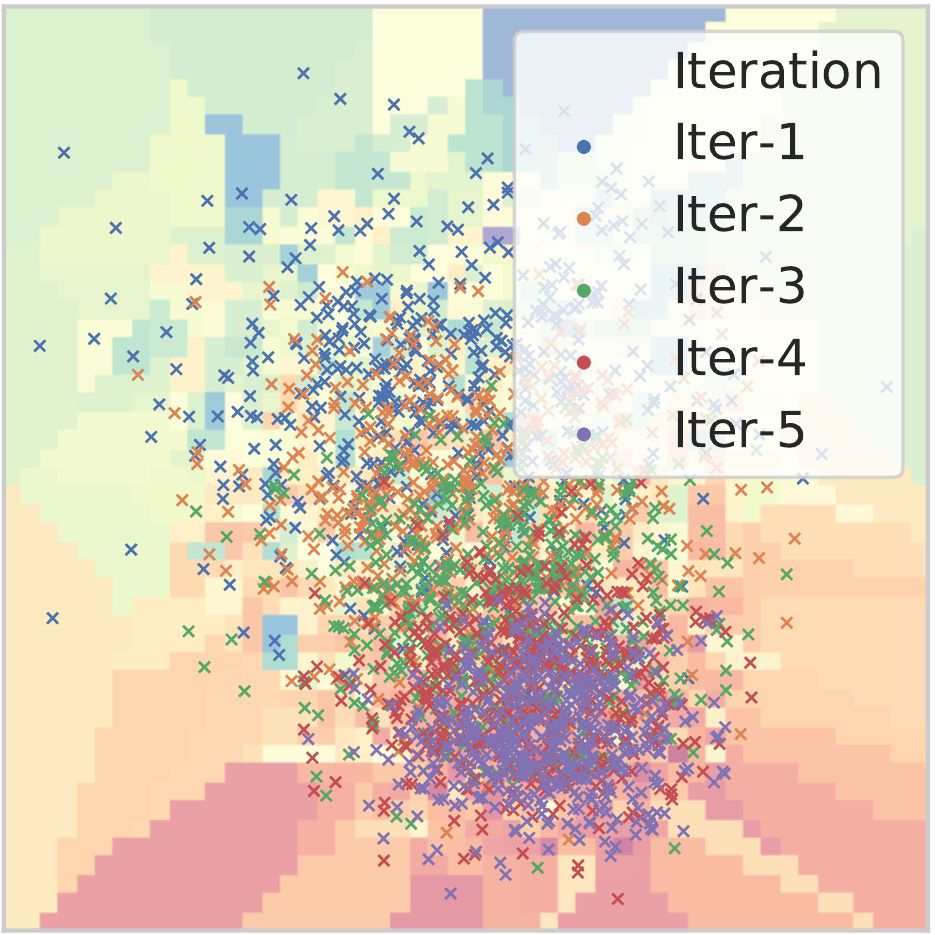}
        (d) \ourmodelshortA{} Population
    \end{subfigure}
    \begin{subfigure}{.15\textwidth}
        \centering
        \includegraphics[width=\linewidth]{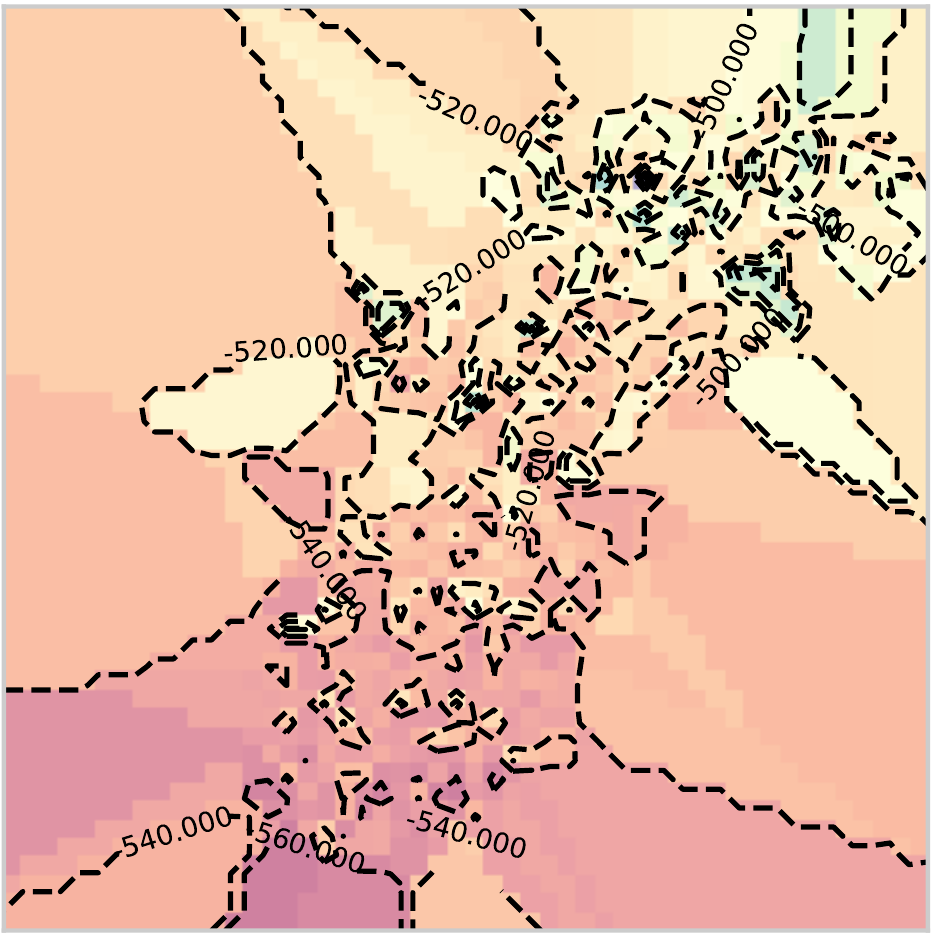}        
        (e) \ourmodelshortP{} Surface
    \end{subfigure}
    \begin{subfigure}{.15\textwidth}
        \centering
        \includegraphics[width=\linewidth]{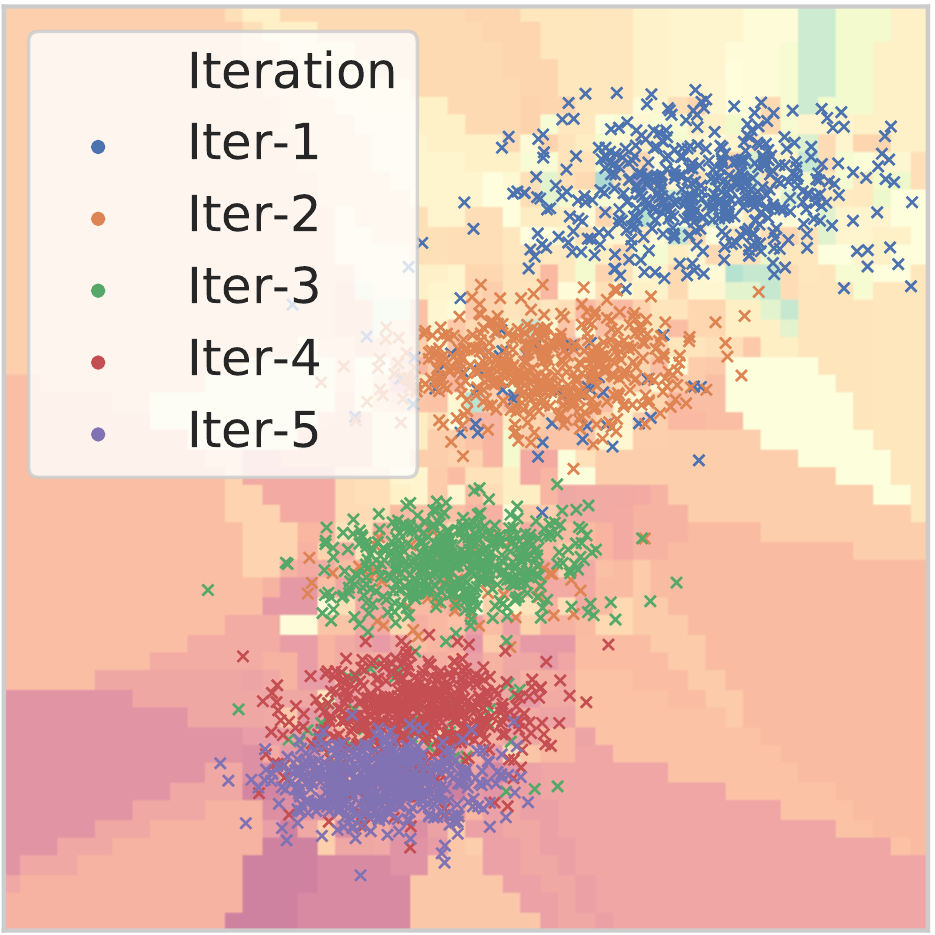}
        (f) \ourmodelshortP{} Population
    \end{subfigure}
    \caption{The reward optimization surface in the solution space. The expected reward is higher from color blue to color red.
    We visualize candidates using different colors as defined in the legend. The full results can be seen in appendix~\ref{appendix:full_reward_surface}.}
    \label{fig:reward_surface_sol_space}
\end{figure}
In this section, we explore the reasons for the effectiveness of \ourmodelshort{}.
In Figure~\ref{fig:popsize}, we show the performance of PETS, \ourmodelshortA{} and \ourmodelshortP{} with different population sizes.
As we can see,
PETS and \ourmodelshortA{},
which are the two algorithms that add search noise in the action space,
cannot increase their performance by having bigger population size.
However, \ourmodelshortP{} is able to efficiently increase performance with bigger population size.
We then visualize the candidates in their reward or optimization surface in Figure~\ref{fig:reward_surface}.
We use PCA (principal component analysis) to transform the action sequences into 2D features.
As we can see, the reward surface is not smooth, with lots of local-minima and local-maxima islands.
The CEM distribution of PETS algorithm is almost fixed across iterations on this surface,
even if there are potentially higher reward regions.
\ourmodelshort{} is able to efficiently search through the jagged reward surface,
from the low-reward center to the high reward left-down corner.
To further understand why \ourmodelshort{} is much better at searching through the reward surface,
we then plot the figures in the solution space in Figure~\ref{fig:reward_surface_sol_space}.
More specifically, we now perform PCA on the policy parameters for \ourmodelshortP{}.
As we can see in Figure~\ref{fig:reward_surface_sol_space} (c),
the reward surface in parameter space is much smoother than the reward surface in action space,
which are shown in Figure~\ref{fig:reward_surface_sol_space} (a), (b).
\ourmodelshortP{} can efficiently search through the smoother reward surface in parameter space.

\begin{wrapfigure}[10]{r}{0.30\textwidth}%
    \vspace{-2mm}
    \centering
    \includegraphics[width=0.30\textwidth]{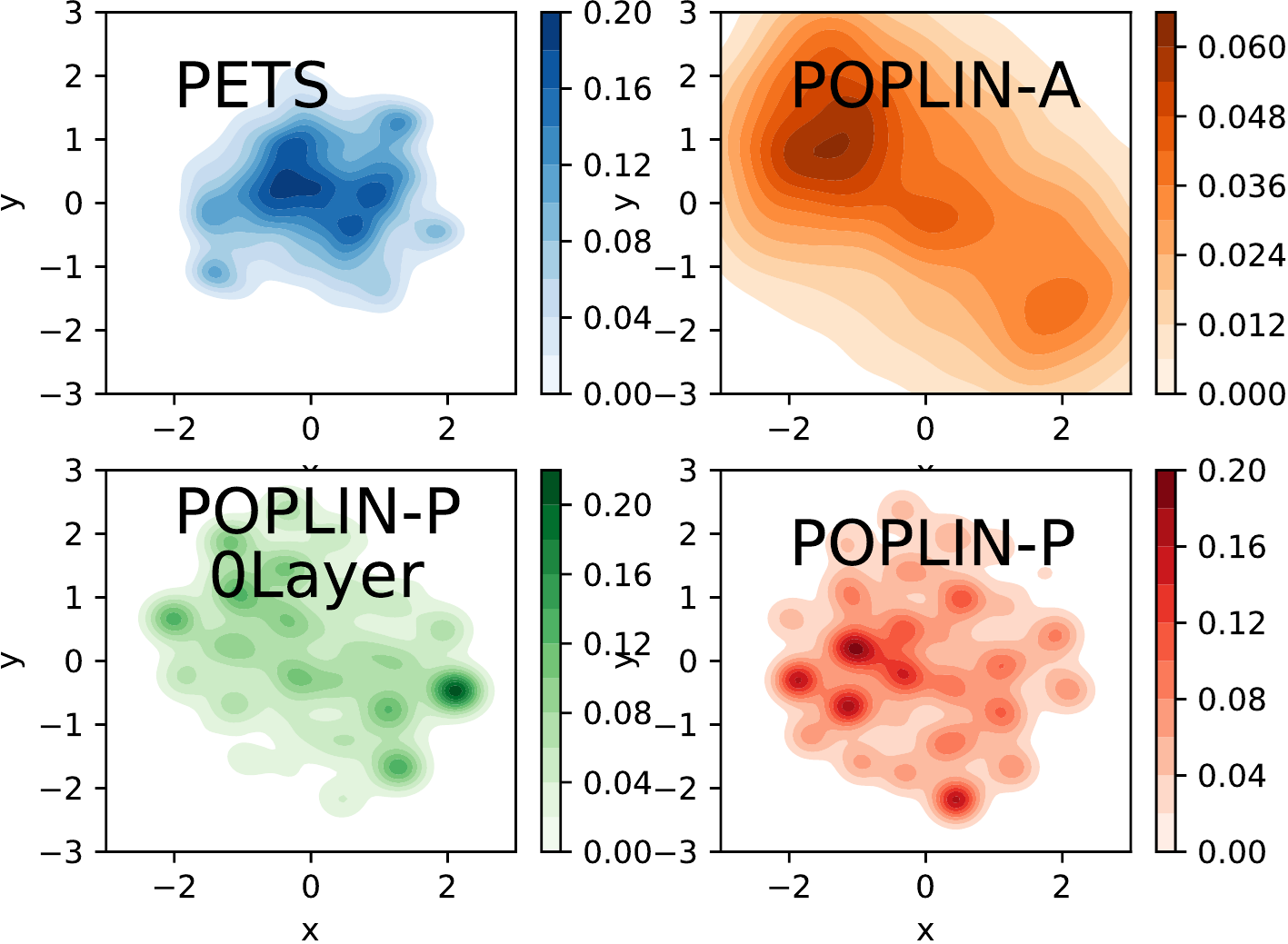}
    \caption{The action distribution in a episode visualized in the projected 2D PCA space. 
    }
    \label{fig:action_distribution}
\end{wrapfigure}

In Figure~\ref{fig:action_distribution}, we also visualize the actions distribution in one episode taken by PETS, \ourmodelshortA{} and \ourmodelshortP{} using policy networks of different number of hidden layers.
We again use PCA to project the actions into 2D feature space.
As we can see,
\ourmodelshortP{} shows a clear pattern of being more multi-model with the use of deeper the network.

\subsection{Ablation Study}\label{section:ablation_study}
In this section,
we study how sensitive our algorithms are with respect to some of the crucial hyper-parameters, for example, the initial variance of the CEM noise distribution.
We also show the performance of different algorithm variants.
The full ablation study and performance against different random seeds are included in appendix~\ref{appendix:ablation_study}.

In Figure~\ref{fig:ablation_study} (a), we show the performance of \ourmodelshortA{} using different training schemes.
We try both training with only the real data samples, which we denote as "Real",
and training also with imaginary data the agent plans into the future, which we denote as "Hallucination".
In practice, \ourmodelshortA{}-Init performs better than \ourmodelshortA{}-Replan,
which suggests that there can be divergent or overconfident update in \ourmodelshortA{}-Replan.
And training with or without imaginary does not have big impact on the performance.
In Figure\ref{fig:ablation_study} (b) and (c),
we also compare the performance of \ourmodelshortP{}-Uni with \ourmodelshortP{}-Sep,
where we show that \ourmodelshortP{}-Sep has much better performance than \ourmodelshortP{}-Uni,
indicating the search is not efficient enough in the constrained parameter space.
For \ourmodelshortP{}-Avg, with bigger initial variance of the noise distribution,
the agent gets better at planning.
However, increasing initial noise variance does not increase the performance of PETS algorithm,
as shown in~\ref{fig:ablation_study} (b), (d). 
It is worth mentioning that \ourmodelshortP{}-GAN is highly sensitive to the entropy penalty we add to the discriminator,
with the 3 curves in Figure\ref{fig:ablation_study} (c) using entropy penalty of 0.003, 0.001 and 0.0001 respectively,

\begin{figure}[!t]
    \centering
    \begin{subfigure}{.24\textwidth}
        \centering
        \includegraphics[width=\linewidth]{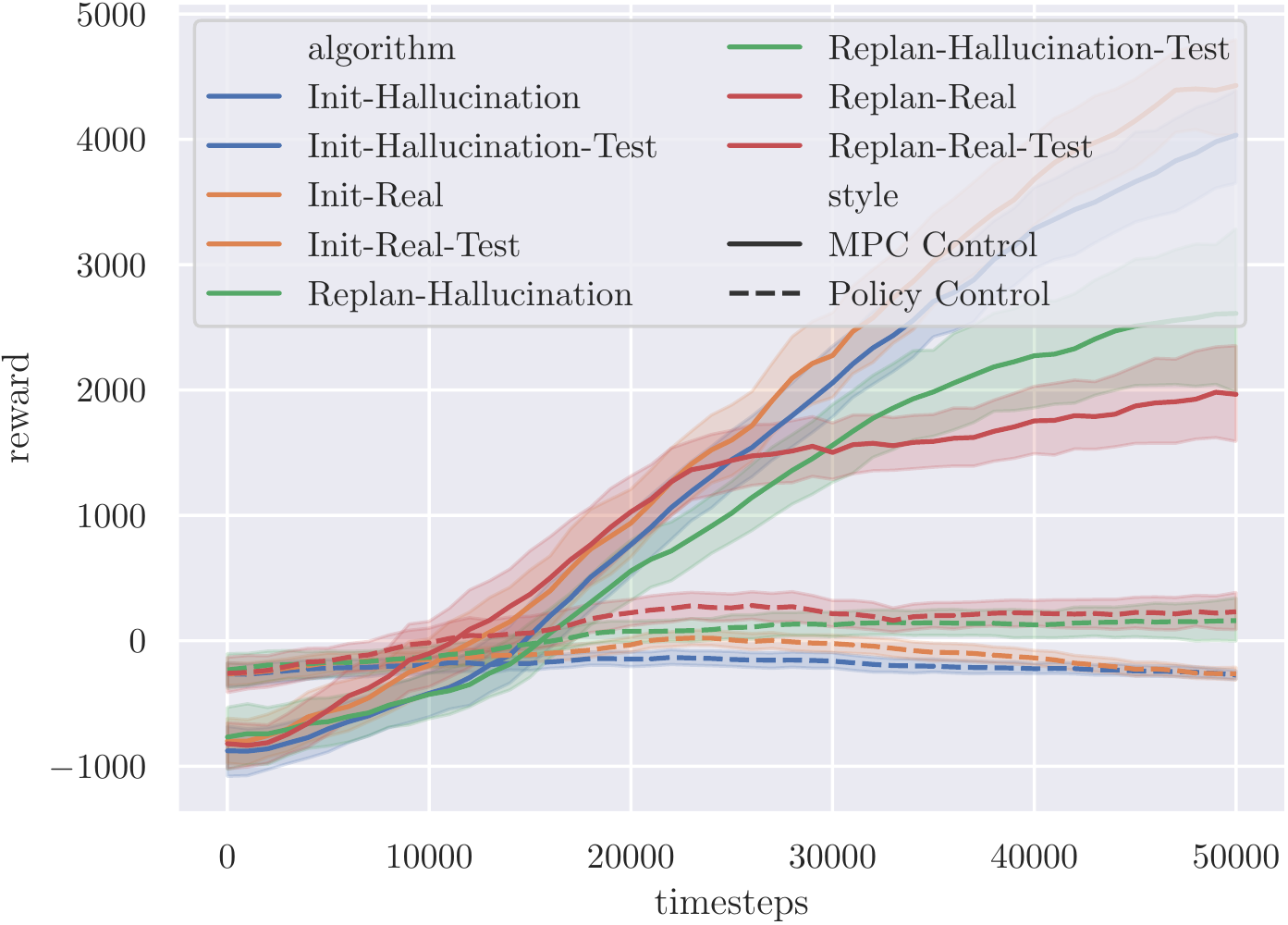}
        (a) \ourmodelshortA{}
    \end{subfigure}
    \begin{subfigure}{.24\textwidth}
        \centering
        \includegraphics[width=\linewidth]{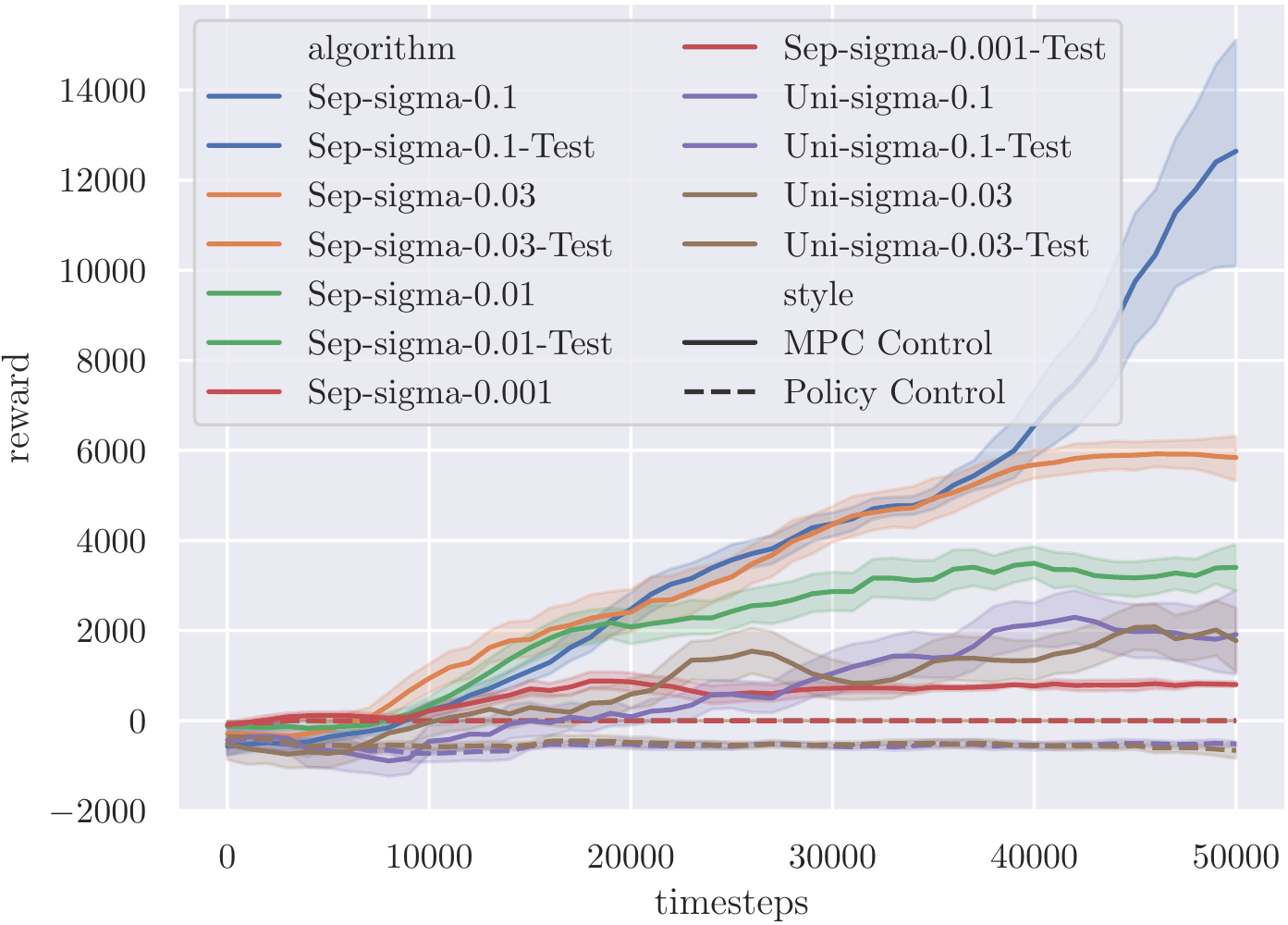}
        (b) \ourmodelshortP{}-Avg
    \end{subfigure}
    \begin{subfigure}{.24\textwidth}
        \centering
        \includegraphics[width=\linewidth]{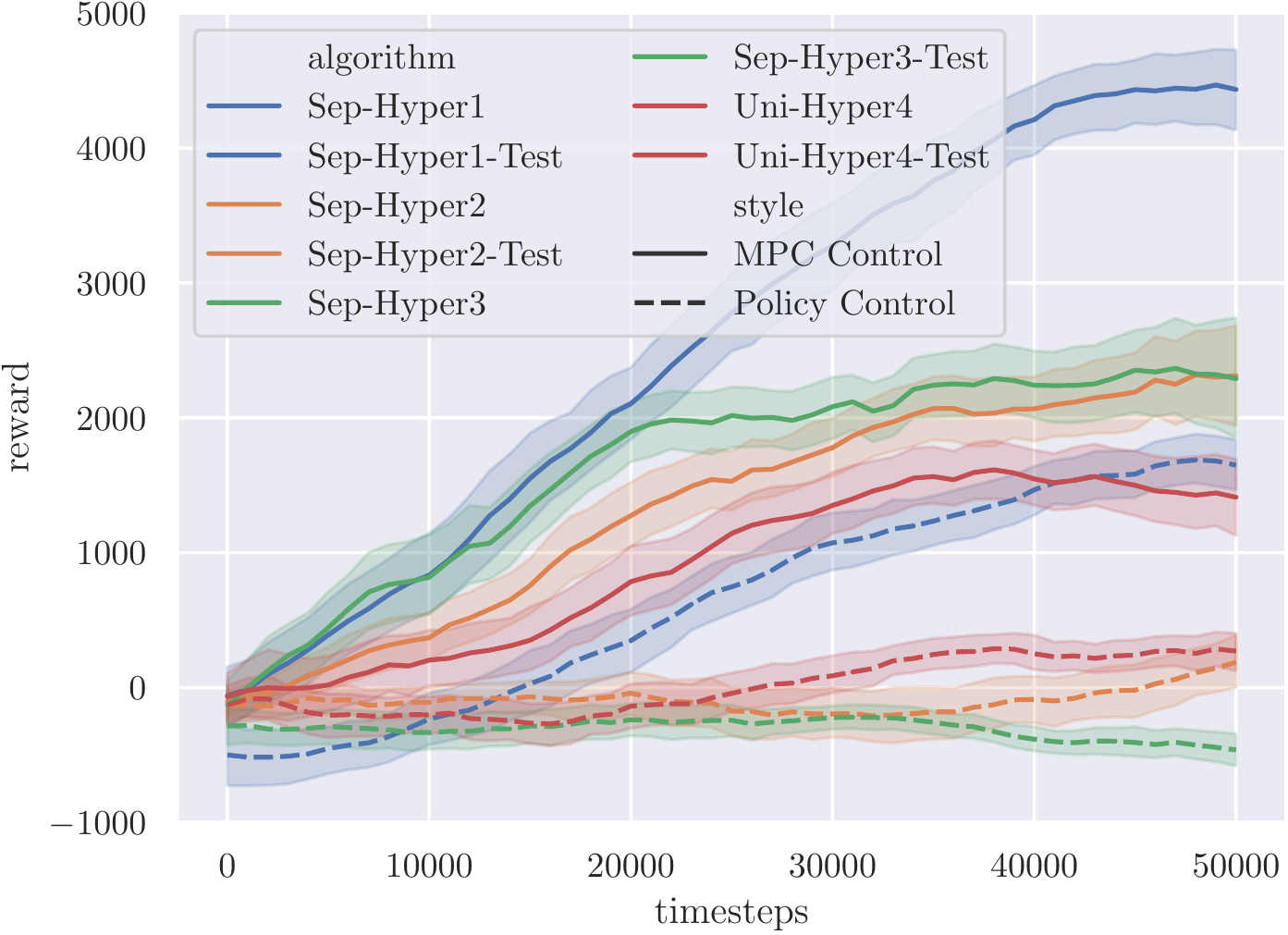}
        (c) \ourmodelshortP{}-GAN
    \end{subfigure}
    \begin{subfigure}{.24\textwidth}
        \centering
        \includegraphics[width=\linewidth]{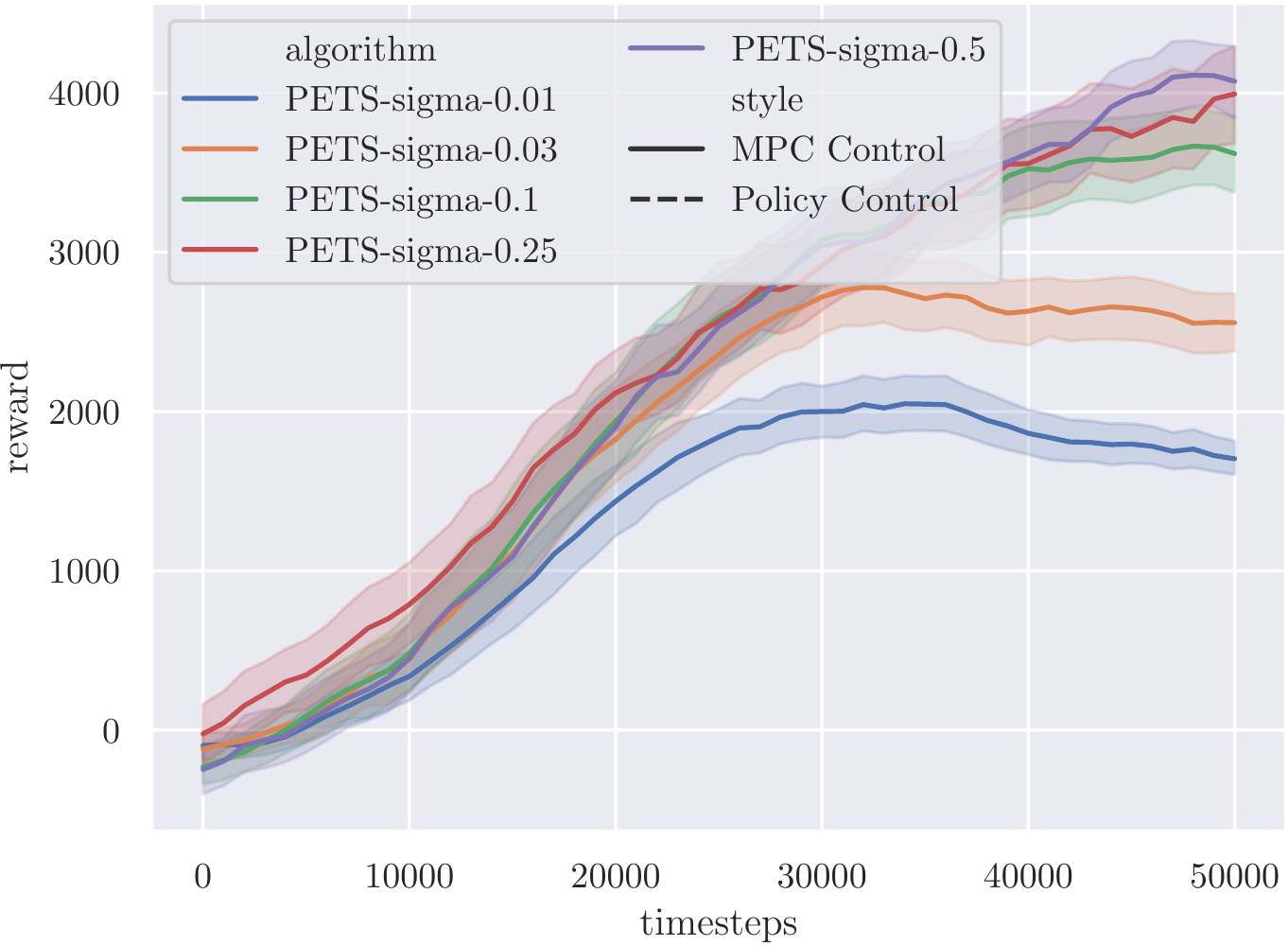}
        (d) PETS
    \end{subfigure}
    \caption{The performance of \ourmodelshortA{}, \ourmodelshortP{}-BC, \ourmodelshortP{}-Avg, \ourmodelshortP{}-GAN using different hyper-parameters.}
    \label{fig:ablation_study}
    \vspace{-0.6cm}
\end{figure}

\section{Conclusions}
In this paper, we explore efficient ways to combine policy networks with model-based planning.
We propose \ourmodelshort{}, which obtains state-of-the-art performance on the MuJoCo benchmarking environments.
We study different distillation schemes to provide fast controllers during testing.
More importantly,
we formulate online planning as optimization using deep neural networks. We believe \ourmodelshort{} will scale to more complex environments in the future.

\newpage
\bibliography{main}\bibliographystyle{plain}

\newpage
\appendix
\section{Appendix}
\subsection{Algorithm Diagrams}\label{appendix:section_diagrams}
To better illustrate the algorithm variants of our proposed methods,
we summarize them in Algorithm~\ref{algorithm:poplina-init},~\ref{algorithm:poplina-replan},~\ref{algorithm:poplinp}.
\begin{algorithm}[!ht]
\begin{algorithmic}[1]
    \State Initialize policy network parameters $\theta$, dynamics network parameters $\phi$, data-set $\mathcal{D}$
    \While {Training iterations not Finished}
        \For {$i^{th}$ time-step of the agent} \Comment{Sampling Data}
        \State Initialize reference action sequence $\{\hat{a}_i, \hat{a}_{i+1}, ..., \hat{a}_{i+\tau}\}.$\Comment{Using Equation~\ref{equation:init_actions}}
        \State Initialize action-sequence noise distribution. $\mu=\mu_0,\,\Sigma=\sigma_0^2\bm{I}$
            \For {$j^{th}$ CEM Update} \Comment{CEM Planning}
                \State Sample action noise sequences $\{\bm{\delta}_i\}$ from $\mathcal{N}(\mu, \Sigma)$.
                \For {Every candidate $\bm{\delta}_i$}\Comment{Trajectory Predicting}
                    \State for $t=i$ to $i+\tau$, $s_{t+1} = f_\phi(s_{t+1}|s_t, a_t = \hat{a}_t + \delta_t)$
                    \State Evaluate expected reward of this candidate.
                \EndFor
                \State Fit distribution of the elite candidates as $\mu', \Sigma'$.
                \State Update noise distribution $\mu = (1-\alpha)\mu + \alpha\mu',\, \Sigma = (1-\alpha)\Sigma+\alpha\Sigma'$
            \EndFor
            \State Execute the first action from the optimal candidate action sequence.
        \EndFor
        \State Update $\phi$ using data-set $\mathcal{D}$ \Comment{Dynamics Update}
        \State Update $\theta$ using data-set $\mathcal{D}$\Comment{Policy Distillation}
    \EndWhile
\end{algorithmic}
\caption{\ourmodelshortA{}-Init}\label{algorithm:poplina-init}
\end{algorithm}

\begin{algorithm}[!ht]
\begin{algorithmic}[1]
    \State Initialize policy network parameters $\theta$, dynamics network parameters $\phi$, data-set $\mathcal{D}$
    \While {Training iterations not Finished}
        \For {$i^{th}$ time-step of the agent} \Comment{Sampling Data}
        \State Initialize action-sequence noise distribution. $\mu=\mu_0,\,\Sigma=\sigma_0^2\bm{I}$
            \For {$j^{th}$ CEM Update} \Comment{CEM Planning}
                \State Sample action noise sequences $\{\bm{\delta}_i\}$ from $\mathcal{N}(\mu, \Sigma)$.
                \For {Every candidate $\bm{\delta}_i$}\Comment{Trajectory Predicting}
                    \State for $t=i$ to $i+\tau$, $s_{t+1} = f_\phi(s_{t+1}|s_t, a_t = \pi_{\theta}(s_t) + \delta_t)$
                    \State Evaluate expected reward of this candidate.
                \EndFor
                \State Fit distribution of the elite candidates as $\mu', \Sigma'$.
                \State Update noise distribution $\mu = (1-\alpha)\mu + \alpha\mu',\, \Sigma = (1-\alpha)\Sigma+\alpha\Sigma'$
            \EndFor
            \State Execute the first action from the optimal candidate action sequence.
        \EndFor
        \State Update $\phi$ using data-set $\mathcal{D}$ \Comment{Dynamics Update}
        \State Update $\theta$ using data-set $\mathcal{D}$\Comment{Policy Distillation}
    \EndWhile
\end{algorithmic}
\caption{\ourmodelshortA{}-Replan}\label{algorithm:poplina-replan}
\end{algorithm}

\begin{algorithm}[!ht]
\begin{algorithmic}[1]
    \State Initialize policy network parameters $\theta$, dynamics network parameters $\phi$, data-set $\mathcal{D}$
    \While {Training iterations not Finished}
        \For {$i^{th}$ time-step of the agent} \Comment{Sampling Data}
        \State Initialize parameter-sequence noise distribution. $\mu=\mu_0,\,\Sigma=\sigma_0^2\bm{I}$
            \For {$j^{th}$ CEM Update} \Comment{CEM Planning}
                \State Sample parameter noise sequences $\{\bm{\omega}_i\}$ from $\mathcal{N}(\mu, \Sigma)$.
                \For {Every candidate $\bm{\omega}_i$}\Comment{Trajectory Predicting}
                    \State for $t=i$ to $i+\tau$, $s_{t+1} = f_\phi(s_{t+1}|s_t, a_t = \pi_{\theta + \omega_t}(s_t))$
                    \State Evaluate expected reward of this candidate.
                \EndFor
                \State Fit distribution of the elite candidates as $\mu', \Sigma'$.
                \State Update noise distribution $\mu = (1-\alpha)\mu + \alpha\mu',\, \Sigma = (1-\alpha)\Sigma+\alpha\Sigma'$
            \EndFor
            \State Execute the first action from the optimal candidate action sequence.
        \EndFor
        \State Update $\phi$ using data-set $\mathcal{D}$ \Comment{Dynamics Update}
        \State Update $\theta$ using data-set $\mathcal{D}$\Comment{Policy Distillation}
    \EndWhile
\end{algorithmic}
\caption{\ourmodelshortP{}}\label{algorithm:poplinp}
\end{algorithm}

\subsection{Bench-marking Environments}\label{appendix:environments}
\begin{figure}[!ht]
    \centering
    \begin{subfigure}{.33\textwidth}
        \centering
        \includegraphics[width=\linewidth]{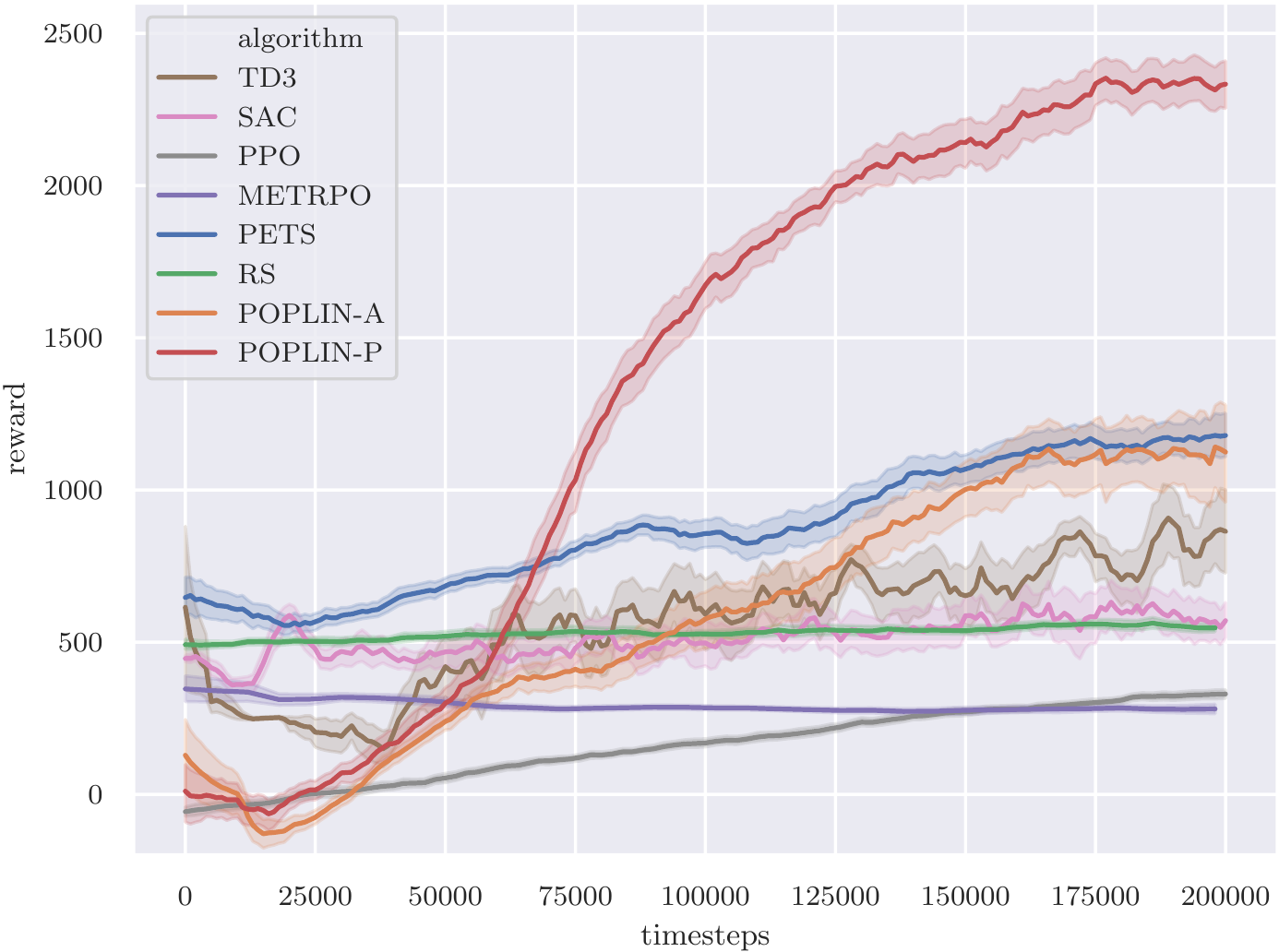}
        (a) Ant
    \end{subfigure}%
    \begin{subfigure}{.33\textwidth}
        \centering
        \includegraphics[width=\linewidth]{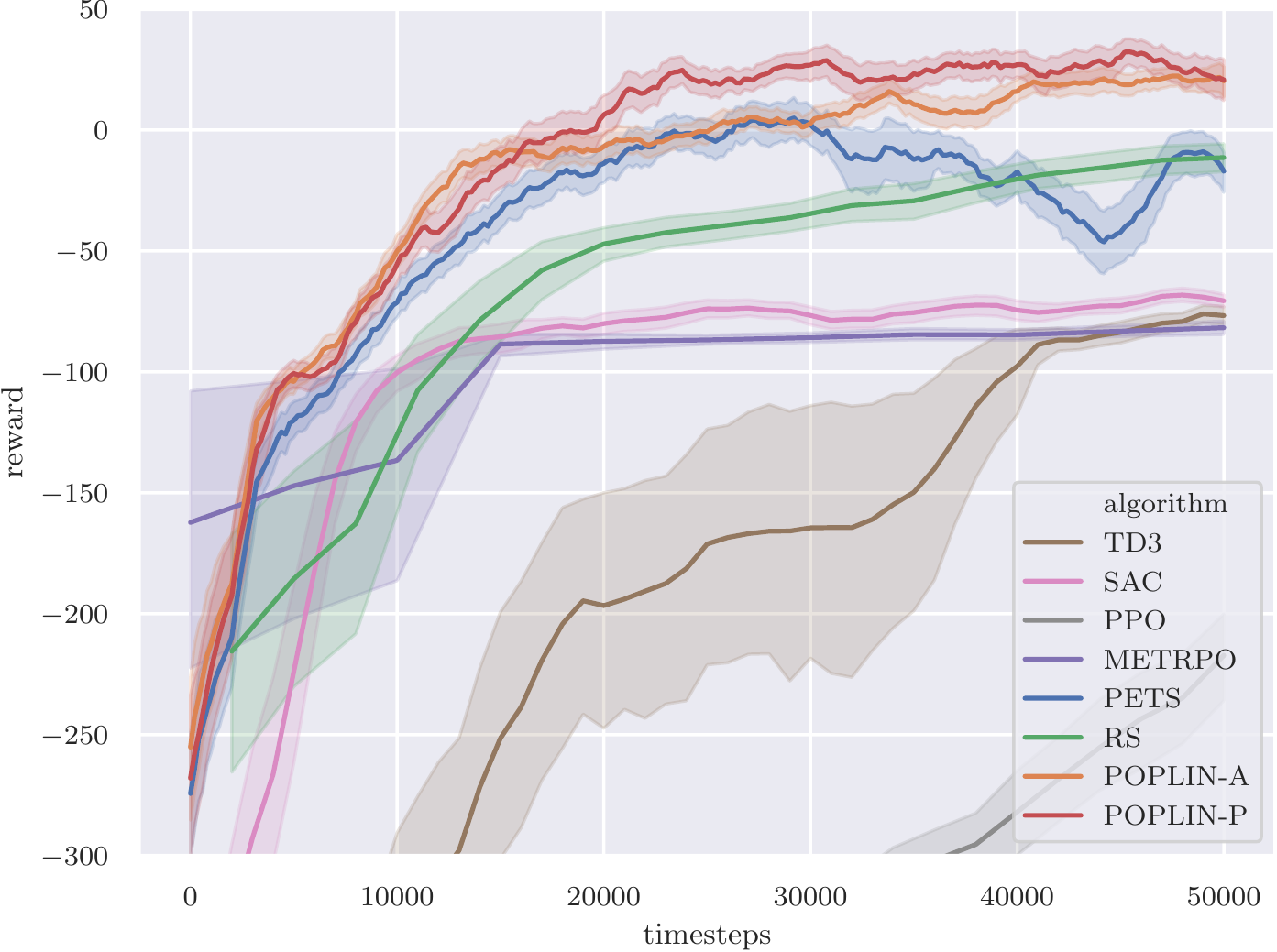}
        (b) Acrobot
    \end{subfigure}
    \begin{subfigure}{.33\textwidth}
        \centering
        \includegraphics[width=\linewidth]{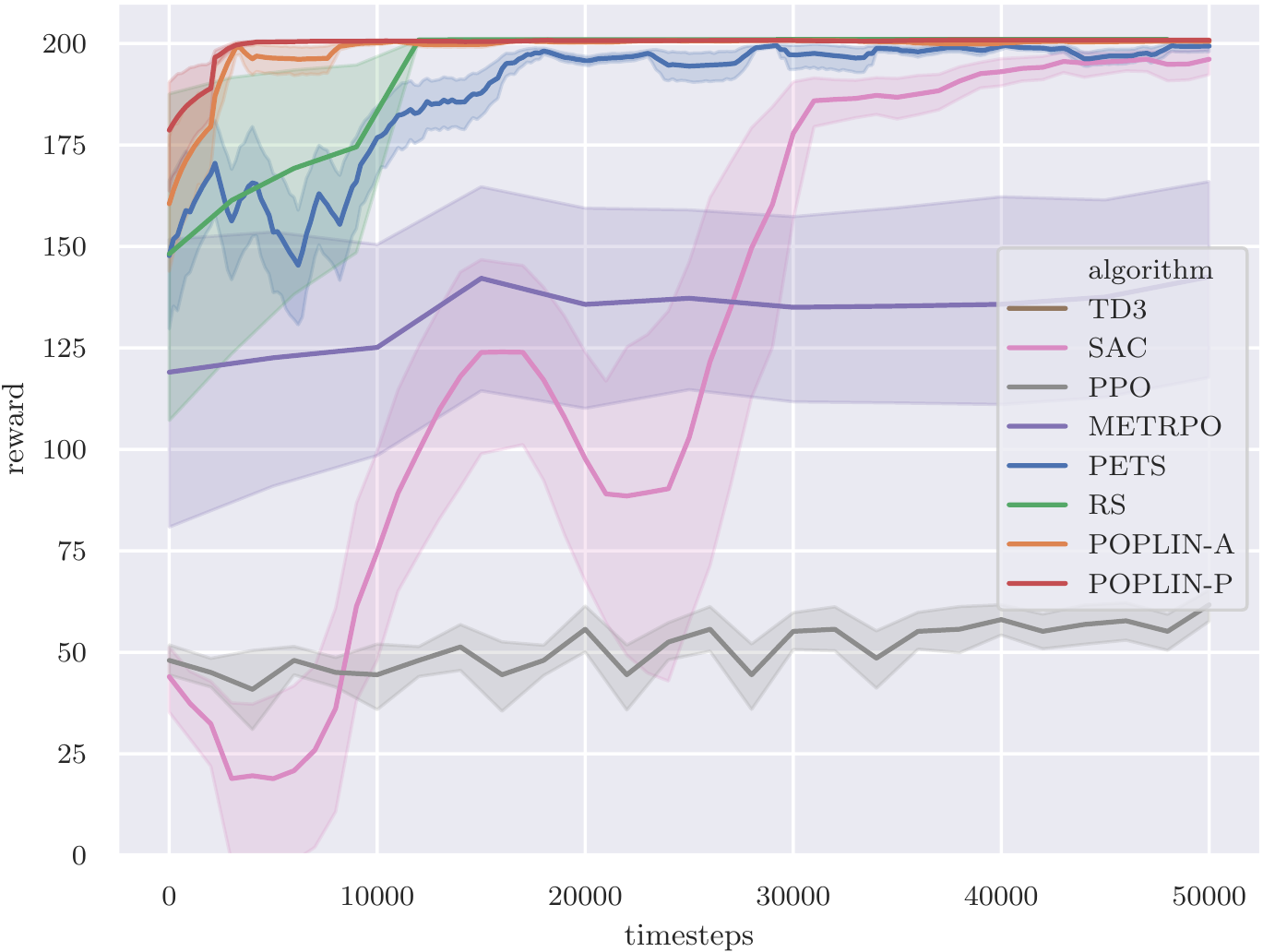}
        (c) Cartpole
    \end{subfigure}
    
    \begin{subfigure}{.33\textwidth}
        \centering
        \includegraphics[width=\linewidth]{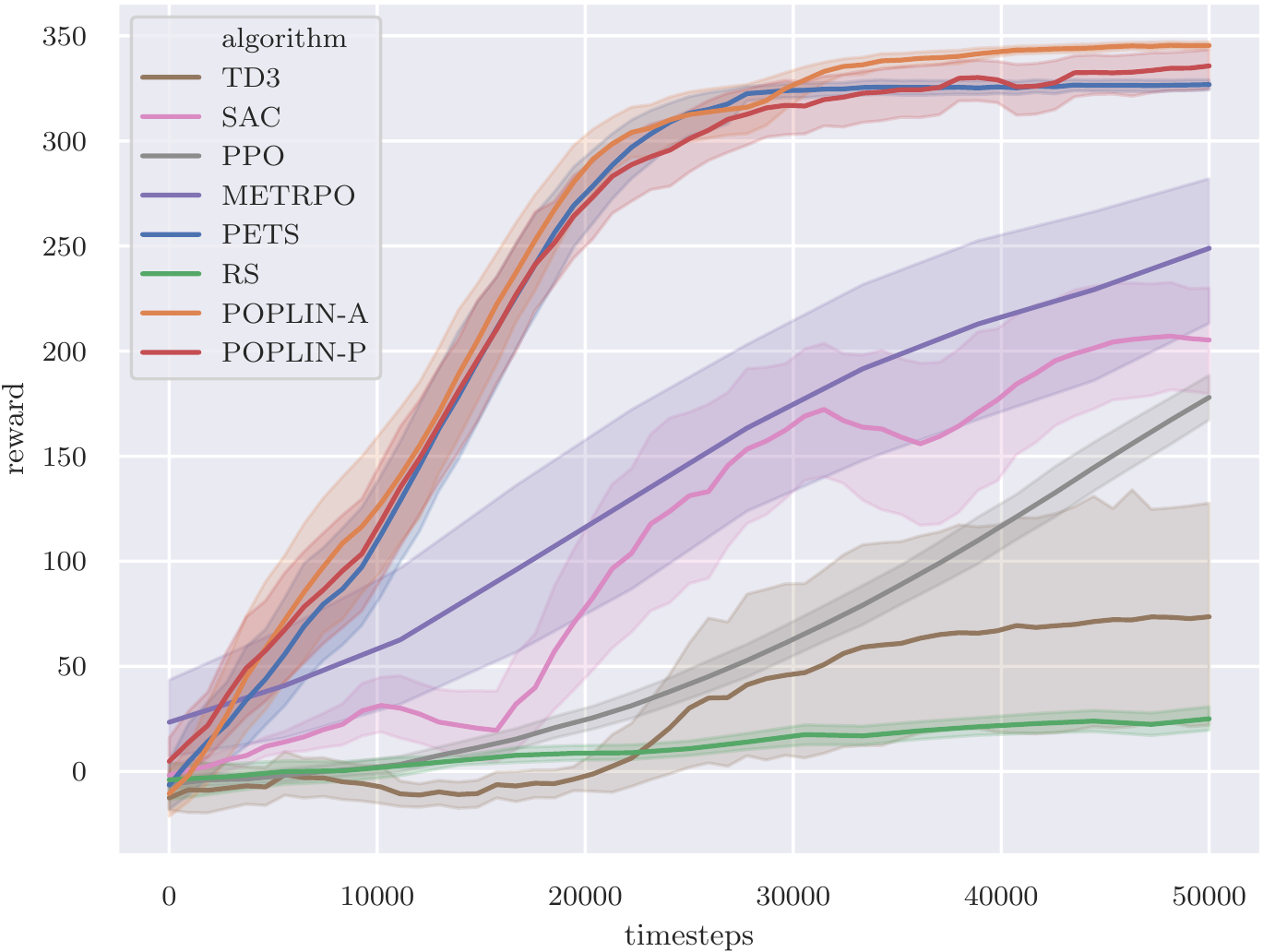}
        (d) Swimmer
    \end{subfigure}
    \begin{subfigure}{.33\textwidth}
        \centering
        \includegraphics[width=\linewidth]{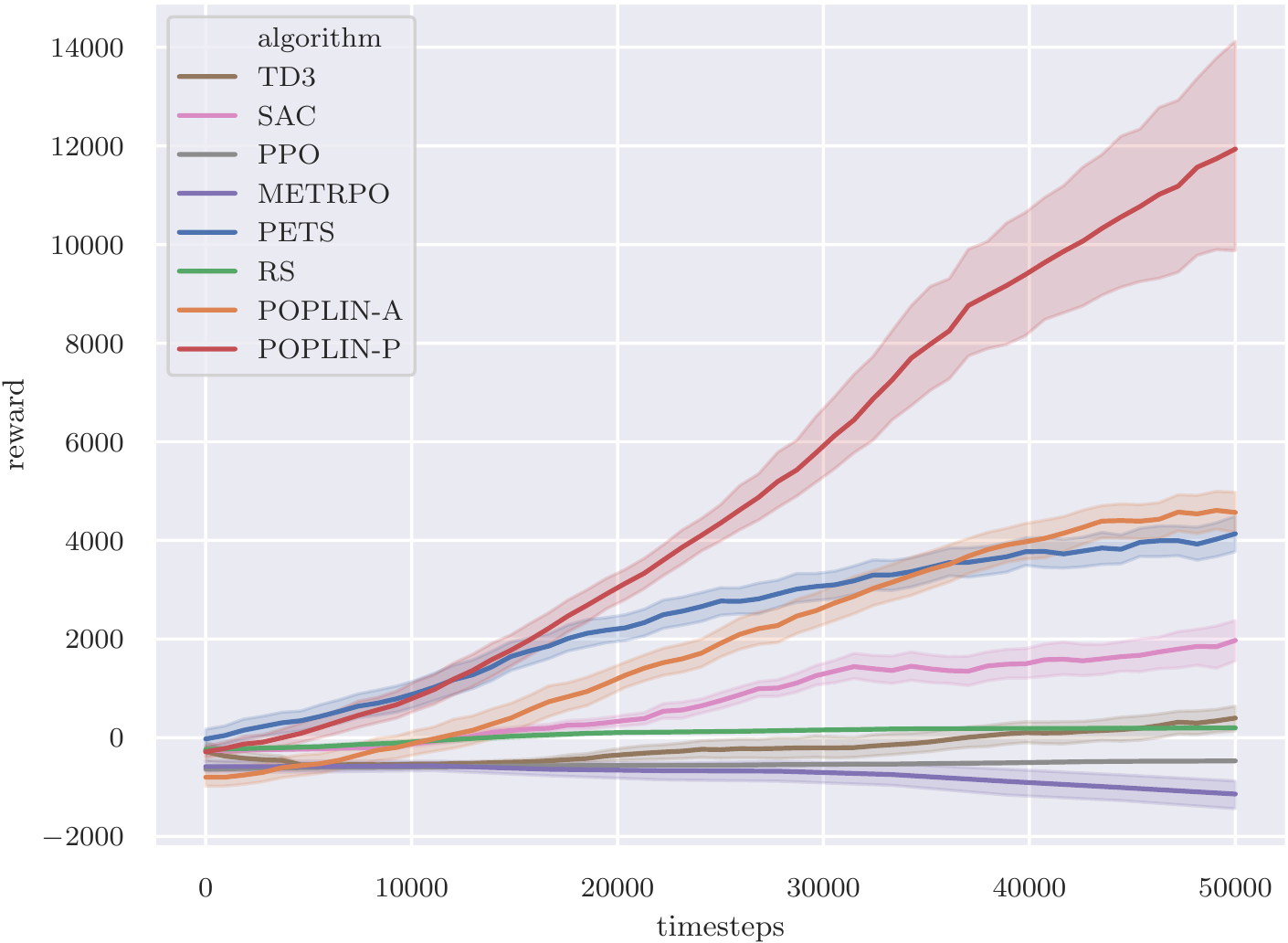}
        (e) Cheetah
    \end{subfigure}%
    \begin{subfigure}{.33\textwidth}
        \centering
        \includegraphics[width=\linewidth]{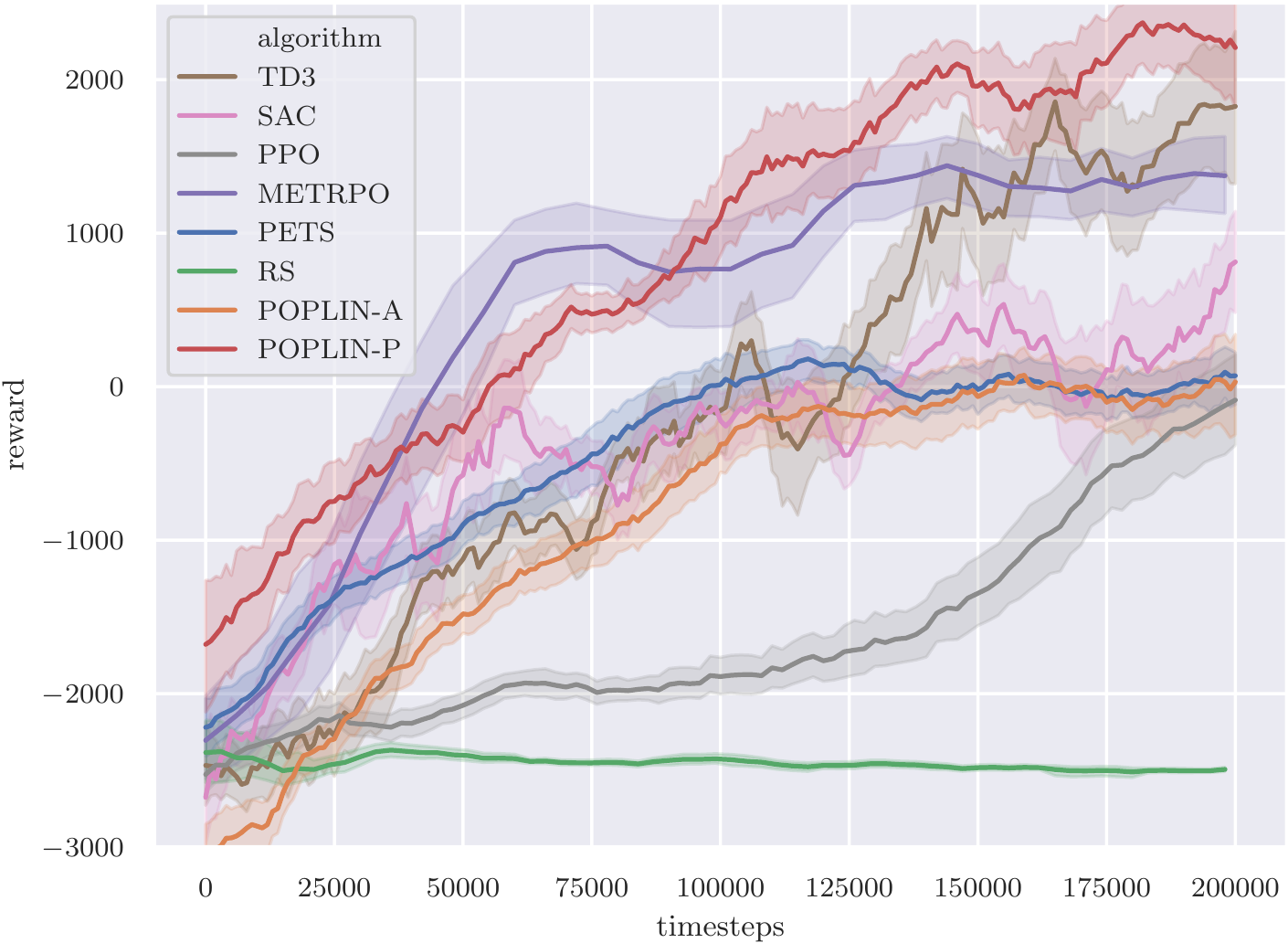}
        (f) Hopper
    \end{subfigure}%
    
    \begin{subfigure}{.33\textwidth}
        \centering
        \includegraphics[width=\linewidth]{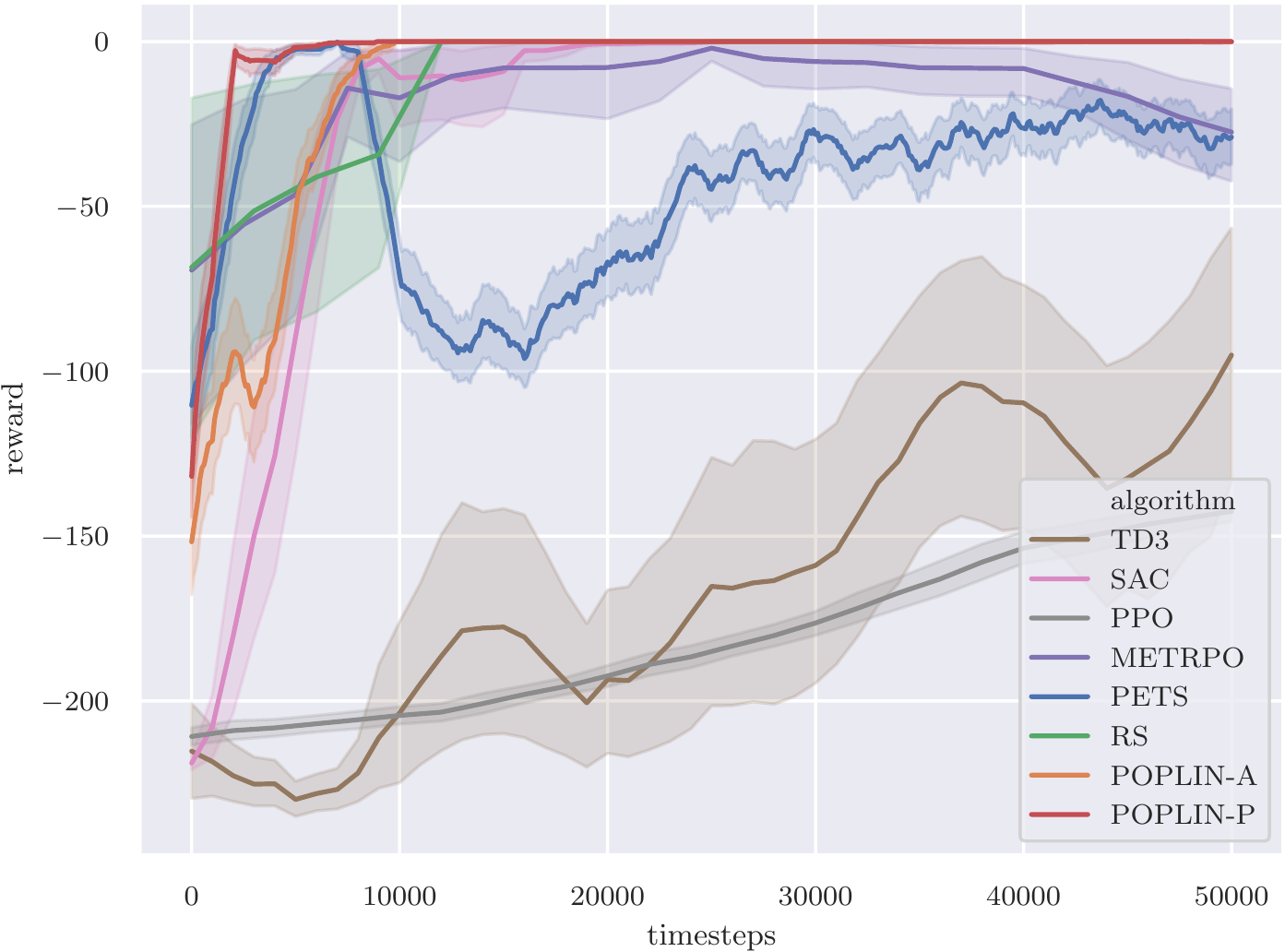}
        (g) InvertedPendulum
    \end{subfigure}
    \begin{subfigure}{.33\textwidth}
        \centering
        \includegraphics[width=\linewidth]{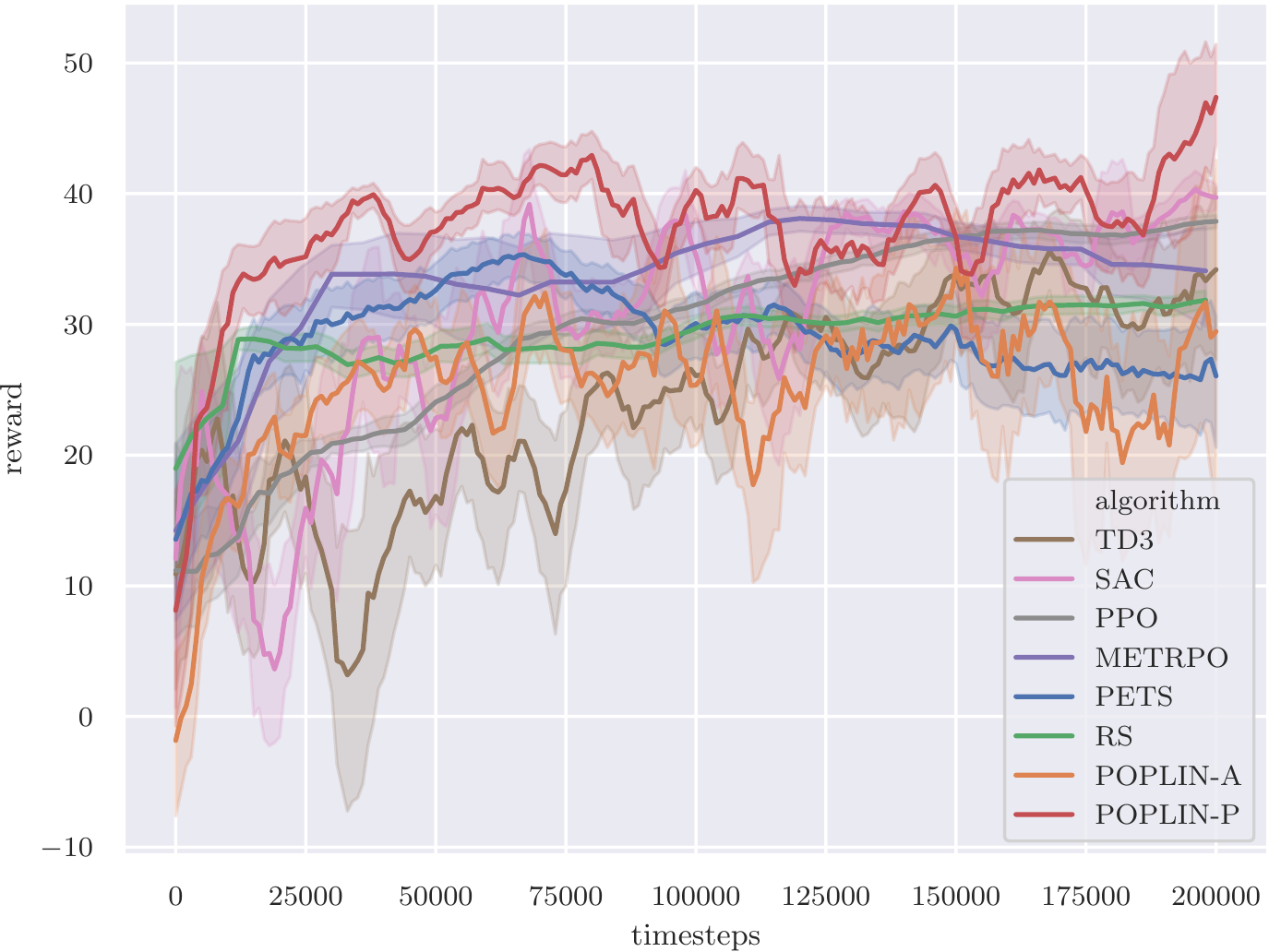}
        (h) Swimmer-v0
    \end{subfigure}%
    \begin{subfigure}{.33\textwidth}
        \centering
        \includegraphics[width=\linewidth]{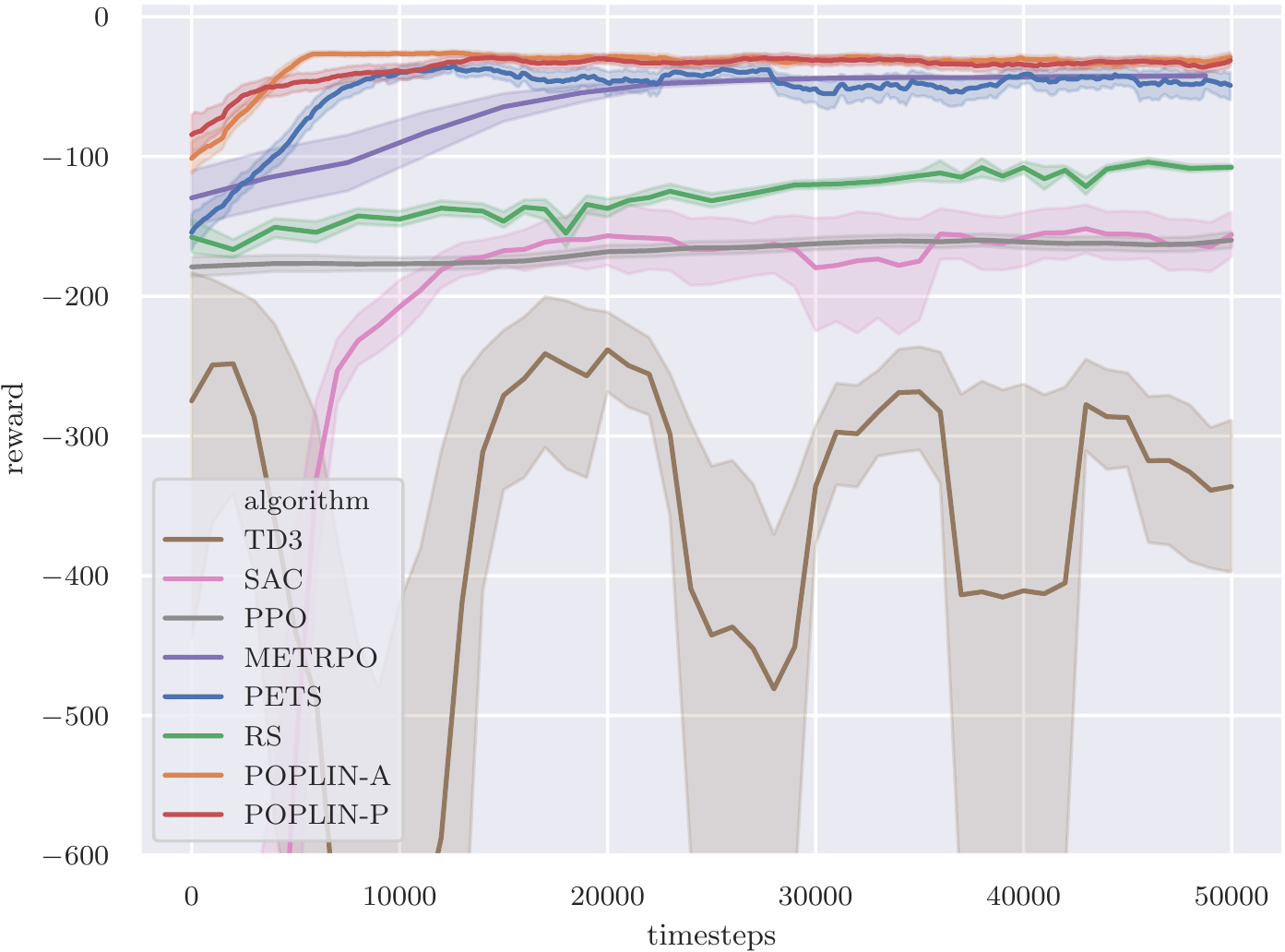}
        (i) Reacher3D
    \end{subfigure}%
    
    \begin{subfigure}{.33\textwidth}
        \centering
        \includegraphics[width=\linewidth]{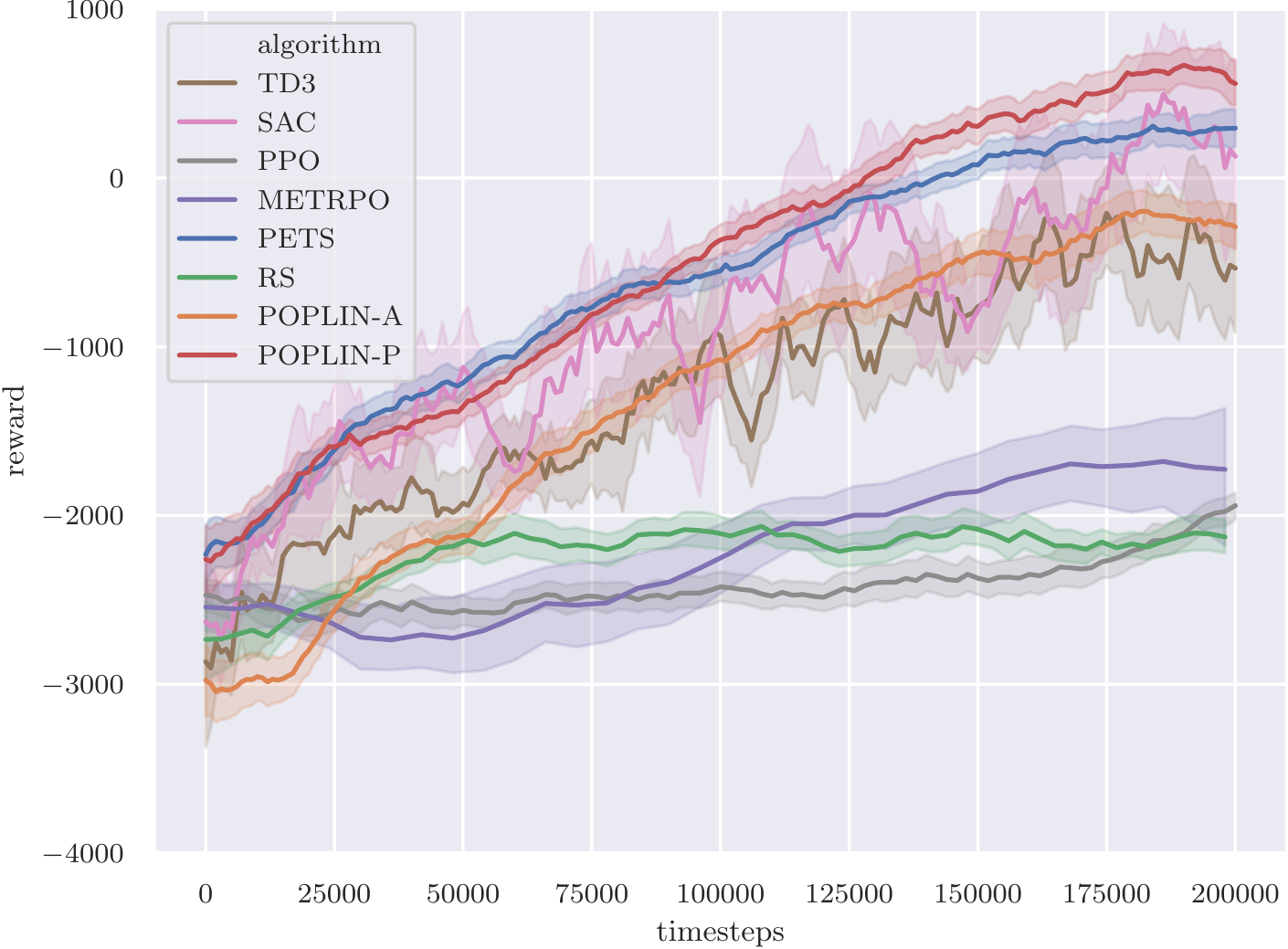}
        (j) Walker2d
    \end{subfigure}%
    \begin{subfigure}{.33\textwidth}
        \centering
        \includegraphics[width=\linewidth]{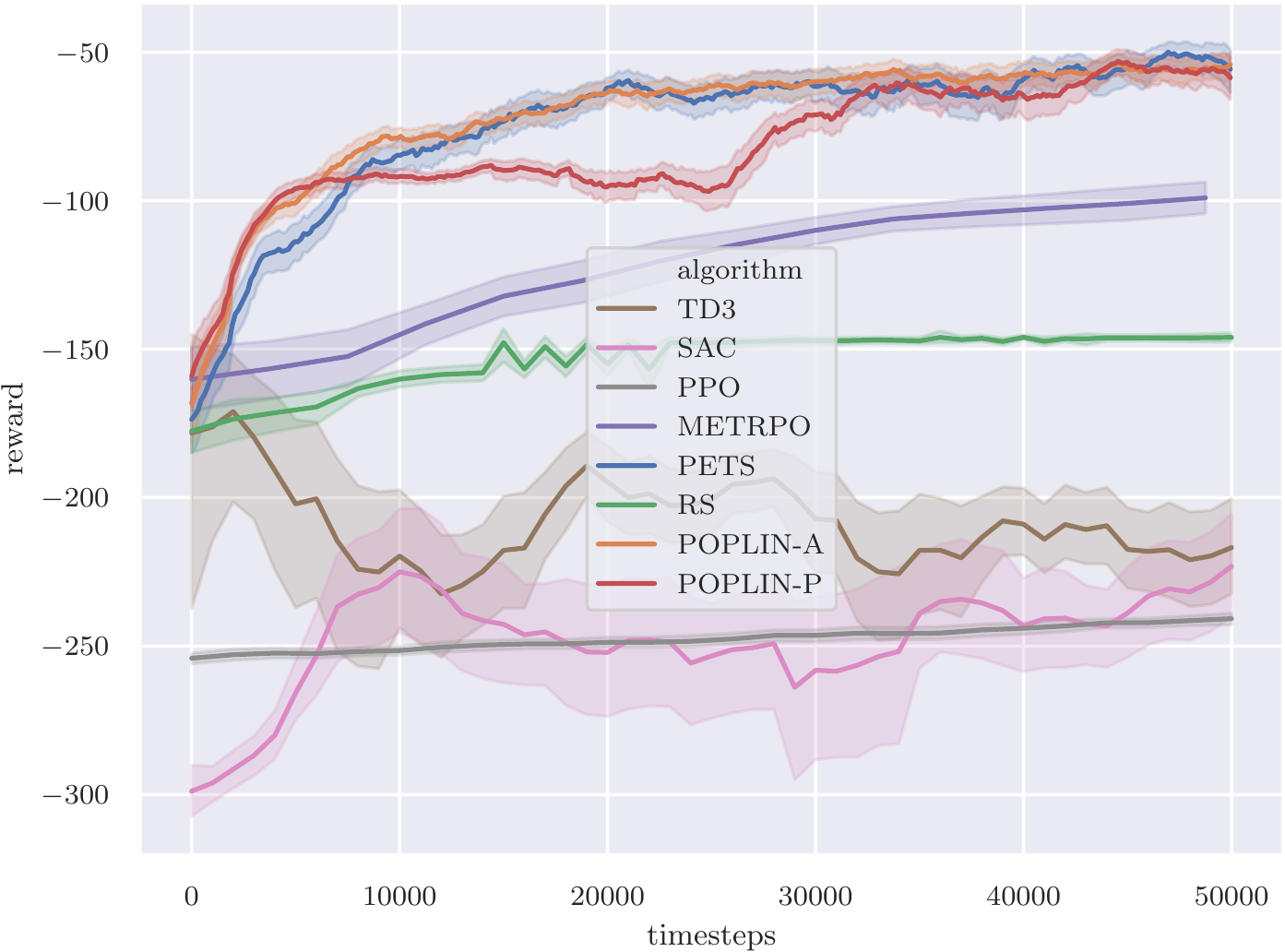}
        (k) Pusher
    \end{subfigure}%
    \begin{subfigure}{.33\textwidth}
        \centering
        \includegraphics[width=\linewidth]{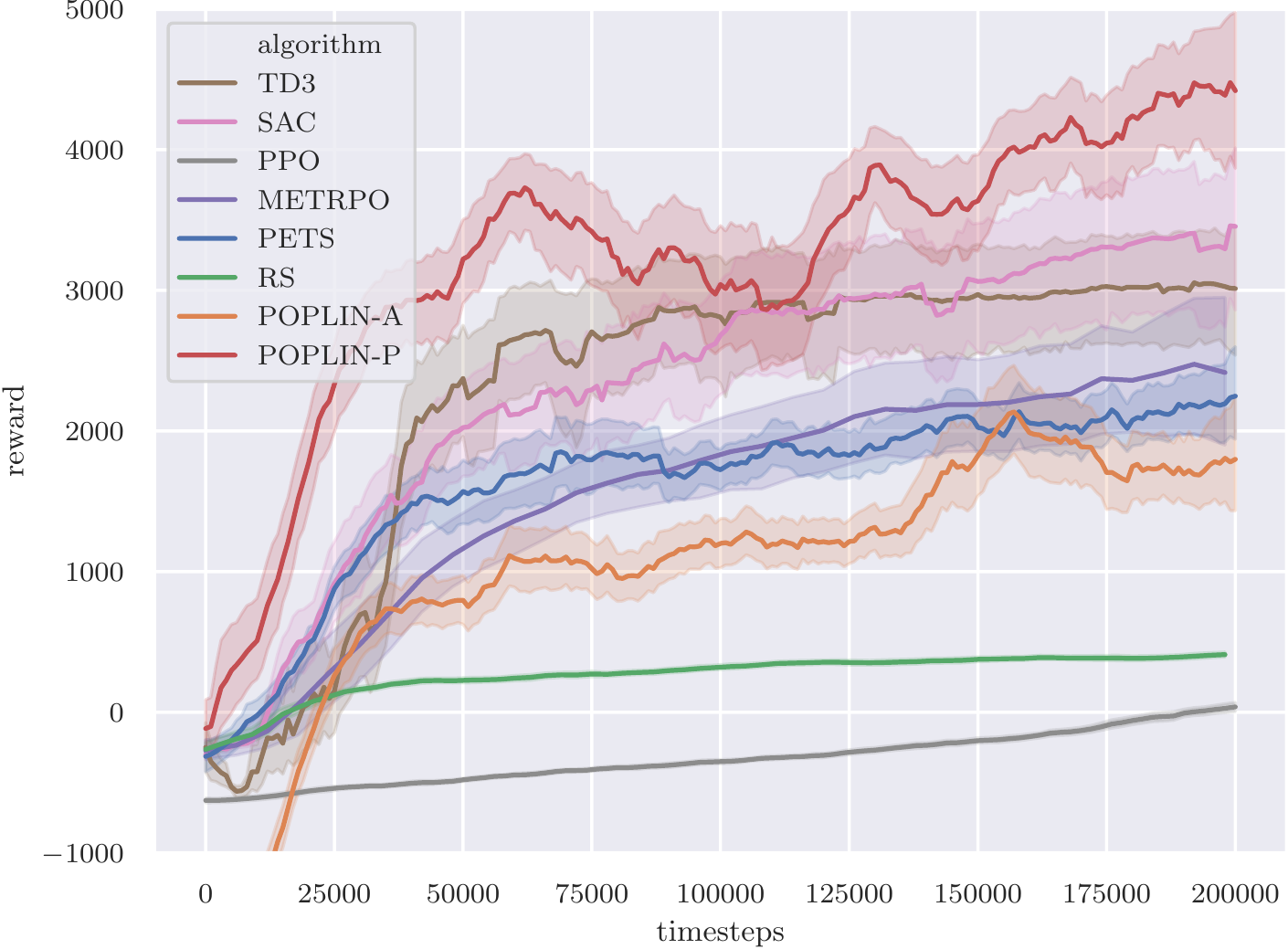}
        (l) Cheetah-v0
    \end{subfigure}%

    \caption{Full Performance of \ourmodelshortP{}, \ourmodelshortA{} and other state-of-the-art algorithms on 12 different bench-marking environments. In the figure, we include baselines such as TD3, SAC, PPO, METRPO, PETS, RS and our proposed algorithm.}
    \label{fig:full_benchmarking_performance}
\end{figure}

\begin{table}[!t]
\resizebox{1\textwidth}{!}{\begin{tabular}{c|c|c|c|c|c|c|c}
\toprule
& Cheetah  & Ant            & Hopper       & Swimmer        & Cheetah-v0  & Walker2d      & Swimmer-v0    \\
\midrule
POPLIN-P & 12227.9 $\pm$ 5652.8 & 2330.1 $\pm$ 320.9 & 2055.2 $\pm$ 613.8   & 334.4 $\pm$ 34.2  & 4235.0 $\pm$ 1133.0 & 597.0 $\pm$ 478.8   & 37.1 $\pm$ 4.6  \\
POPLIN-A & 4651.1 $\pm$ 1088.5  & 1148.4 $\pm$ 438.3 & 202.5 $\pm$ 962.5    & 344.9 $\pm$ 7.1   & 1562.8 $\pm$ 1136.7 & -105.0 $\pm$ 249.8  & 26.7 $\pm$ 13.2 \\
PETS     & 4204.5 $\pm$ 789.0   & 1165.5 $\pm$ 226.9 & 114.9 $\pm$ 621.0    & 326.2 $\pm$ 12.6  & 2288.4 $\pm$ 1019.0 & 282.5 $\pm$ 501.6   & 29.7 $\pm$ 13.5 \\
RS       & 191.1 $\pm$ 21.2     & 535.5 $\pm$ 37.0   & -2491.5 $\pm$ 35.1   & 22.4 $\pm$ 9.7    & 421.0 $\pm$ 55.2    & -2060.3 $\pm$ 228.0 & 26.8 $\pm$ 2.3  \\
MBMF     & -459.5 $\pm$ 62.5    & 134.2 $\pm$ 50.4   & -1047.4 $\pm$ 1098.7 & 110.7 $\pm$ 45.6  & 126.9 $\pm$ 72.7    & -2218.1 $\pm$ 437.7 & 30.6 $\pm$ 4.9  \\
TRPO     & -412.4 $\pm$ 33.3    & 323.3 $\pm$ 24.9   & -2100.1 $\pm$ 640.6  & 47.8 $\pm$ 11.1   & -12.0 $\pm$ 85.5    & -2286.3 $\pm$ 373.3 & 26.3 $\pm$ 2.6  \\
PPO      & -483.0 $\pm$ 46.1    & 321.0 $\pm$ 51.2   & -103.8 $\pm$ 1028.0  & 155.5 $\pm$ 14.9  & 17.2 $\pm$ 84.4     & -1893.6 $\pm$ 234.1 & 24.7 $\pm$ 4.0  \\
GPS      & 129.4 $\pm$ 140.4    & 445.5 $\pm$ 212.9  & -768.5 $\pm$ 200.9   & -30.9 $\pm$ 6.3   & 52.3 $\pm$ 41.7     & -1730.8 $\pm$ 441.7 & 8.2 $\pm$ 10.2  \\
METRPO   & -744.8 $\pm$ 707.1   & 282.2 $\pm$ 18.0   & 1272.5 $\pm$ 500.9   & 225.5 $\pm$ 104.6 & 2283.7 $\pm$ 900.4  & -1609.3 $\pm$ 657.5 & 35.4 $\pm$ 2.2  \\
TD3      & 218.9 $\pm$ 593.3    & 870.1 $\pm$ 283.8  & 1816.6 $\pm$ 994.8   & 72.1 $\pm$ 130.9  & 3015.7 $\pm$ 969.8  & -516.4 $\pm$ 812.2  & 17.0 $\pm$ 12.9 \\
SAC      & 1745.9 $\pm$ 839.2   & 548.1 $\pm$ 146.6  & 788.3 $\pm$ 738.2    & 204.6 $\pm$ 69.3  & 3459.8 $\pm$ 1326.6 & 164.5 $\pm$ 1318.6  & 23.0 $\pm$ 17.3 \\
Random   & -284.2 $\pm$ 83.3    & 478.0 $\pm$ 47.8   & -2768.0 $\pm$ 571.6  & -12.4 $\pm$ 12.8  & -312.4 $\pm$ 44.2   & -2450.1 $\pm$ 406.5 & 2.4 $\pm$ 12.0 \\
\midrule
Time-step & 50000 & 200000 & 200000 & 50000 & 200000 & 200000 & 200000 \\
\bottomrule
\end{tabular}}
\vspace{0.1cm}
\caption{Performance of each algorithm on environments based on OpenAI Gym~\cite{gym} MuJoCo\cite{2012mujoco} environments. In the table, we record the performance at 200,000 time-step.}\label{table:full_benchmarking_stats_harder}
\end{table}

\begin{table}[!t]
\resizebox{1\textwidth}{!}{\begin{tabular}{c|c|c|c|c|c|c}
\toprule
& Reacher3D  & Pusher       & Pendulum    & InvertedPendulum & Acrobot      & Cartpole    \\
\midrule
POPLIN-P & -29.0 $\pm$ 25.2   & -55.8 $\pm$ 23.1  & 167.9 $\pm$ 45.9       & -0.0 $\pm$ 0.0     & 23.2 $\pm$ 27.2   & 200.8 $\pm$ 0.3    \\
POPLIN-A & -27.7 $\pm$ 25.2   & -56.0 $\pm$ 24.3  & 178.3 $\pm$ 19.3       & -0.0 $\pm$ 0.0     & 20.5 $\pm$ 20.1   & 200.6 $\pm$ 1.3    \\
PETS     & -47.7 $\pm$ 43.6   & -52.7 $\pm$ 23.5  & 155.7 $\pm$ 79.3       & -29.5 $\pm$ 37.8   & -18.4 $\pm$ 46.3  & 199.6 $\pm$ 4.6    \\
RS       & -107.6 $\pm$ 5.2   & -146.4 $\pm$ 3.2  & 161.2 $\pm$ 11.5       & -0.0 $\pm$ 0.0     & -12.5 $\pm$ 14.3  & 201.0 $\pm$ 0.0    \\
MBMF     & -168.6 $\pm$ 23.2  & -285.8 $\pm$ 15.2 & 163.7 $\pm$ 15.2       & -202.3 $\pm$ 17.0  & -146.8 $\pm$ 29.9 & 22.5 $\pm$ 67.7    \\
TRPO     & -176.5 $\pm$ 24.3  & -235.5 $\pm$ 6.2  & 158.7 $\pm$ 9.1        & -134.6 $\pm$ 6.9   & -291.2 $\pm$ 6.7  & 46.3 $\pm$ 6.0     \\
PPO      & -162.2 $\pm$ 15.7  & -243.2 $\pm$ 6.9  & 160.9 $\pm$ 12.5       & -137.3 $\pm$ 12.4  & -205.4 $\pm$ 51.5 & 68.8 $\pm$ 4.9     \\
GPS      & -552.8 $\pm$ 577.7 & -151.2 $\pm$ 1.3  & 164.3 $\pm$ 4.1        & -14.7 $\pm$ 20.7   & -214.3 $\pm$ 15.3 & -18.7 $\pm$ 101.1  \\
METRPO   & -43.5 $\pm$ 3.7    & -98.5 $\pm$ 12.6  & 174.8 $\pm$ 6.2        & -29.3 $\pm$ 29.5   & -78.7 $\pm$ 5.0   & 138.5 $\pm$ 63.2   \\
TD3      & -331.6 $\pm$ 134.6 & -216.4 $\pm$ 39.6 & 168.6 $\pm$ 12.7       & -102.9 $\pm$ 101.0 & -76.5 $\pm$ 10.2  & -409.2 $\pm$ 928.8 \\
SAC      & -161.6 $\pm$ 43.7  & -227.6 $\pm$ 42.2 & 159.5 $\pm$ 12.1       & -0.2 $\pm$ 0.1     & -69.4 $\pm$ 7.0   & 195.5 $\pm$ 8.7    \\
Random   & -183.1 $\pm$ 41.5  & -199.0 $\pm$ 10.0 & -249.5 $\pm$ 228.4     & -205.9 $\pm$ 12.1  & -374.1 $\pm$ 15.6 & 31.3 $\pm$ 36.3    \\
\midrule
Time-step & 50000 & 50000 & 50000 & 50000 & 50000 & 50000 \\
\bottomrule
\end{tabular}}
\vspace{0.1cm}
\caption{Performance of each algorithm on environments based on OpenAI Gym~\cite{gym} classic control environments. In the table, we record the performance at 50000 time-step.}\label{table:full_benchmarking_stats_easier}
\end{table}
In the original PETS paper~\cite{chua2018deep}, the authors only experiment with 4 environments,
which are namely Reacher3D, Pusher, Cartpole and Cheetah.
In this paper, we experiment with the 9 more environments based on the standard bench-marking environments from OpenAI Gym~\cite{gym}.
More specifically, we experiment with InvertedPendulum, Acrobot, Pendulum, Ant, Hopper, Swimmer, Walker2d.
We also note that the Cheetah environment in PETS~\cite{chua2018deep} is different from the standard HalfCheetah-v1 in OpenAI Gym.
Therefore we experiment with both versions in our paper,
where the Cheetah from PETS is named as "Cheetah",
and the HalfCHeetah from OpenAI Gym is named as "Cheetah-v0".
Empirically, Cheetah is much easier to solve than Cheetah-v0,
as show in Table~\ref{table:full_benchmarking_stats_harder} and Table~\ref{table:full_benchmarking_stats_easier}.
We also include two swimmer, which we name as Swimmer and Swimmer-v0, which we explain in section~\ref{appendix:fixed_swimmer}.

\subsubsection{Fixing the Swimmer Environments}\label{appendix:fixed_swimmer}
We also notice that after an update in the Gym environments,
the swimmer became unsolvable for almost all algorithms.
The reward threshold for solving is around 340 for the original swimmer,
but almost all algorithms, including the results shown in many published papers~\cite{ppo},
will be stuck at the 130 reward local-minima.
We note that this is due the fact that the velocity sensor is on the neck of the swimmer,
making swimmer extremely prone to this performance local-minimum.
We provide a fixed swimmer, which we name as Swimmer, by moving the sensor from the neck to the head.
We believe this modification is necessary to test the effectiveness of the algorithms.

\subsection{Full Results of Bench-marking Performance}\label{appendix:full_benchmarking}
In this section, we show the figures of all the environments in Figure~\ref{fig:full_benchmarking_performance}.
We also include the final performance in the Table~\ref{table:full_benchmarking_stats_harder} and~\ref{table:full_benchmarking_stats_easier}.
As we can see, \ourmodelshort{} has consistently the best performance among almost all the environments.
We also include the time-steps we use on each environment for all the algorithms in Table~\ref{table:full_benchmarking_stats_harder} and~\ref{table:full_benchmarking_stats_easier}.

\subsubsection{Hyper-parameters}\label{appendix:hyper_search}
In this section, we introduce the hyper-parameters we search during the experiments.
One thing to notice is that,
for all of the experiments on PETS, \ourmodelshort{},
we use the model type PE (probabilistic ensembles) and propagation method of E (expectation).
While other combinations of model type and propagation methods might result in better performance,
they are usually prohibitively computationally expensive.
For example, the combination of PE-DS requires a training time of about 68 hours for one random seed,
for PETS to train with 200 iteration, which is 200,000 time-step.
As a matter of fact, PE-E is actually one of the best combination in many environments.
Since \ourmodelshort{} is based on PETS,
we believe this is a fair comparison for all the algorithms.

We show the hyper-parameter search we perform for PETS in the paper in Table~\ref{table:pets_hyper}.
For the hyper-parameters specific to \ourmodelshort{},
we summarize them in~\ref{table:our_A_hyper} and~\ref{table:our_P_hyper}.
\begin{table}[!ht]
    \centering
    \begin{tabular}{l|c}
        \toprule
        Hyper-parameter & Value Tried \\
        \midrule
        Population Size & 100, 200, ..., 2000 \\
        Planning Horizon & 30, 50, 100 \\
        Initial Distribution Sigma & 0.01, 0.03, 0.1, 0.25, 0.3, 0.5 \\
        CEM Iterations & 5, 8, 10, 20\\
        ELite Size $\xi$ & 50, 100, 200\\
        \bottomrule
    \end{tabular}\vspace{0.1cm}
    \caption{Hyper-parameter grid search options for PETS.}
    \label{table:pets_hyper}
\end{table}
\begin{table}[!ht]
    \centering
    \begin{tabular}{l|c}
        \toprule
        Hyper-parameter & Value Tried \\
        \midrule
        Training Data & real data, hallucination data \\
        Variant & Replan, Init \\
        Initial Distribution Sigma & 0.001, 0.003, 0.01, 0.03, 0.1\\
        \bottomrule
    \end{tabular}\vspace{0.1cm}
    \caption{Hyper-parameter grid search options for \ourmodelshortA{}.}
    \label{table:our_A_hyper}
\end{table}
\begin{table}[!ht]
    \centering
    \begin{tabular}{l|c}
        \toprule
        Hyper-parameter & Value Tried \\
        \midrule
        Training Data & real data, hallucination data \\
        Training Variant & BC, GAN, Avg \\
        Noise Variant & Uni, Sep \\
        Initial Distribution Sigma & 0.001, 0.003, 0.01, 0.03, 0.1\\
        \bottomrule
    \end{tabular}\vspace{0.1cm}
    \caption{Hyper-parameter grid search options for \ourmodelshortP{}. 
    We also experiment with using WGAN in~\cite{salimans2016improved} to train the policy network, which does not results in good performance and is not put into the article.}
    \label{table:our_P_hyper}
\end{table}

\subsection{Full Results of Policy Control}\label{appendix:full_testing}
\begin{figure}[!th]
    \centering
    \begin{subfigure}{.33\textwidth}
        \centering
        \includegraphics[width=\linewidth]{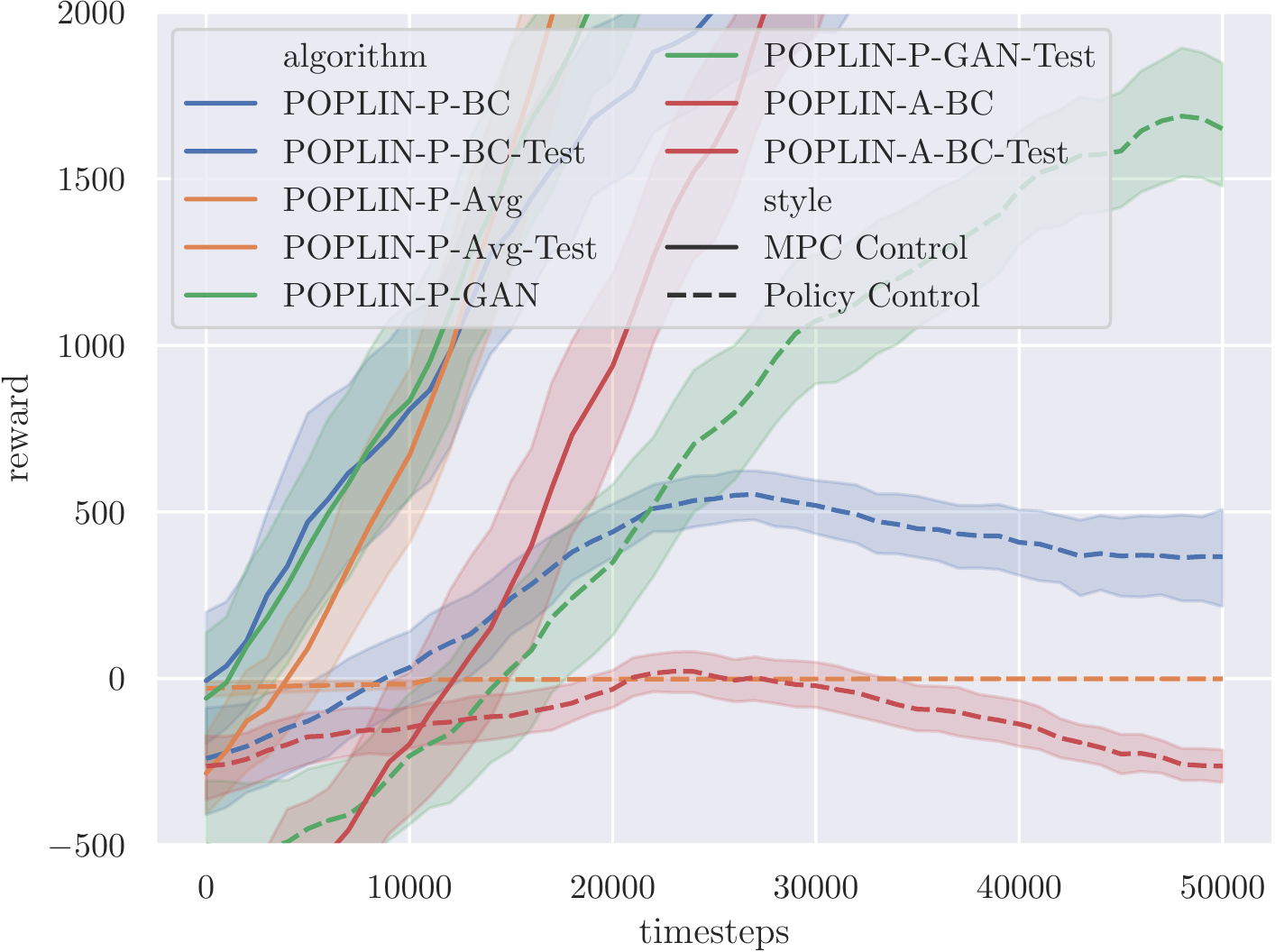}
        (a) Cheetah
    \end{subfigure}%
    \begin{subfigure}{.33\textwidth}
        \centering
        \includegraphics[width=\linewidth]{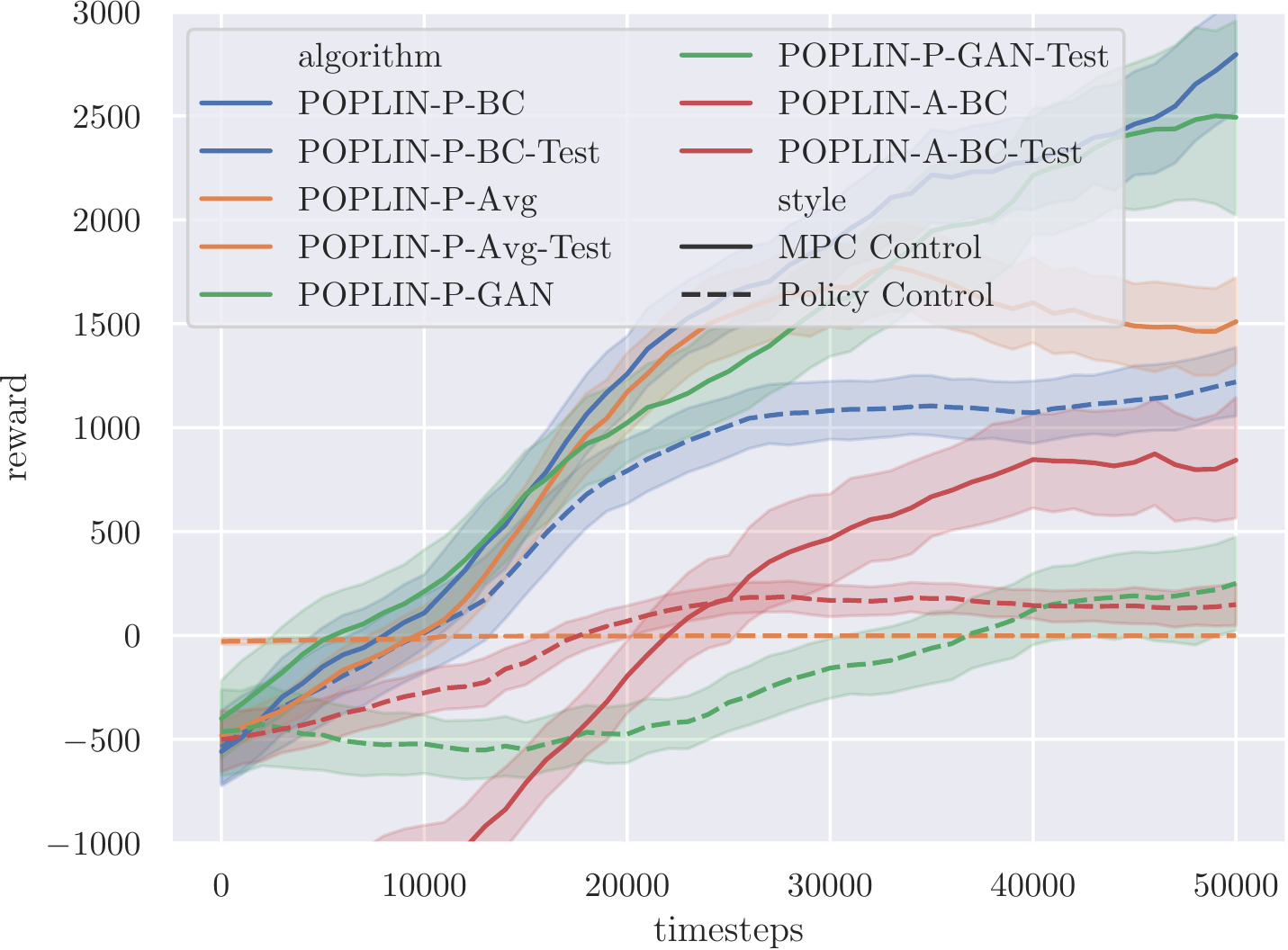}
        (b) Cheetah-v0
    \end{subfigure}%
    \begin{subfigure}{.33\textwidth}
        \centering
        \includegraphics[width=\linewidth]{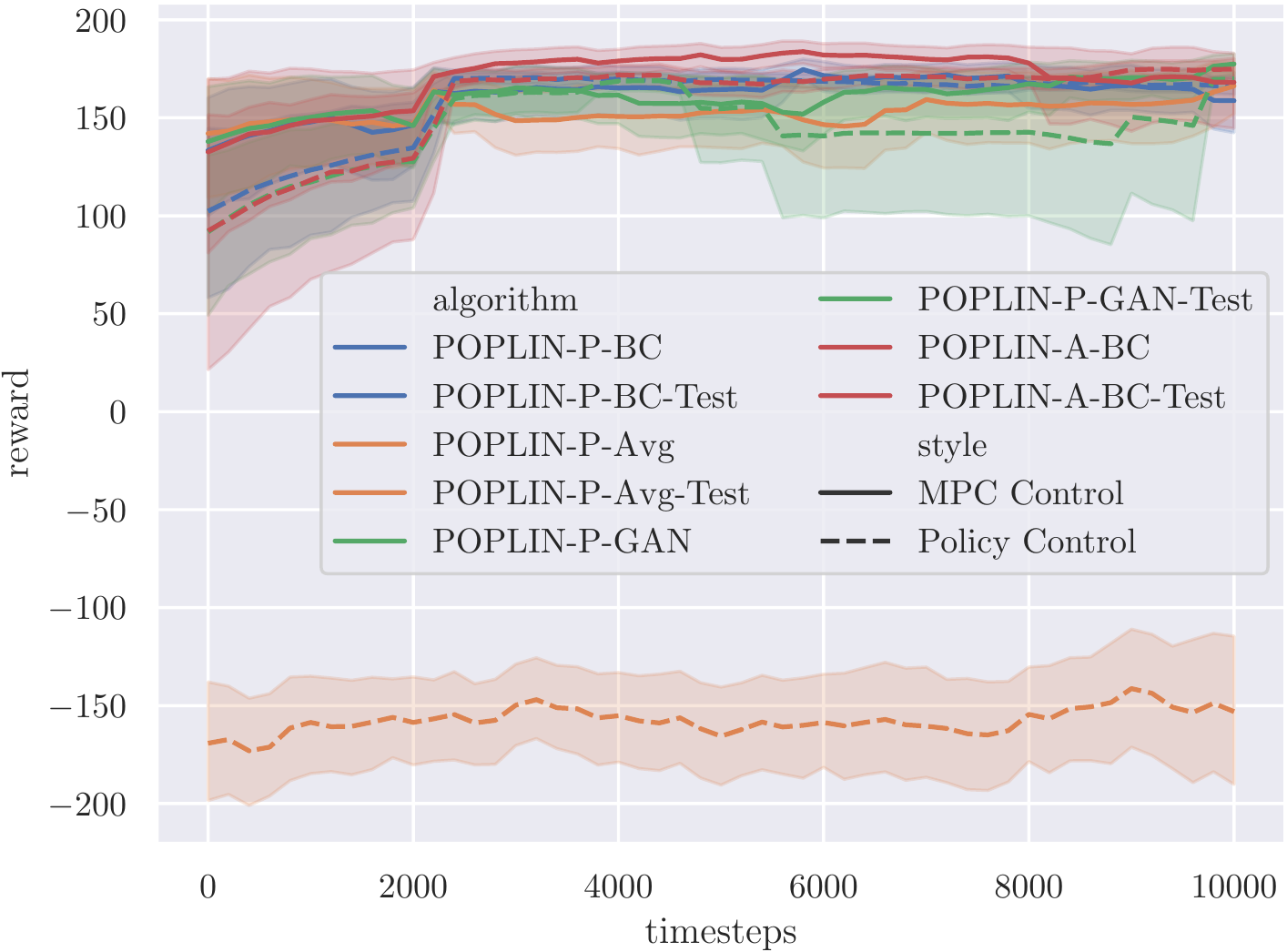}
        (c) Pendulum
    \end{subfigure}
    
    \begin{subfigure}{.33\textwidth}
        \centering
        \includegraphics[width=\linewidth]{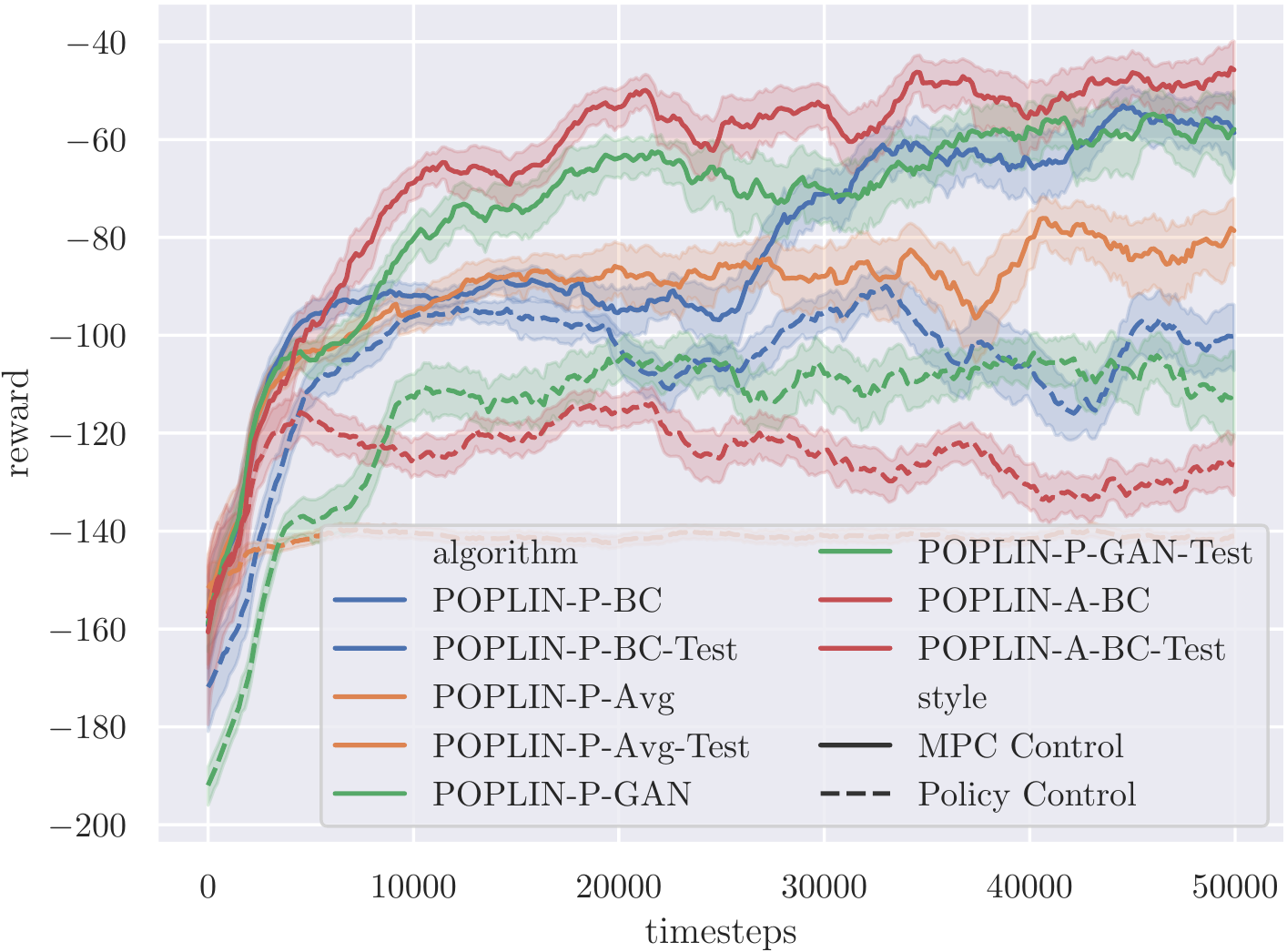}
        (e) Pusher
    \end{subfigure}
    \begin{subfigure}{.33\textwidth}
        \centering
        \includegraphics[width=\linewidth]{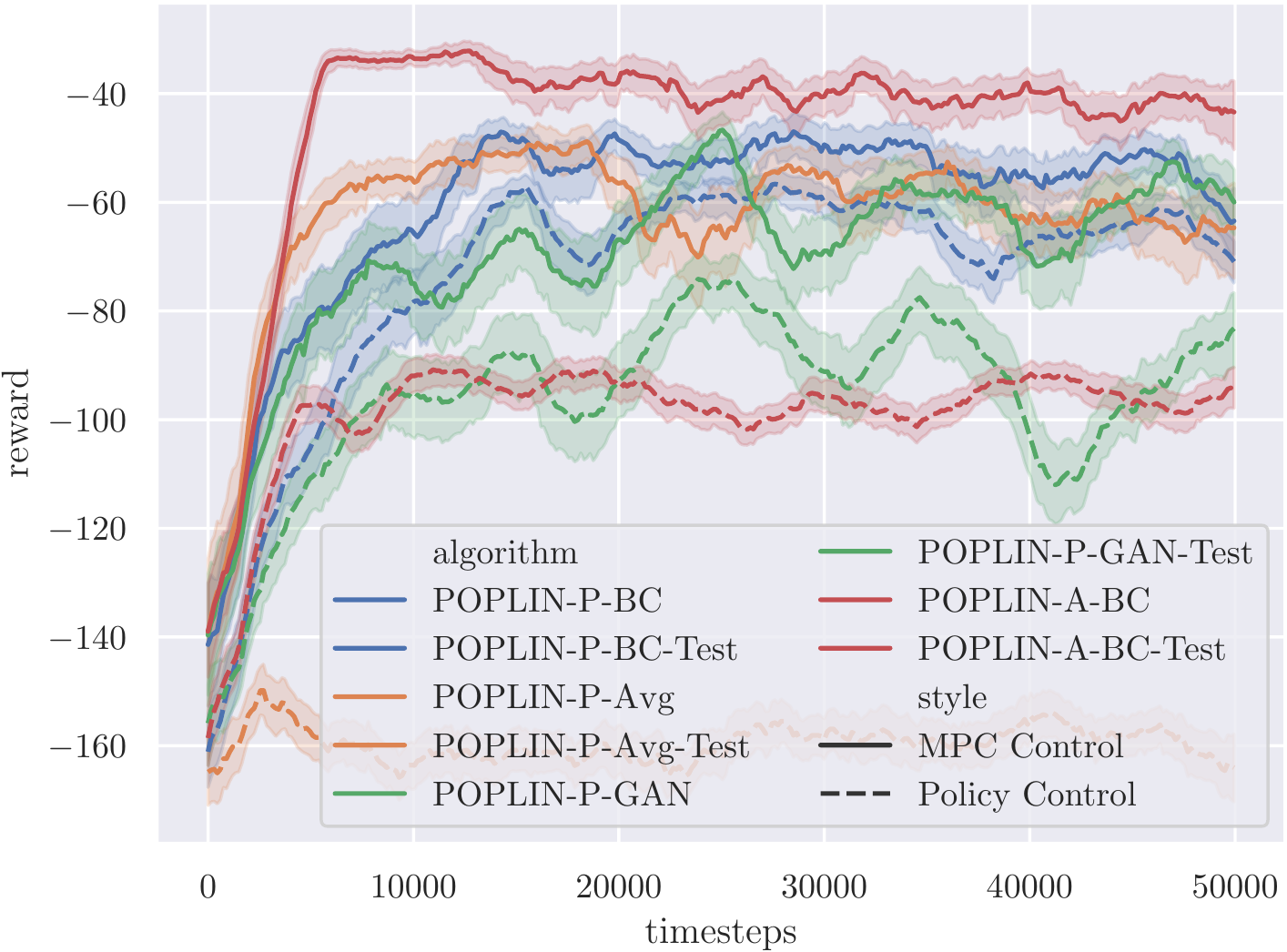}
        (f) Reacher3D
    \end{subfigure}%
    \caption{The planning performance and the testing performance of the proposed \ourmodelshortA{}, and \ourmodelshortP{} with its three training schemes, which are namely behavior cloning (BC), generative adversarial network training (GAN) and setting parameter average (Avg).}
    \label{fig:test_performance_full}
\end{figure}

Due to the space limit, we are not able to put all of the results of policy control in the main article.
More specifically, we add the figure for the original Cheetah-v0 compared to the figures shown in the main article,
as can be seen in~\ref{fig:test_performance_full} (b).
Again, we note that \ourmodelshortP{}-BC and \ourmodelshortP{}-GAN are comparable to each other,
as mentioned in the main article.
\ourmodelshortP{}-BC and \ourmodelshortP{}-GAN are the better algorithms respectively in Cheetah and Cheetah-v0,
which are essentially the same environment with different observation functions.

\subsection{Ablation Study for Different Variant of \ourmodelshort{}}\label{appendix:ablation_study}
\begin{figure}[!t]
    \centering
    \begin{subfigure}{.33\textwidth}
        \centering
        \includegraphics[width=\linewidth]{pcem_3/pwcem_exp_3_POPLIN-A-crop.pdf}
        (a) \ourmodelshortA{}
    \end{subfigure}
    \begin{subfigure}{.33\textwidth}
        \centering
        \includegraphics[width=\linewidth]{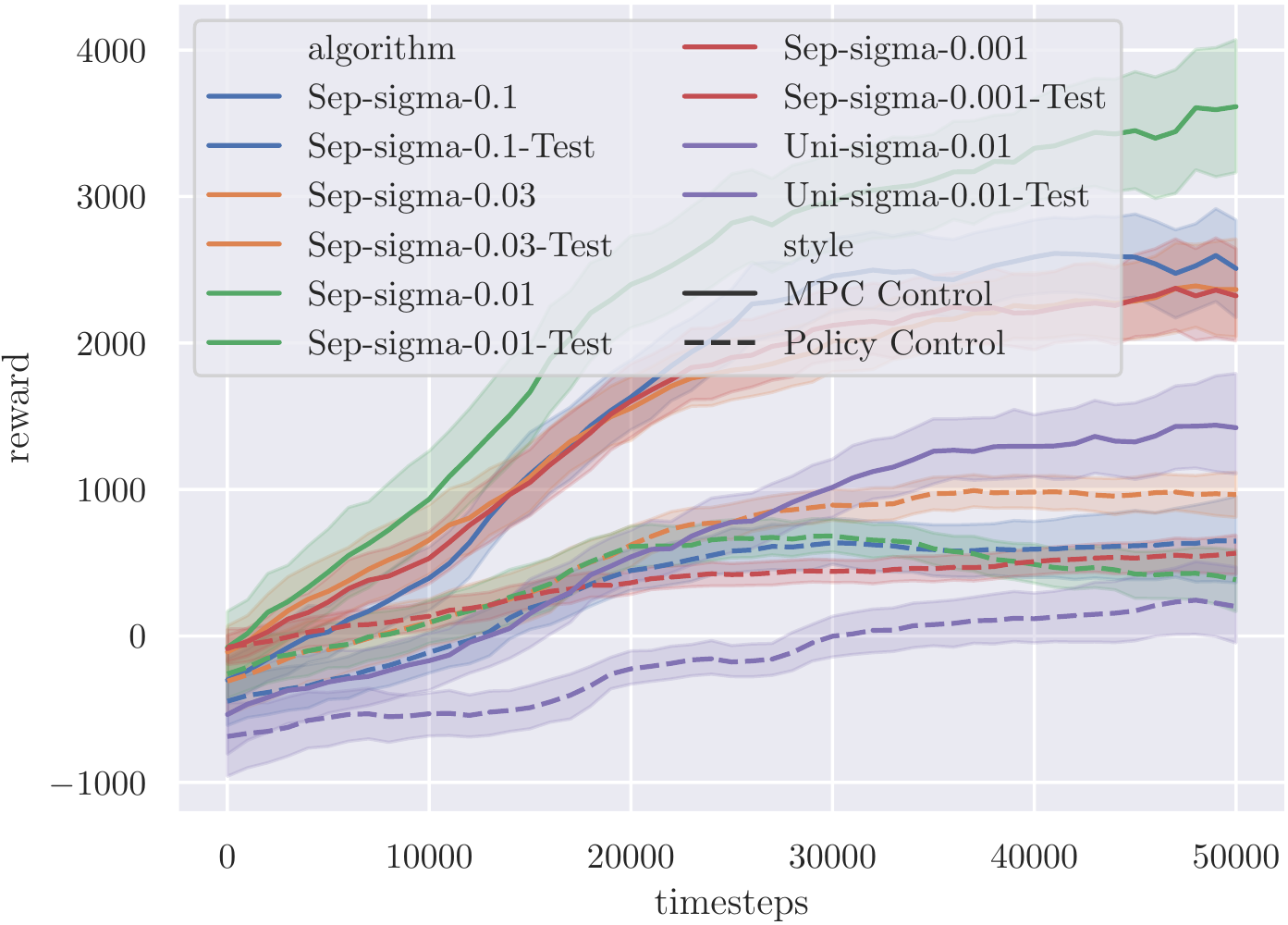}
        (b) \ourmodelshortP{}-BC
    \end{subfigure}%
    \begin{subfigure}{.33\textwidth}
        \centering
        \includegraphics[width=\linewidth]{pcem_3/pwcem_exp_3_POPLIN-P-Avg-crop.pdf}
        (c) \ourmodelshortP{}-Avg
    \end{subfigure}
    \begin{subfigure}{.33\textwidth}
        \centering
        \includegraphics[width=\linewidth]{pcem_3/pwcem_exp_3_PETS-crop.pdf}
        (d) PETS
    \end{subfigure}
    \begin{subfigure}{.33\textwidth}
        \centering
        \includegraphics[width=\linewidth]{pcem_3/pwcem_exp_3_POPLIN-P-GAN-crop.pdf}
        (d) \ourmodelshortP{}-GAN
    \end{subfigure}
    \caption{The performance of \ourmodelshortA{}, \ourmodelshortP{}-BC, \ourmodelshortP{}-Avg, \ourmodelshortP{}-GAN using different hyper-parameters.}
    \label{fig:ablation_study_full}
\end{figure}

\begin{figure}[!t]
    \centering
    \begin{subfigure}{.50\textwidth}
        \centering
        \includegraphics[width=\linewidth]{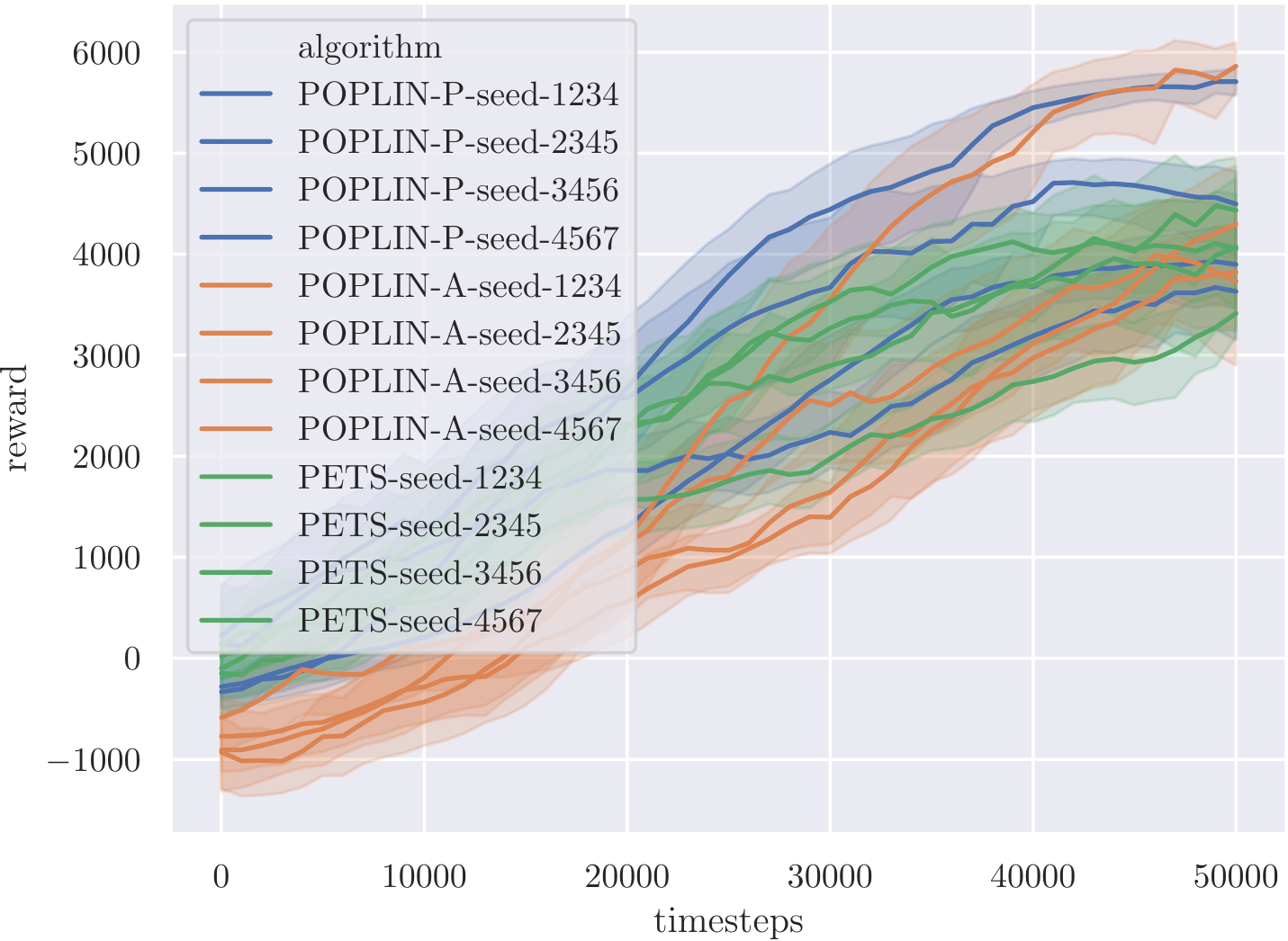}
        (e) Random seeds
    \end{subfigure}
    \caption{The performance of \ourmodelshortA{}, \ourmodelshortP, and PETS of different random seeds.}
    \label{fig:appendix_seed}
\end{figure}

In this section, we show the results of different variant of our algorithm.
In Figure~\ref{fig:appendix_seed}, the performances of different random seeds are visualized, 
where we show that \ourmodelshort{} has similar randomness in performance to PETS.
Additionally, we visualize \ourmodelshortP{}-BC in Figure~\ref{fig:ablation_study_full} (b),
whose best distribution variance for policy planning is $0.01$, while the best setting for testing is $0.03$.

\subsection{Population Size}\label{appendix:sec_popsize}
\begin{figure}[!t]
    \centering
    \begin{subfigure}{.32\textwidth}
        \centering
        \includegraphics[width=\linewidth]{pcem_4/pwcem_exp_4_cem.pdf}
        (a) CEM
    \end{subfigure}
    \begin{subfigure}{.32\textwidth}
        \centering
        \includegraphics[width=\linewidth]{pcem_4/pwcem_exp_4_pocem.pdf}
        (b) \ourmodelshortA{}
    \end{subfigure}
    \begin{subfigure}{.32\textwidth}
        \centering
        \includegraphics[width=\linewidth]{pcem_4/pwcem_exp_4_pwcem-wra.pdf}
        (c) \ourmodelshortP{}-AVG
    \end{subfigure}%
    
    \begin{subfigure}{.32\textwidth}
        \centering
        \includegraphics[width=\linewidth]{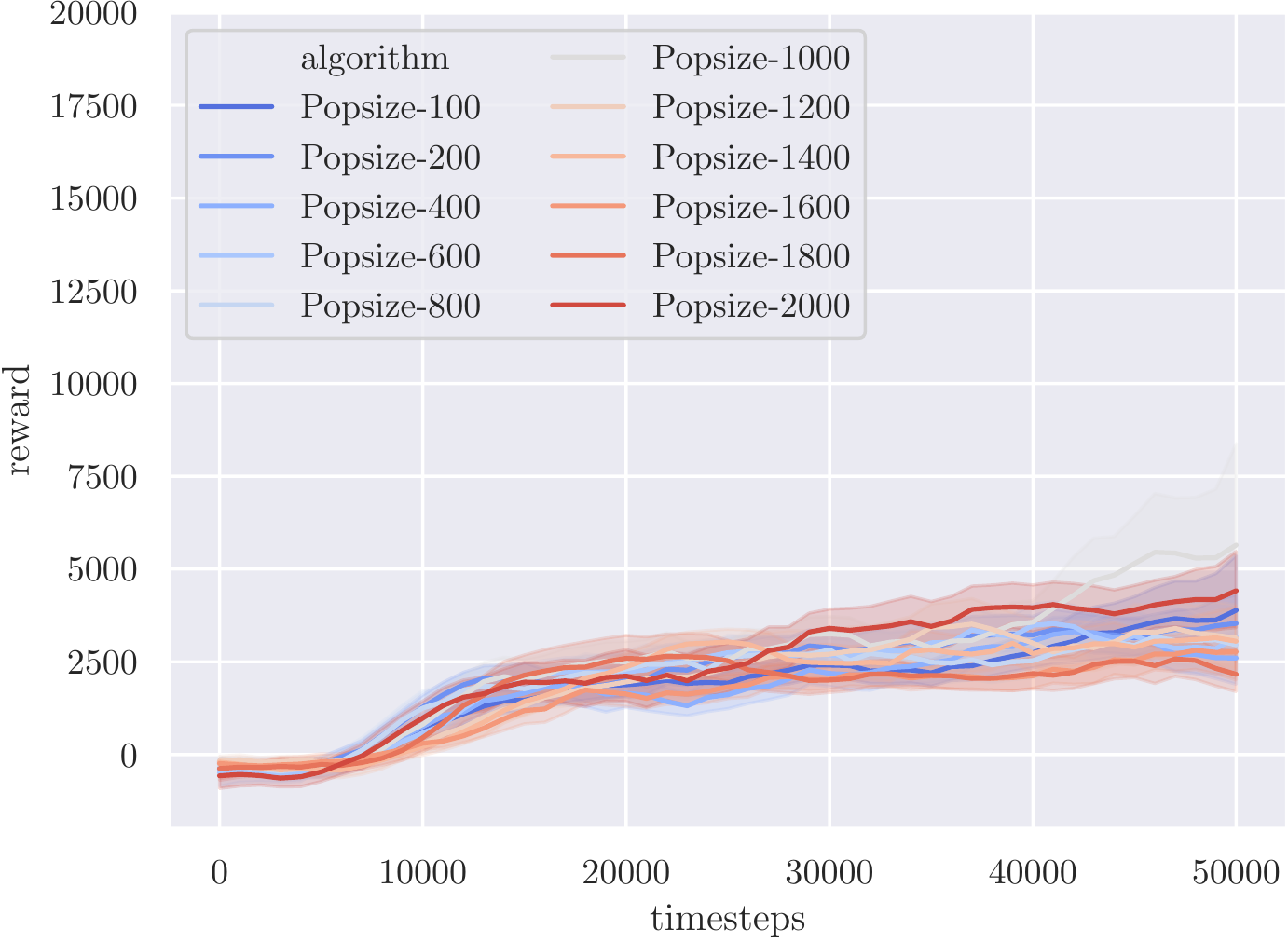}
        (d) \ourmodelshortP{}-BC
    \end{subfigure}%
    \begin{subfigure}{.32\textwidth}
        \centering
        \includegraphics[width=\linewidth]{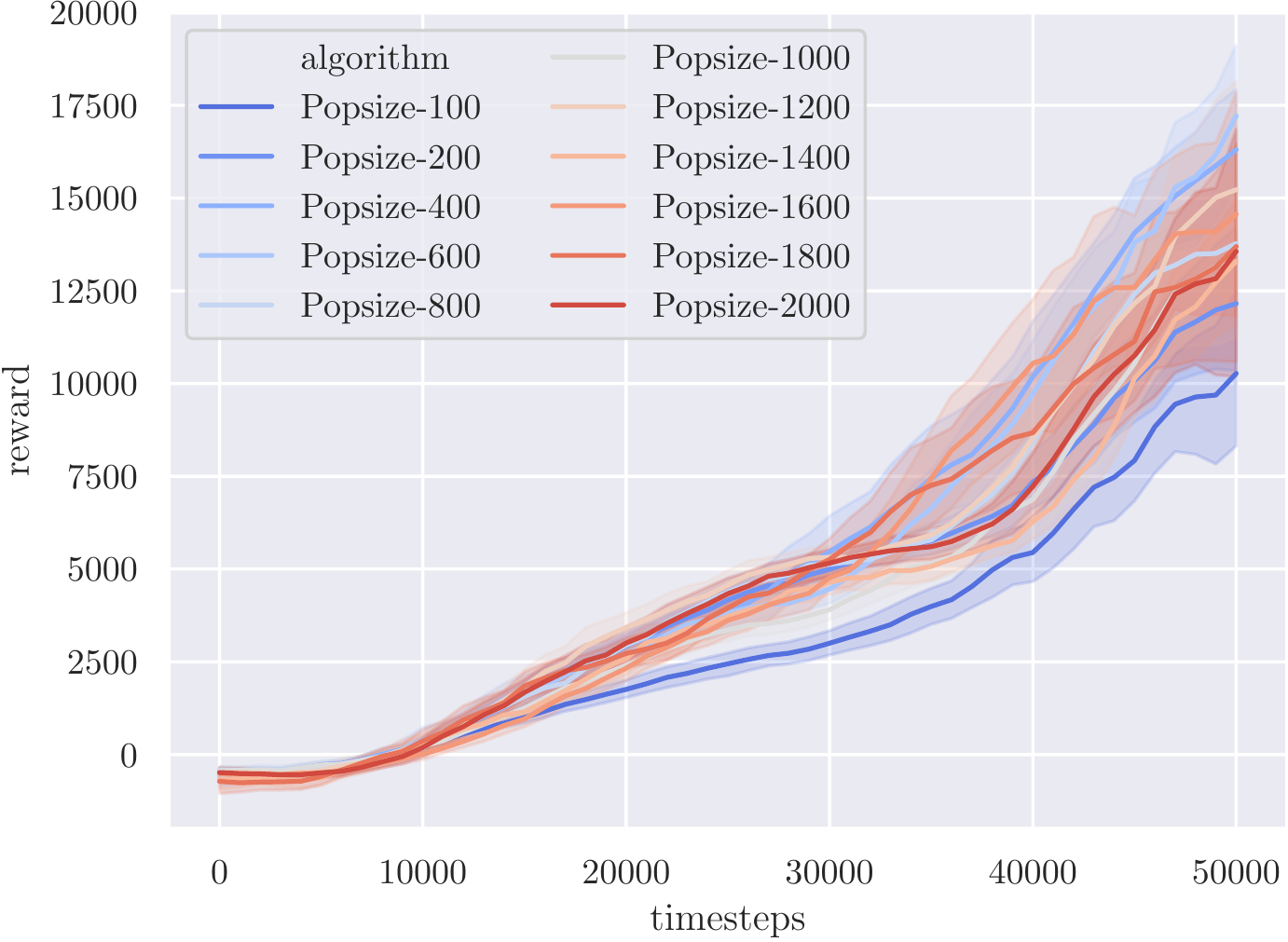}
        (e) \ourmodelshortP{}-ZeroWeight
    \end{subfigure}%
    \caption{The performance of PETS, \ourmodelshortA{}, \ourmodelshortP{}-Avg, \ourmodelshortP{}-BC and \ourmodelshortP{} whose network has fixed parameters of zeros. The variance of the candidates trajectory $\sigma$ in \ourmodelshortP{} is set to 0.1.}
    \label{fig:popsize_appendix}
\end{figure}

In Figure~\ref{fig:popsize_appendix},
we include more detailed figures of the performance of different algorithms with different population size.
One interesting finding is that even with fixed parameters of zeros,
\ourmodelshortP{} can still performance very efficient search.
This is indicating that the efficiency in optimization of \ourmodelshortP{},
especially of \ourmodelshortP{}-AVG, is the key reasons for successful planning.
However, this scheme naturally sacrifices the policy distillation and thus cannot be applied without planning.

\subsection{The Reward Surface of Different Algorithm}\label{appendix:full_reward_surface}
\begin{figure}[!ht]
    \centering
    \begin{subfigure}{.40\textwidth}
        \centering
        \includegraphics[width=\linewidth]{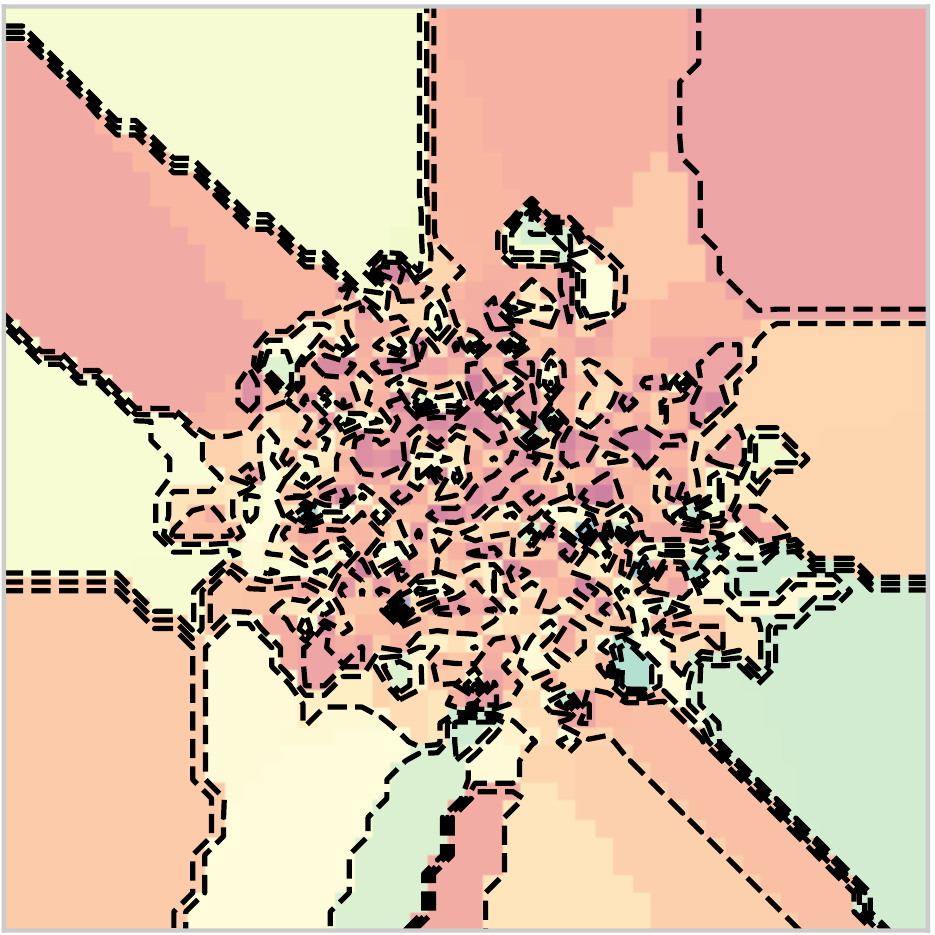}
        (a) PETS reward surface
    \end{subfigure}
    \begin{subfigure}{.40\textwidth}
        \centering
        \includegraphics[width=\linewidth]{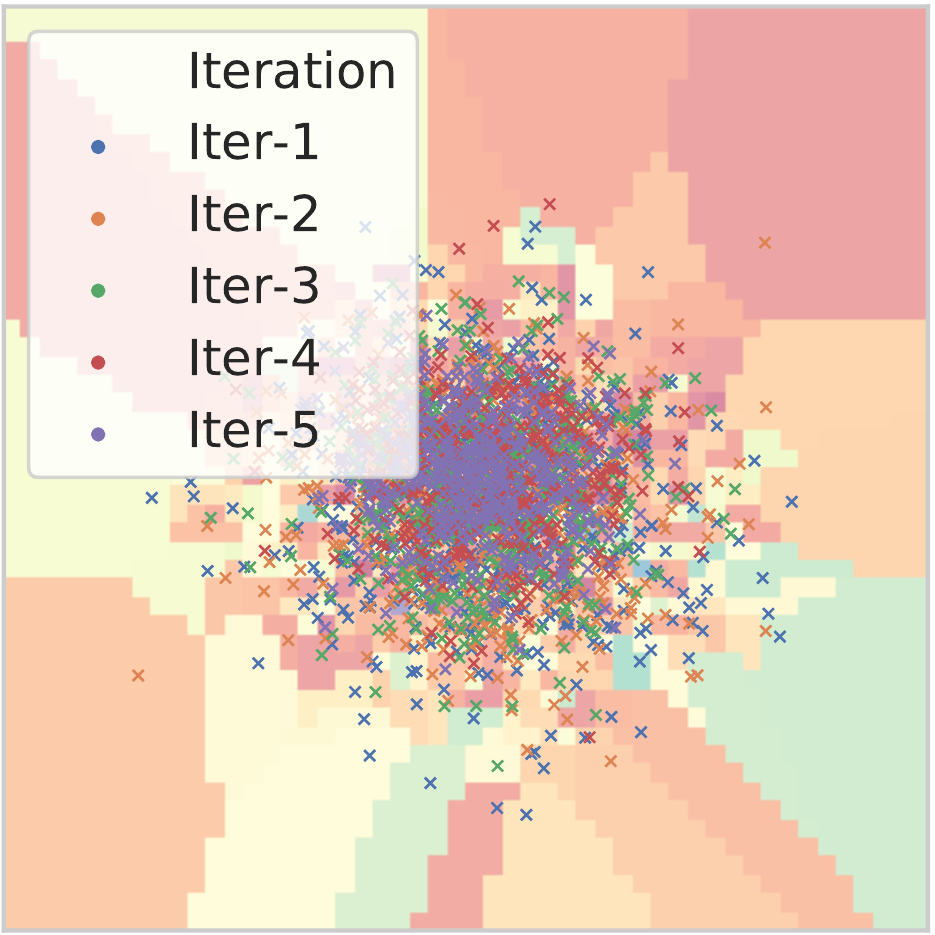}
        (b) PETS candidates
    \end{subfigure}
    \caption{Reward surface in solution space (action space) for PETS algorithm.}
    \label{fig:reward_surface_sol_space_full}
\end{figure}

\begin{figure}[!ht]
    \centering
    \begin{subfigure}{.40\textwidth}
        \centering
        \includegraphics[width=\linewidth]{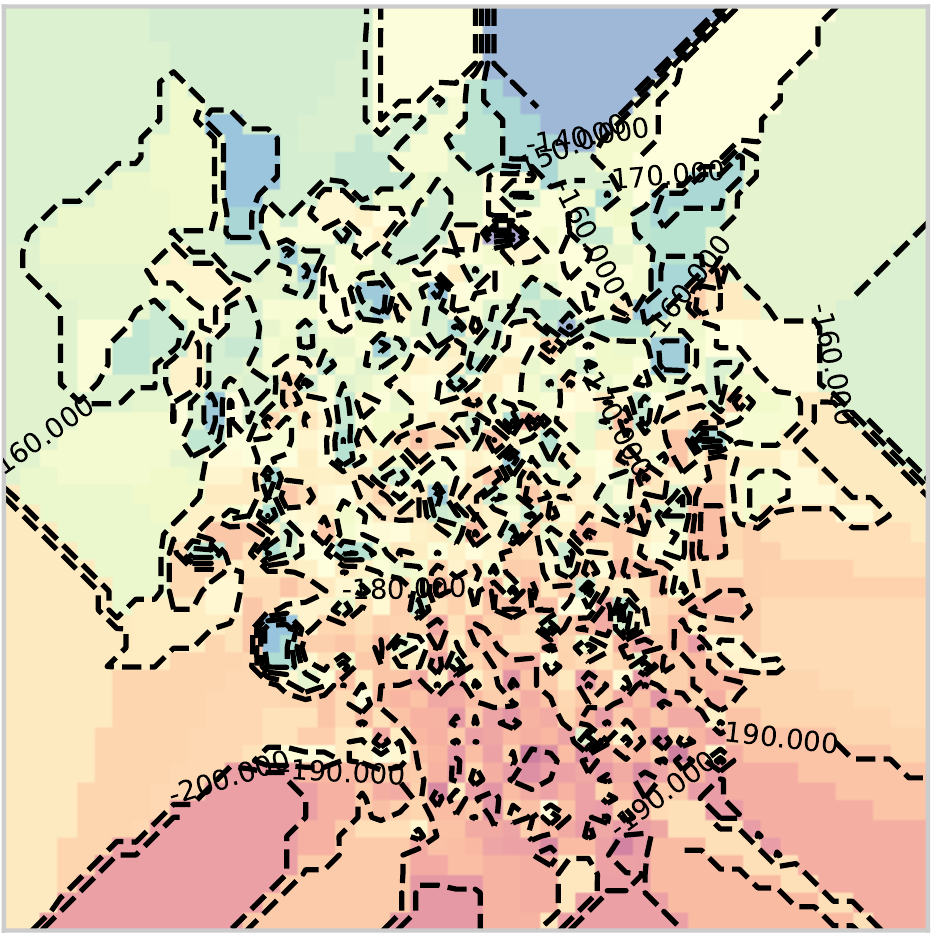}
        (c) \ourmodelshortA{}-Replan reward surface
    \end{subfigure}
    \begin{subfigure}{.40\textwidth}
        \centering
        \includegraphics[width=\linewidth]{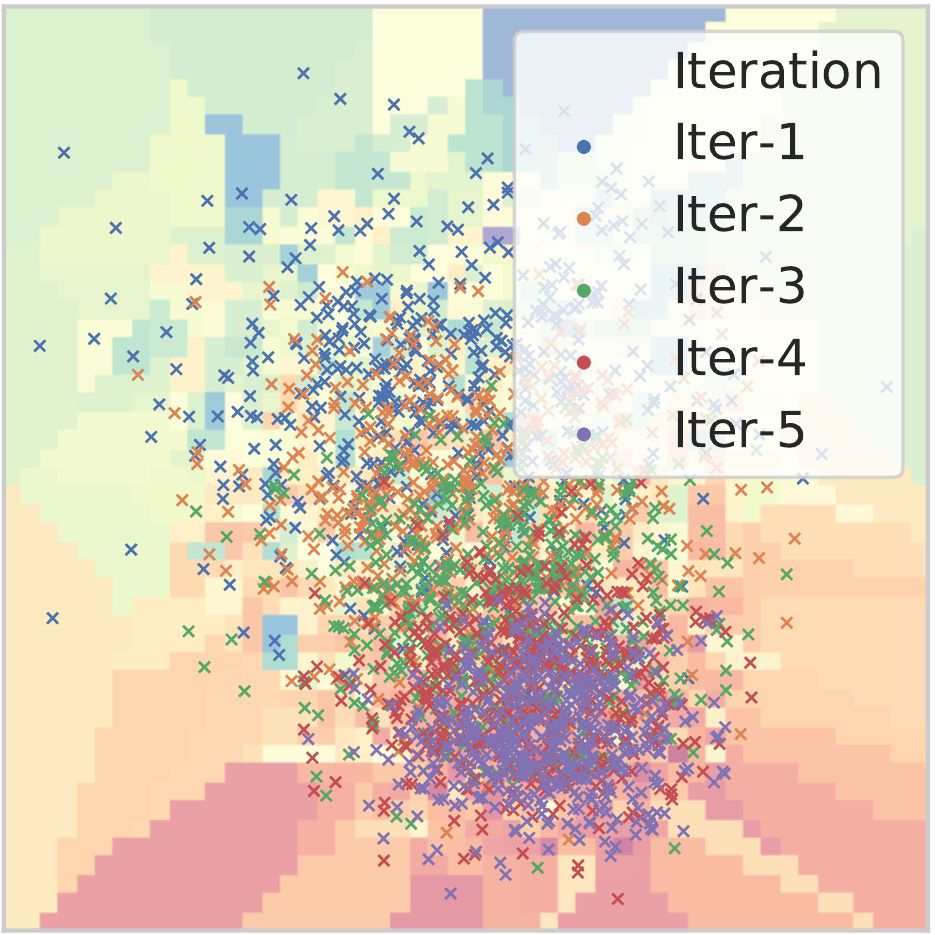}
        (d) \ourmodelshortA{}-Replan candidates
    \end{subfigure}
        \caption{Reward surface in solution space (action space) for \ourmodelshortA{}-Replan.}
    \label{fig:reward_surface_sol_space_full2}
\end{figure}

\begin{figure}[!ht]
    \centering
    \begin{subfigure}{.40\textwidth}
        \centering
        \includegraphics[width=\linewidth]{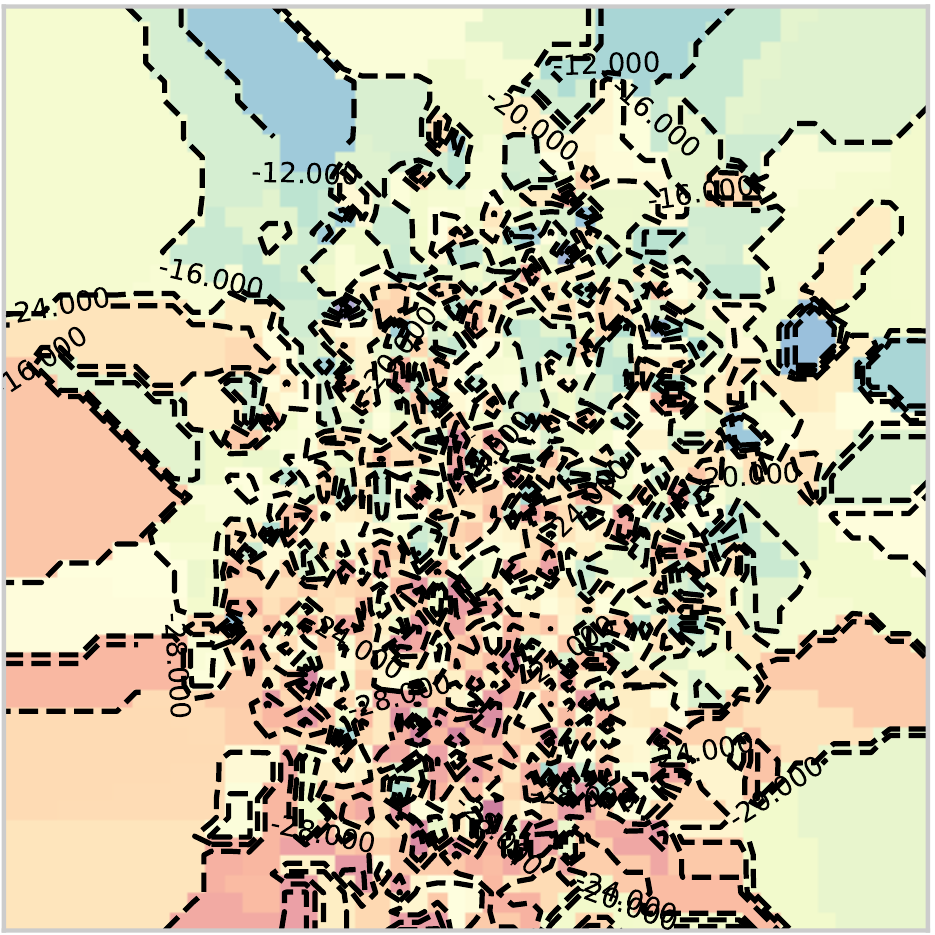}
        (e) \ourmodelshortA{}-Init reward surface
    \end{subfigure}
    \begin{subfigure}{.40\textwidth}
        \centering
        \includegraphics[width=\linewidth]{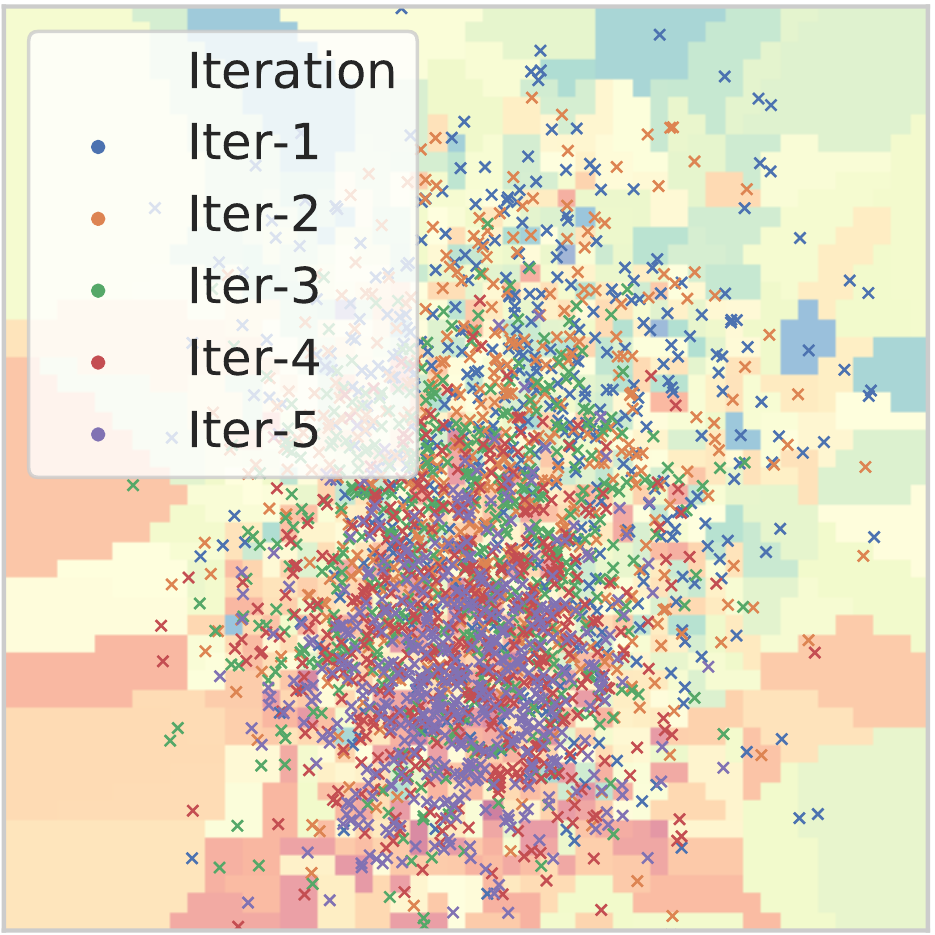}
        (f) \ourmodelshortA{}-Init candidates
    \end{subfigure}
        \caption{Reward surface in solution space (action space) for \ourmodelshortA{}-Init.}
    \label{fig:reward_surface_sol_space_full3}
\end{figure}

\begin{figure}[!ht]
    \centering
    \begin{subfigure}{.40\textwidth}
        \centering
        \includegraphics[width=\linewidth]{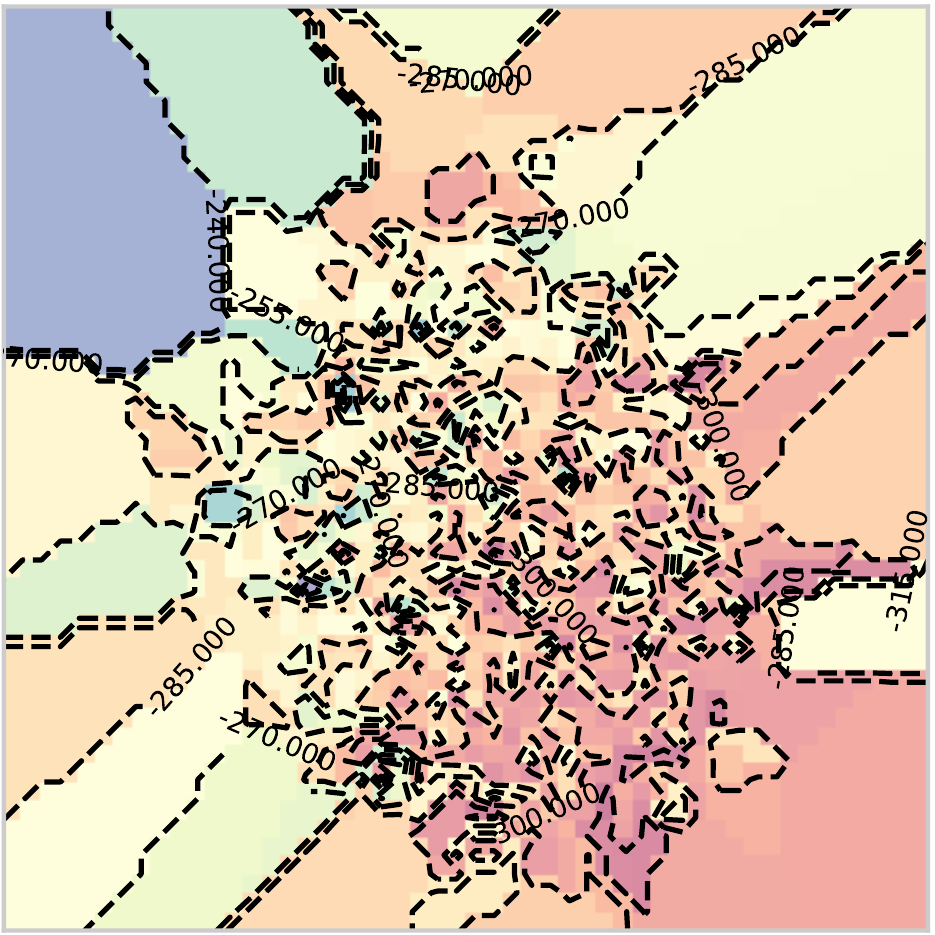}        
        (g) \ourmodelshortP{}-Layer0 reward surface
    \end{subfigure}
    \begin{subfigure}{.40\textwidth}
        \centering
        \includegraphics[width=\linewidth]{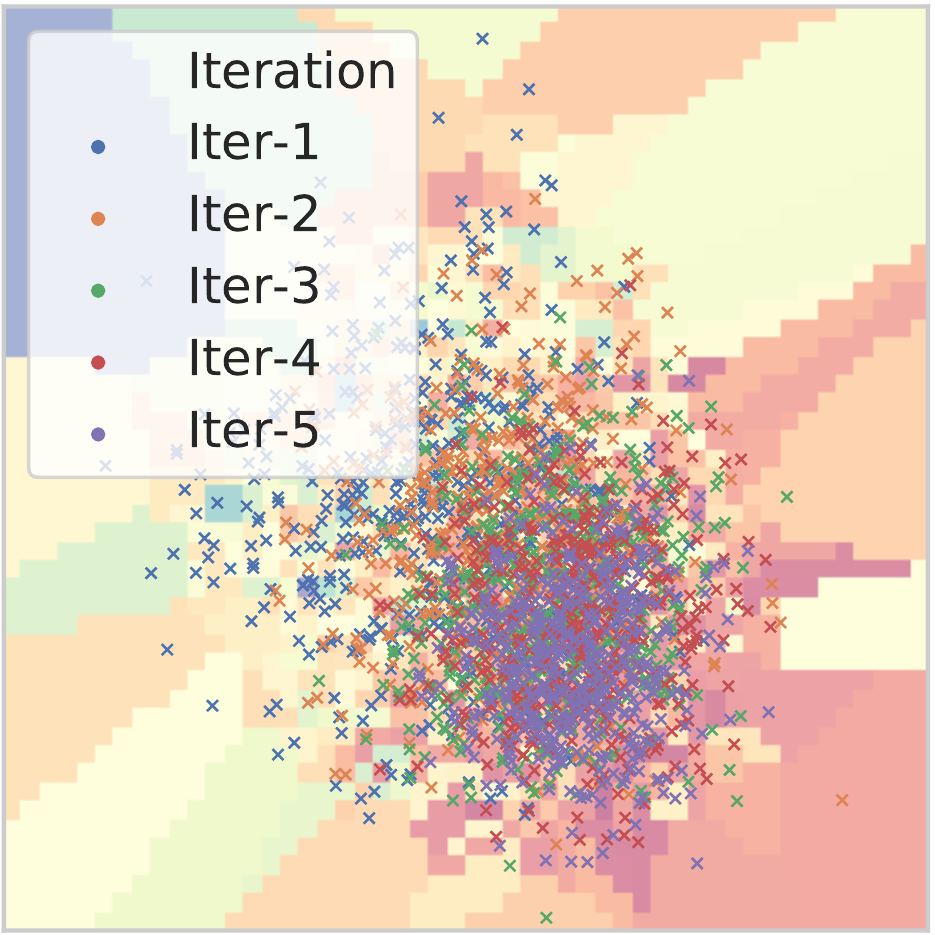}
        (h) \ourmodelshortP{}-Layer0 candidates
    \end{subfigure}
        \caption{Reward surface in solution space (parameter space) for \ourmodelshortP{} with 0 hidden layer.}
    \label{fig:reward_surface_sol_space_full4}
\end{figure}

\begin{figure}[!ht]
    \centering
    \begin{subfigure}{.40\textwidth}
        \centering
        \includegraphics[width=\linewidth]{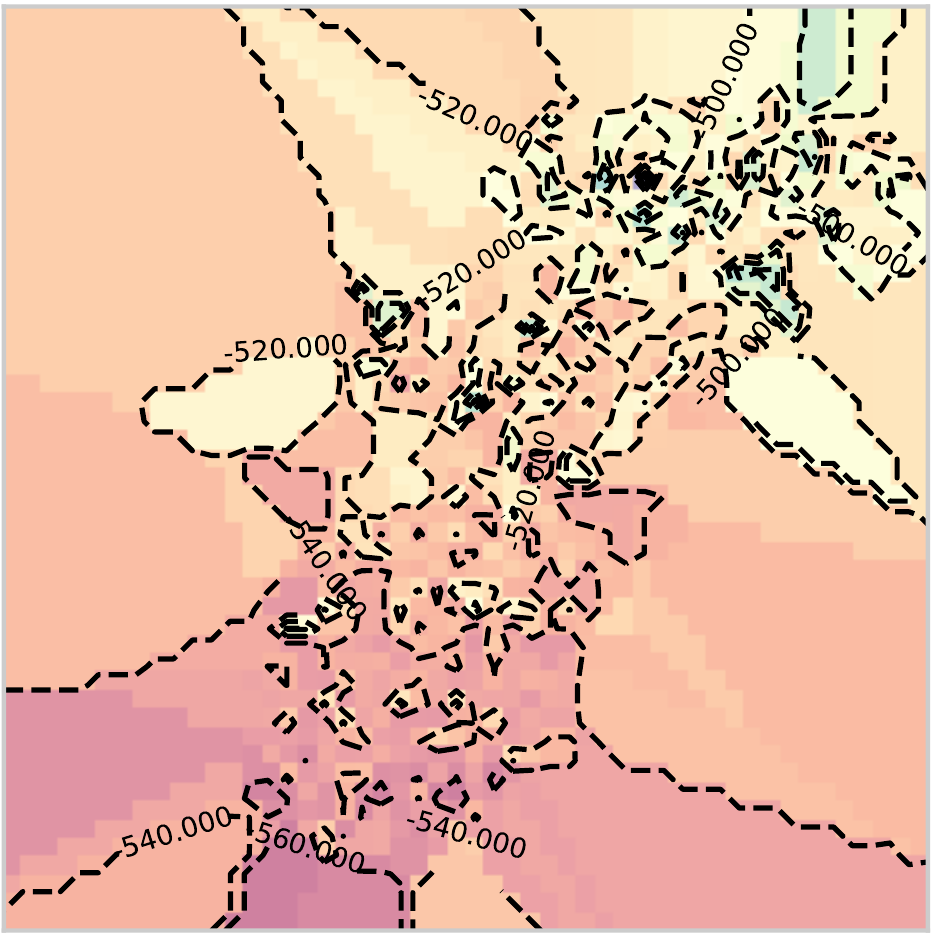}        
        (i) \ourmodelshortP{} reward surface
    \end{subfigure}
    \begin{subfigure}{.40\textwidth}
        \centering
        \includegraphics[width=\linewidth]{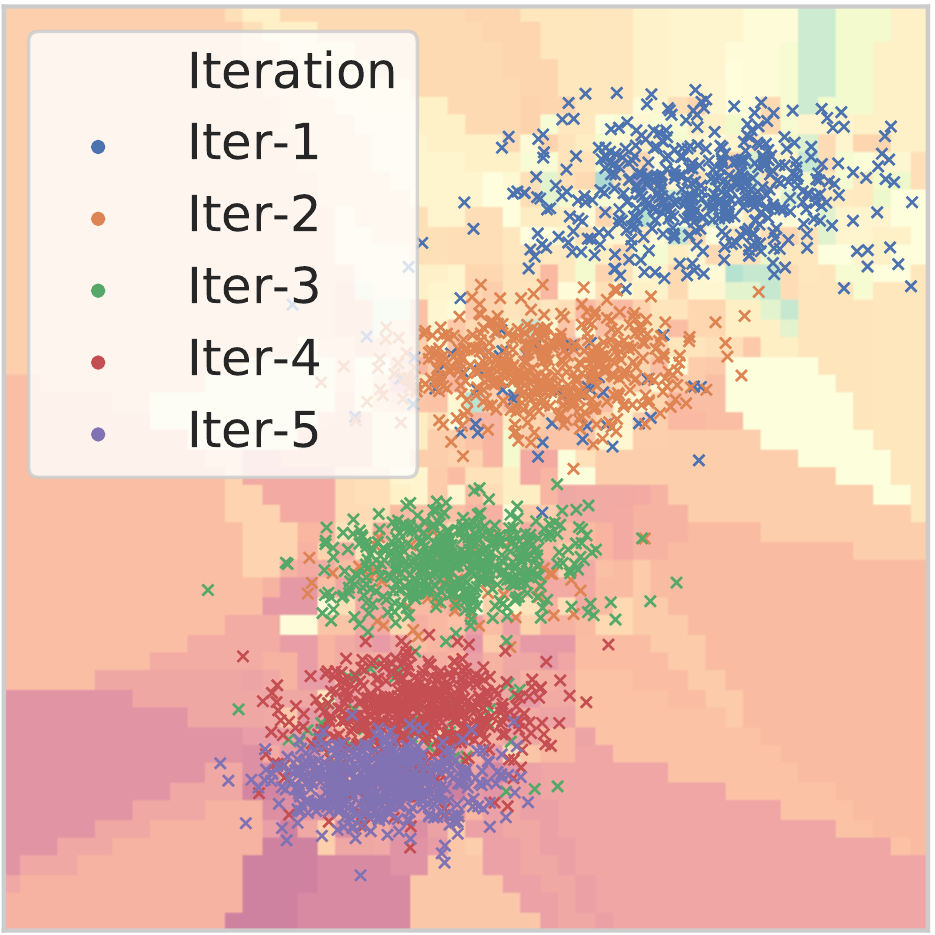}
        (j) \ourmodelshortP{} candidates
    \end{subfigure}
        \caption{Reward surface in solution space (parameter space) for \ourmodelshortP{} using 1 hidden layer.}
    \label{fig:reward_surface_sol_space_full5}
\end{figure}

In this section, we provide a more detailed description of the reward surface with respect the the solution space (action space for PETS and \ourmodelshortA{}, and parameter space for \ourmodelshortP{}) in Figure~\ref{fig:reward_surface_sol_space_full},~\ref{fig:reward_surface_sol_space_full2}, \ref{fig:reward_surface_sol_space_full3},~\ref{fig:reward_surface_sol_space_full4}, \ref{fig:reward_surface_sol_space_full5}.
As we can see, variants of \ourmodelshortA{} are better at searching,
but the reward surface is still not smooth.
\ourmodelshortA{}-Replan is more efficient in searching than \ourmodelshortA{}-Init,
but the errors in dynamics limit its performance.
We also include the results for \ourmodelshortP{} using a 1-layer neural network in solution space in Figure~\ref{fig:reward_surface_sol_space_full4} (g), (h).
The results indicate that the deeper the network, the better the search efficiency.

\begin{figure}[!ht]
    \centering    
    \begin{subfigure}{0.32\textwidth}
        \centering
        \includegraphics[width=\linewidth]{pcem_5/exp_5_pwcemsearch_action_cem.pdf}
        (a) PETS Surface
    \end{subfigure}
    \begin{subfigure}{0.32\textwidth}
        \centering
        \includegraphics[width=\linewidth]{pcem_5/exp_5_pwcemsearch_action_pwcem.pdf}
        (b) POLINA Surface
    \end{subfigure}
    \begin{subfigure}{0.32\textwidth}
        \centering
        \includegraphics[width=\linewidth]{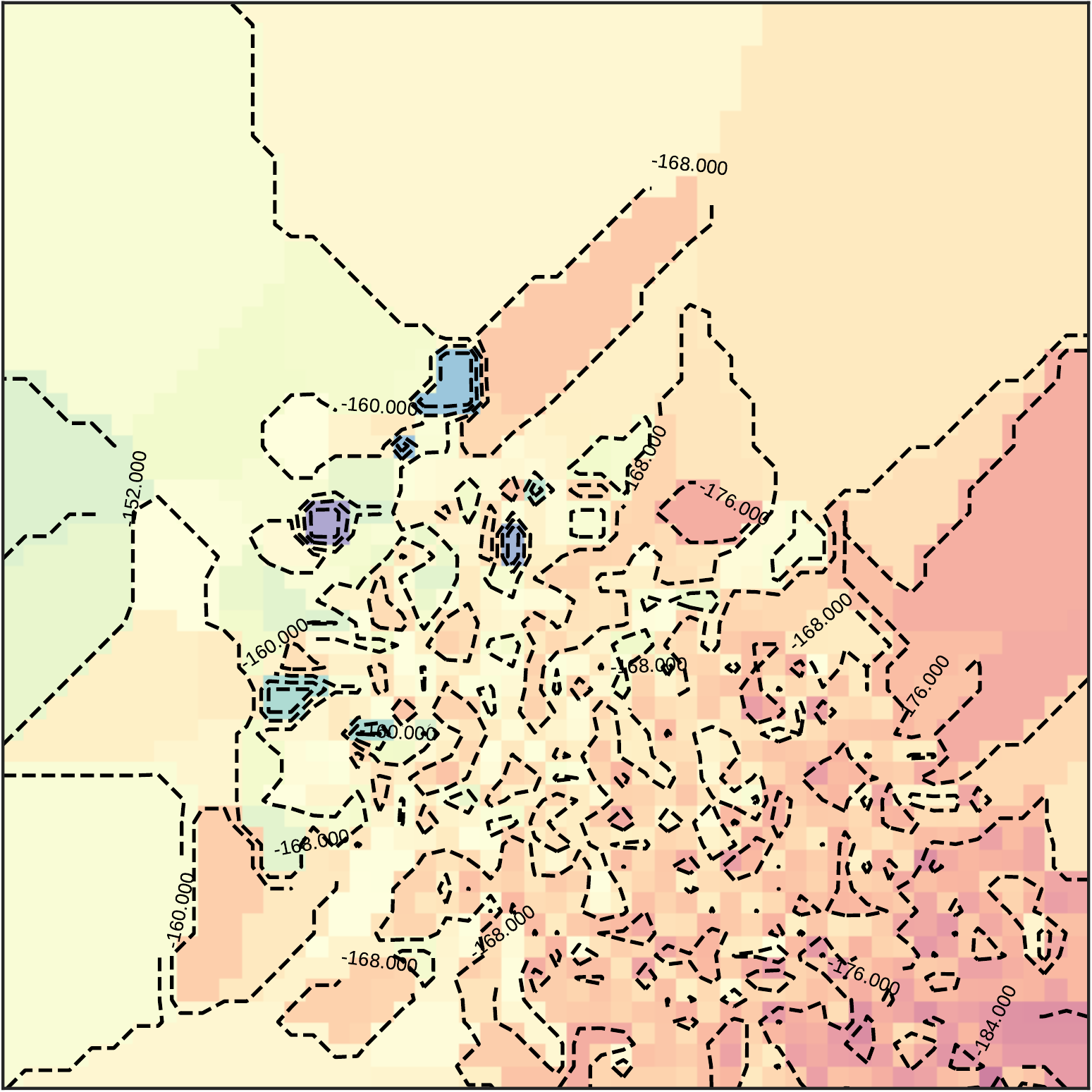}
        (c) POLINA-L0 Surface
    \end{subfigure}
    \caption{The color indicates the expected cost (negative of expected reward).
    We emphasis that all these figures are visualized in the action space. And all of them are very unsmooth.
    For the figures visualized in solution space, we refer to Figure~\ref{fig:reward_surface_sol_space_full}.
    }
    \label{fig:reward_surface_appendix}
\end{figure}

\begin{figure}[!ht]
    \begin{subfigure}{0.19\textwidth}
        \centering
        \includegraphics[width=\linewidth]{pcem_5/exp_5_pwcemsearch_actioncem_iter0.pdf}
        (a1) PETS Iteration 1
    \end{subfigure}
    \begin{subfigure}{0.19\textwidth}
        \centering
        \includegraphics[width=\linewidth]{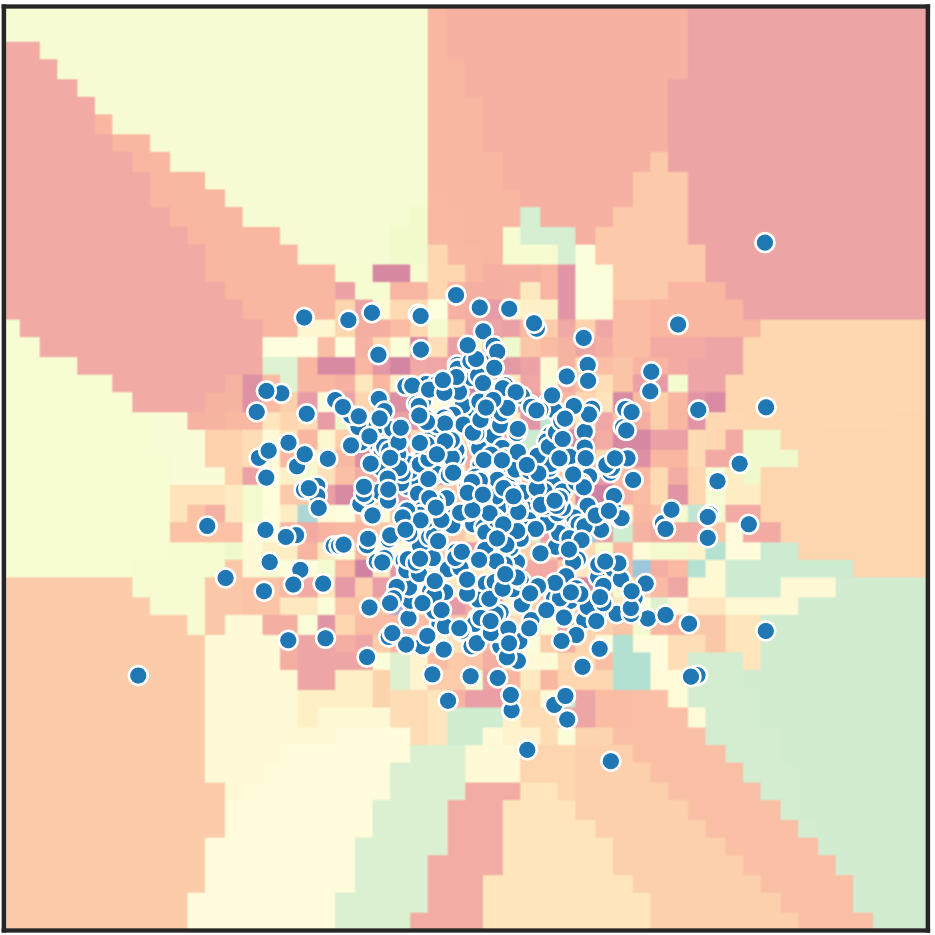}
        (a2) PETS Iteration 2
    \end{subfigure}
    \begin{subfigure}{0.19\textwidth}
        \centering
        \includegraphics[width=\linewidth]{pcem_5/exp_5_pwcemsearch_actioncem_iter2.pdf}
        (a3) PETS Iteration 3
    \end{subfigure}
    \begin{subfigure}{0.19\textwidth}
        \centering
        \includegraphics[width=\linewidth]{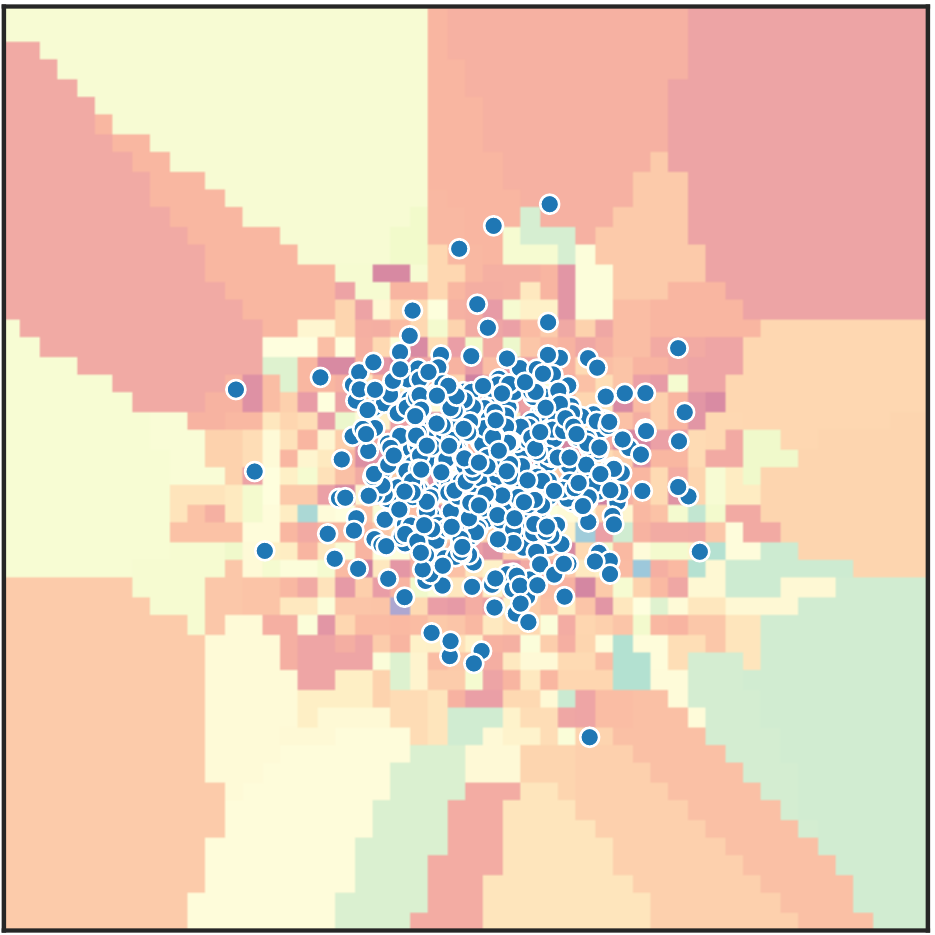}
        (a4) PETS Iteration 4
    \end{subfigure}
    \begin{subfigure}{0.19\textwidth}
        \centering
        \includegraphics[width=\linewidth]{pcem_5/exp_5_pwcemsearch_actioncem_iter4.pdf}
        (a5) PETS Iteration 5
    \end{subfigure}
    \caption{The figures are the planned trajectories of PETS.}
    \label{fig:reward_surface_appendix1}
\end{figure}

\begin{figure}[!ht]
    \begin{subfigure}{0.19\textwidth}
        \centering
        \includegraphics[width=\linewidth]{pcem_5/exp_5_pwcemsearch_actionpwcem_iter0.pdf}
        (b1) \ourmodelshort{} Iteration 1
    \end{subfigure}
    \begin{subfigure}{0.19\textwidth}
        \centering
        \includegraphics[width=\linewidth]{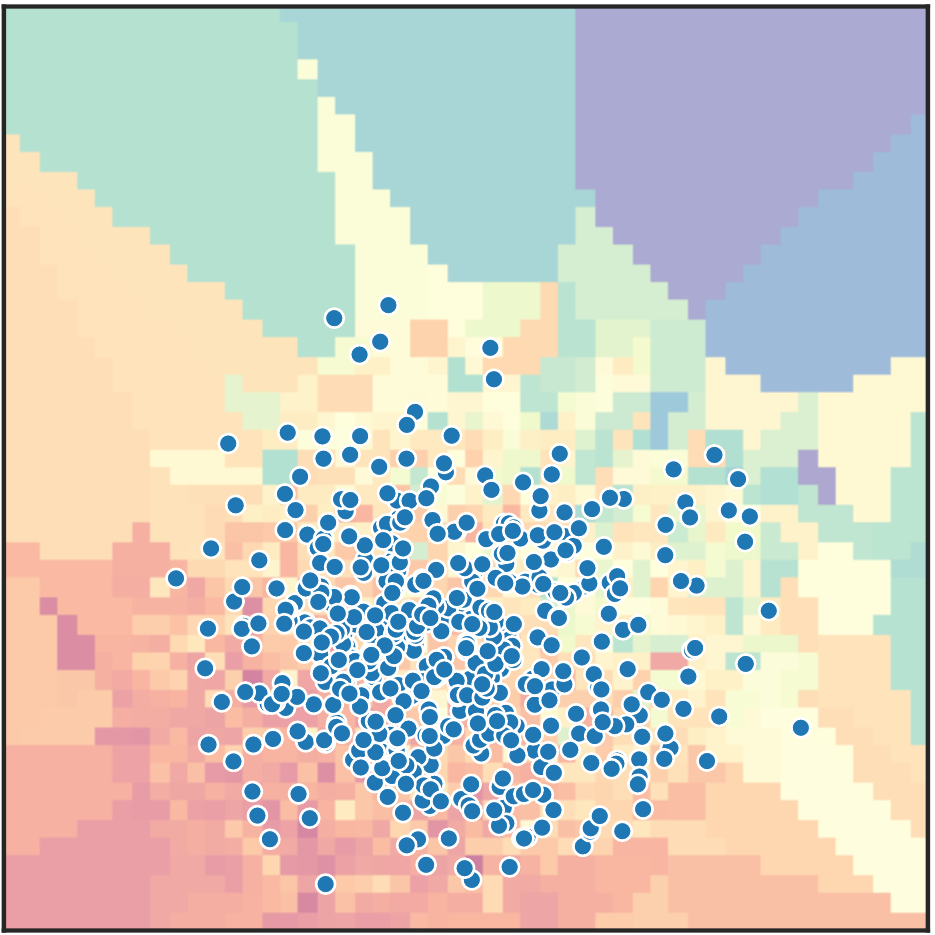}
        (b2) \ourmodelshort{} Iteration 2
    \end{subfigure}
    \begin{subfigure}{0.19\textwidth}
        \centering
        \includegraphics[width=\linewidth]{pcem_5/exp_5_pwcemsearch_actionpwcem_iter2.pdf}
        (b3) \ourmodelshort{} Iteration 3
    \end{subfigure}
    \begin{subfigure}{0.19\textwidth}
        \centering
        \includegraphics[width=\linewidth]{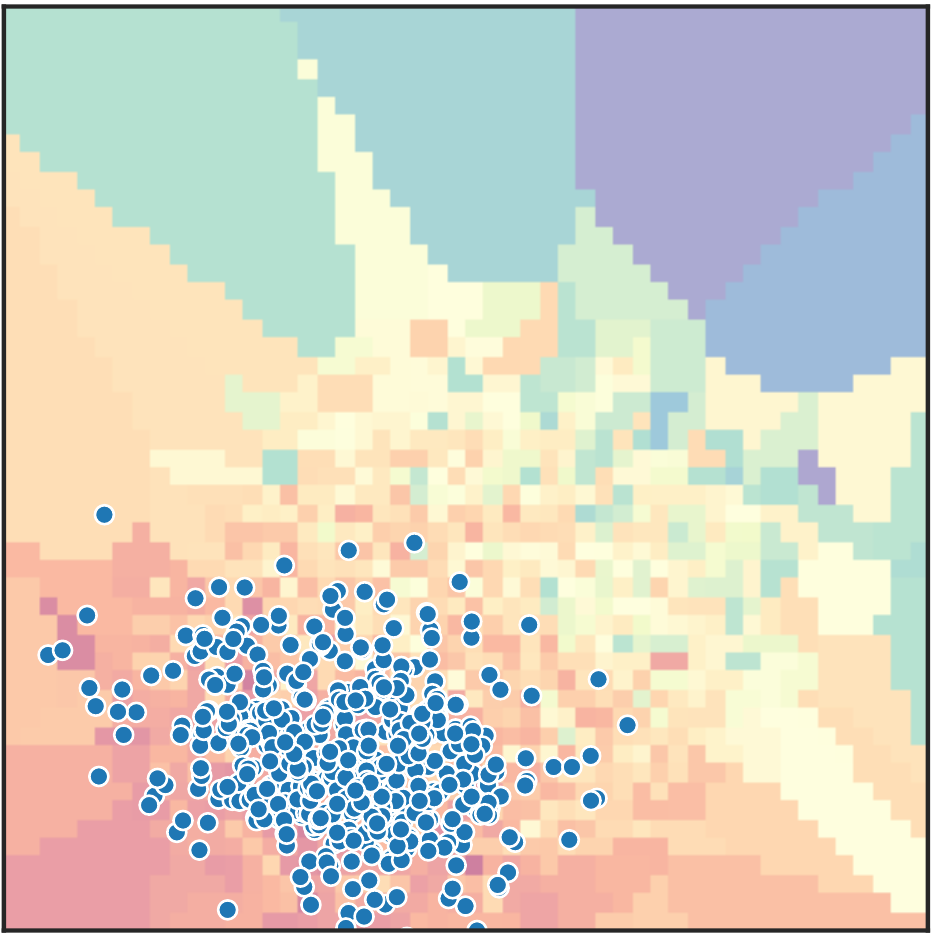}
        (b4) \ourmodelshort{} Iteration 4
    \end{subfigure}
    \begin{subfigure}{0.19\textwidth}
        \centering
        \includegraphics[width=\linewidth]{pcem_5/exp_5_pwcemsearch_actionpwcem_iter4.pdf}
        (b5) \ourmodelshort{} Iteration 5
    \end{subfigure}
    \caption{The figures are the planned trajectories of \ourmodelshort{}-P using 1 hidden layer MLP.}
    \label{fig:reward_surface_appendix2}
\end{figure}

\begin{figure}[!ht]
    \begin{subfigure}{0.19\textwidth}
        \centering
        \includegraphics[width=\linewidth]{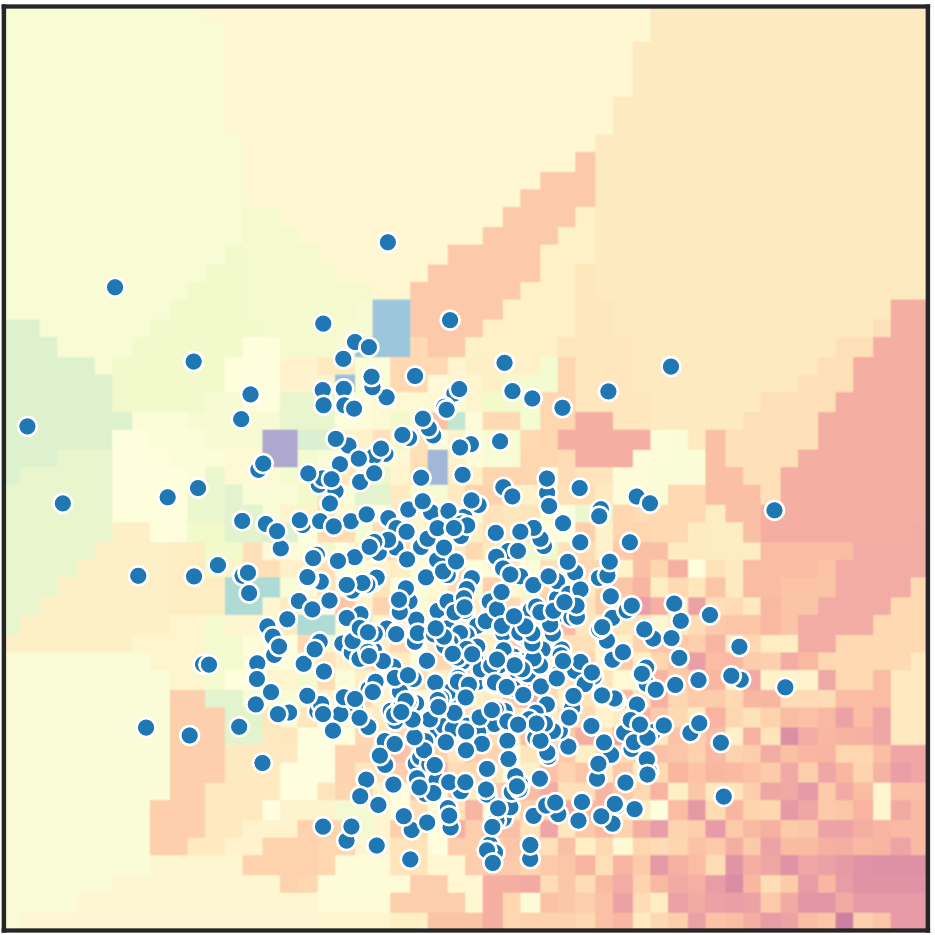}
        (b2) \ourmodelshort{}-L0 Iteration 1
    \end{subfigure}
    \begin{subfigure}{0.19\textwidth}
        \centering
        \includegraphics[width=\linewidth]{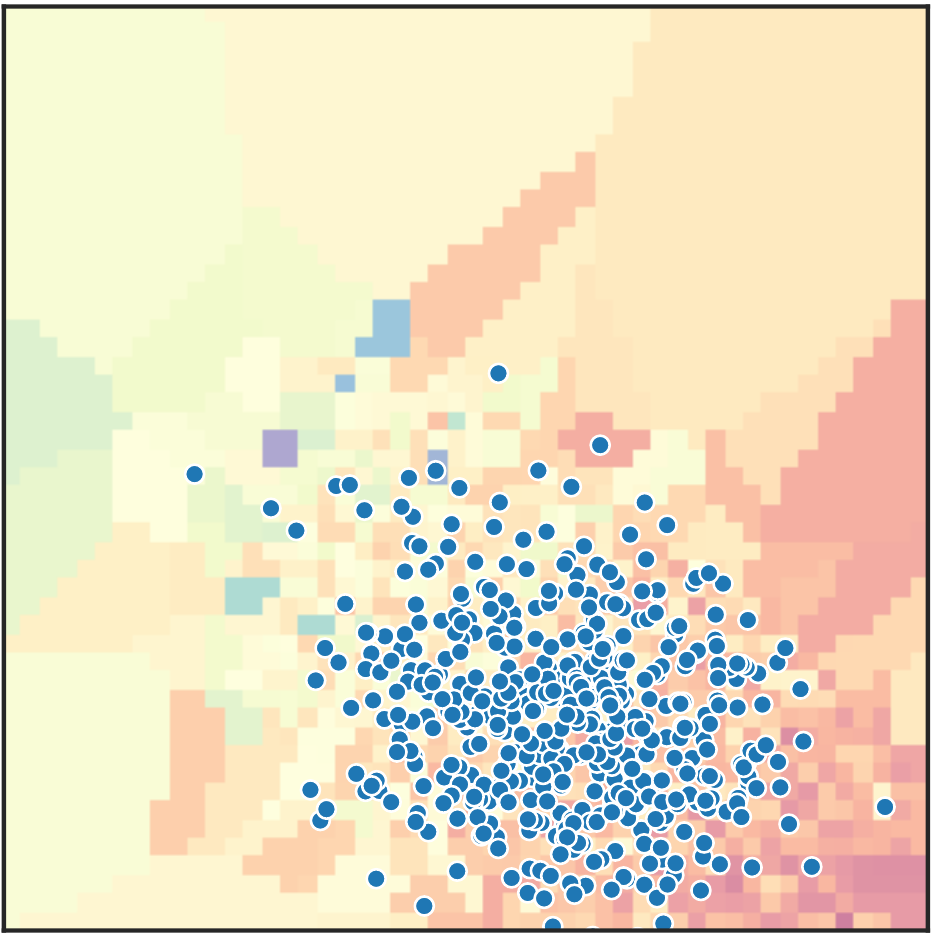}
        (b2) \ourmodelshort{}-L0 Iteration 2
    \end{subfigure}
    \begin{subfigure}{0.19\textwidth}
        \centering
        \includegraphics[width=\linewidth]{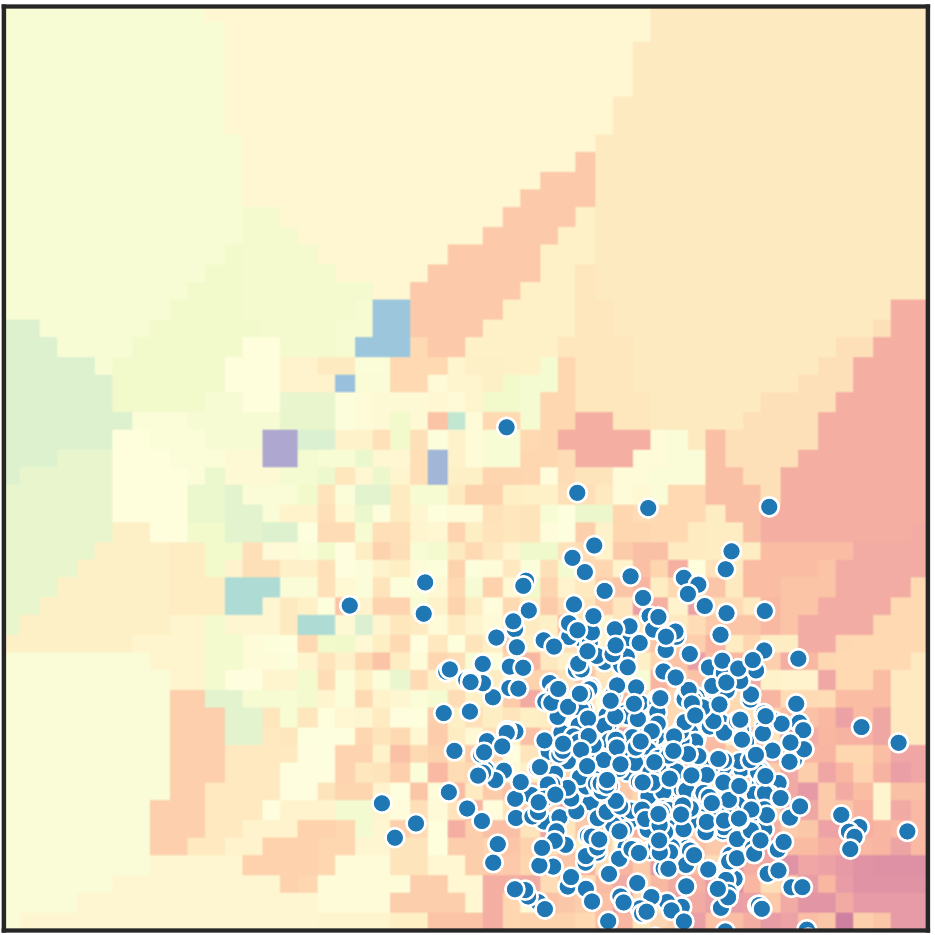}
        (b3) \ourmodelshort{}-L0 Iteration 3
    \end{subfigure}
    \begin{subfigure}{0.19\textwidth}
        \centering
        \includegraphics[width=\linewidth]{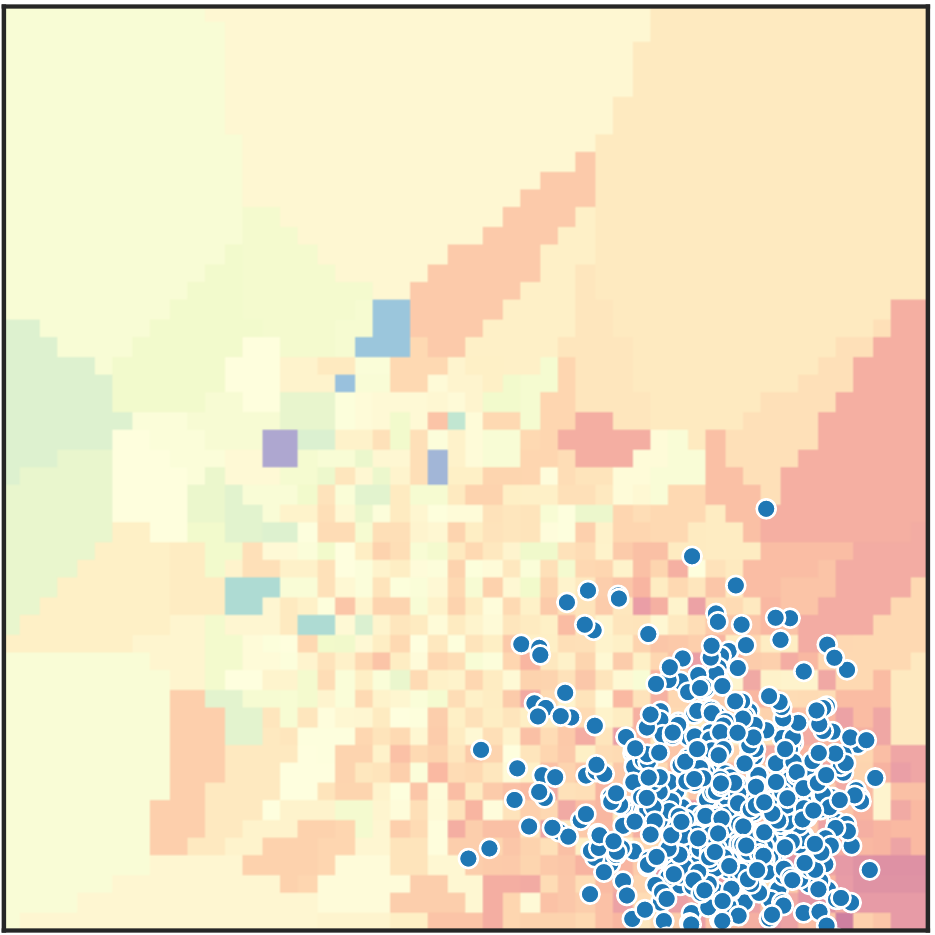}
        (b4) \ourmodelshort{}-L0 Iteration 4
    \end{subfigure}
    \begin{subfigure}{0.19\textwidth}
        \centering
        \includegraphics[width=\linewidth]{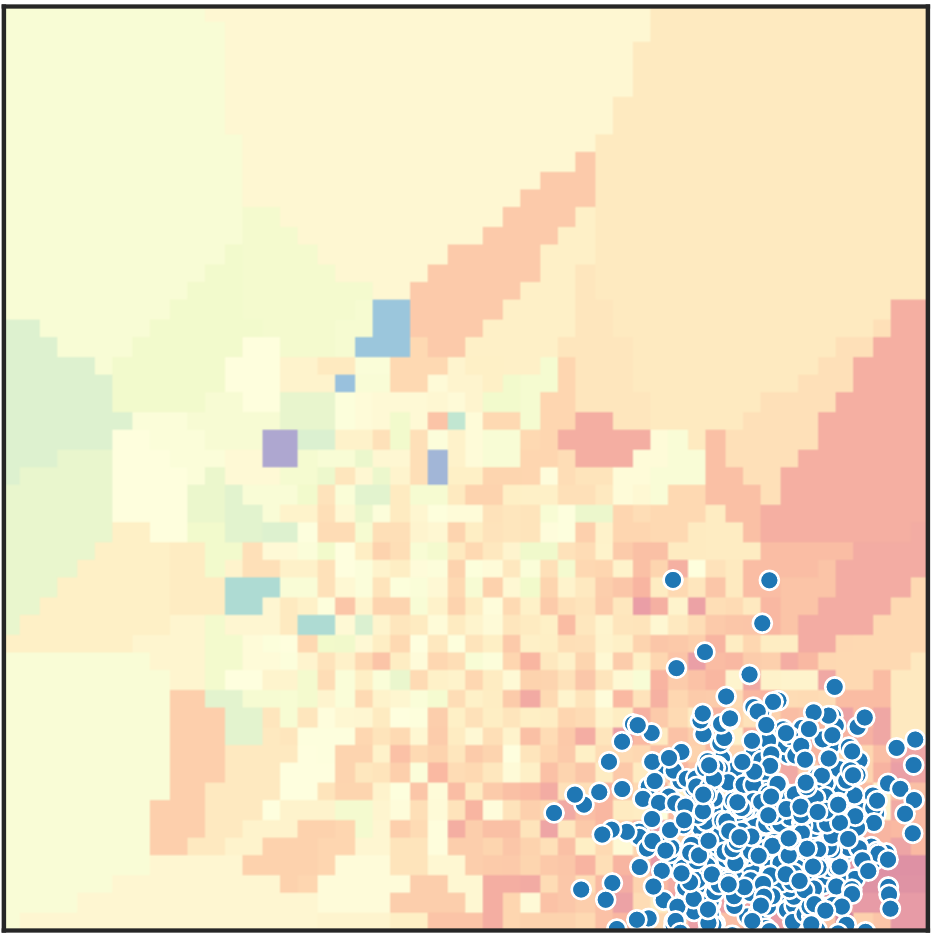}
        (b4) \ourmodelshort{}-L0 Iteration 5
    \end{subfigure}
    
    \caption{The figures are the planned trajectories of \ourmodelshort{}-P using 0 hidden layer MLP.}
    \label{fig:reward_surface_appendix3}
\end{figure}

We also provide more detailed version of Figure~\ref{fig:reward_surface} in Figure~\ref{fig:reward_surface_appendix}.
We respectively show the surface for PETS, \ourmodelshortP{}-P using 1 and 0 hidden layers.
Their planned trajectories across different CEM updates are visualized in Figure~\ref{fig:reward_surface_appendix1},~\ref{fig:reward_surface_appendix2},~\ref{fig:reward_surface_appendix3}.
Originally in Figure~\ref{fig:reward_surface}, we use the trajectories in iteration 1, 3, 5 for better illustration.
In the appendix, we also provide all the iteration data.
Again, the color indicates the expected cost (negative of expected reward).
From left to right, we show the updated the trajectories in each iteration with blue scatters.

\end{document}